%% file: main.tex
\documentclass{article}
\usepackage{graphicx}

\usepackage{fullpage}
\usepackage{microtype}
\usepackage{newtxtext}
\usepackage[utf8]{inputenc}
\usepackage[T1]{fontenc}
\usepackage{textcomp}
\usepackage{authblk}
\usepackage[numbers]{natbib}

\usepackage{xspace}
\usepackage{siunitx}
\usepackage{enumerate}
\usepackage{url}

\usepackage{booktabs}
\usepackage{multirow}
\usepackage{multicol}
\usepackage{makecell}
\usepackage{xcolor,colortbl}
\definecolor{Gray}{gray}{0.90}
\newcolumntype{g}{>{\columncolor{Gray}}c}

\usepackage{mathmacros}
\usepackage{hyperref}
\definecolor{mydarkblue}{rgb}{0,0.08,0.45}
\hypersetup{ %
    pdftitle={},
    pdfsubject={},
    pdfkeywords={},
    pdfborder=0 0 0,
    pdfpagemode=UseNone,
    colorlinks=true,
    linkcolor=mydarkblue,
    citecolor=mydarkblue,
    filecolor=mydarkblue,
    urlcolor=mydarkblue,
}
\usepackage[capitalize, noabbrev]{cleveref}

\crefname{theoremmd}{Theorem}{Theorems}
\crefname{propositionmd}{Proposition}{Propositions}
\crefname{corollarymd}{Corollary}{Corollaries}
\crefname{lemmamd}{Lemma}{Lemmas}
\crefname{definitionmd}{Definition}{Definitions}

\usepackage{autonum}

\title{Probability Bounding: Post-Hoc Calibration \\via Box-Constrained Softmax}
\author[1,3]{Kyohei Atarashi\thanks{Corresponding Author: atarashi@i.kyoto-u.ac.jp}}
\author[2,3]{Satoshi Oyama}
\author[3]{Hiromi Arai}
\author[1,3]{Hisashi Kashima}
\affil[1]{Kyoto University}
\affil[2]{Nagoya City University}
\affil[3]{RIKEN AIP}
\date{\today}

\begin{document}
\sloppy
\maketitle
\begin{abstract}
    Many studies have observed that modern neural networks achieve high accuracy while producing poorly calibrated probabilities, making calibration a critical practical issue.
    In this work, we propose \textbf{probability bounding} (PB), a novel post-hoc calibration method that mitigates both underconfidence and overconfidence by learning lower and upper bounds on the output probabilities.
    To implement PB, we introduce the \textbf{box-constrained softmax} ($\BCSoftmax$) function, a generalization of $\Softmax$ that explicitly enforces lower and upper bounds on the output probabilities.
    While $\BCSoftmax$ is formulated as the solution to a box-constrained optimization problem, we develop an exact and efficient algorithm for computing $\BCSoftmax$.
    We further provide theoretical guarantees for PB and introduce two variants of PB.
    We demonstrate the effectiveness of our methods experimentally on four real-world datasets, consistently reducing calibration errors.
    Our Python implementation is available at \url{https://github.com/neonnnnn/torchbcsoftmax}.
\end{abstract}

\section{Introduction}
    \label{sec:intro}

    \begin{figure}[t]
        \centering
        \includegraphics[width=0.7\linewidth]{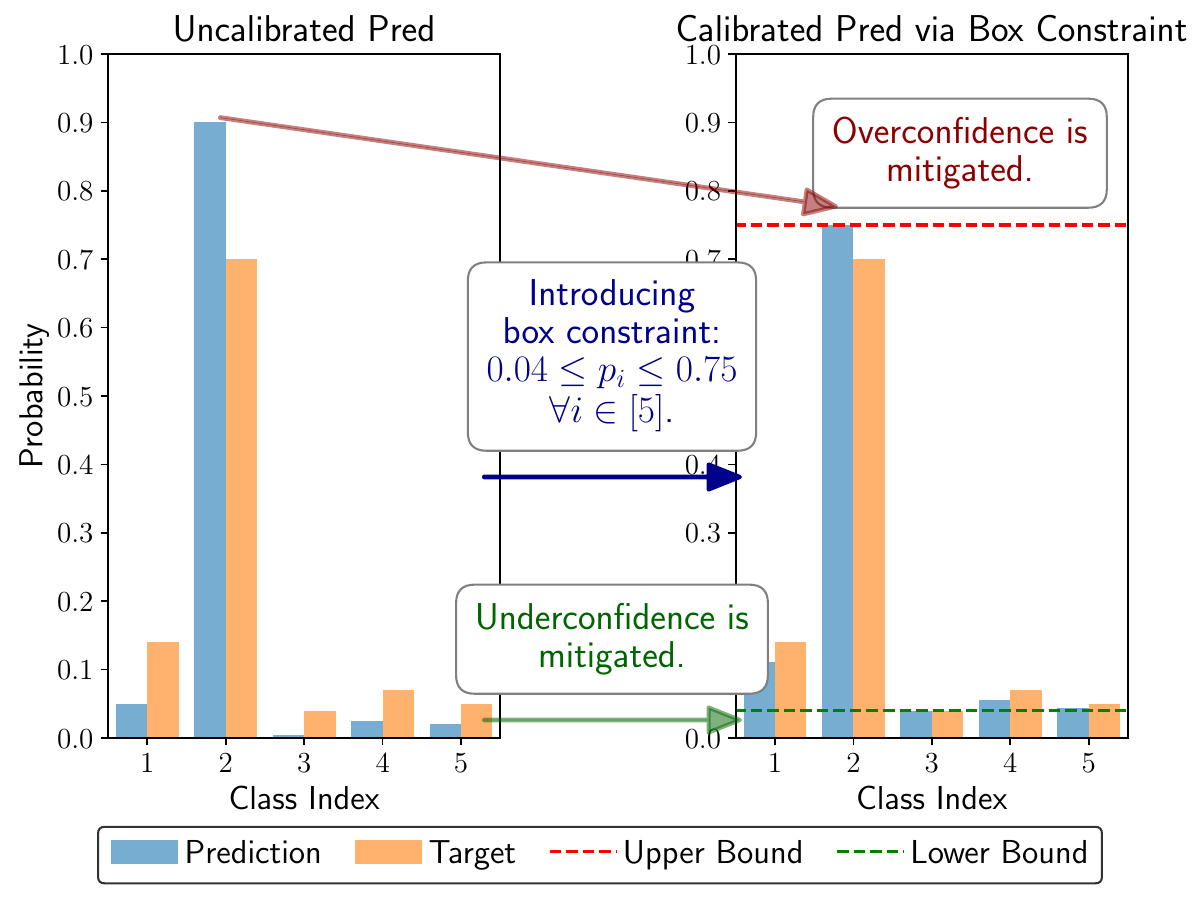}
        \caption{Probability calibration via box constraint. The left figure shows the predicted probabilities $\bm{p}$ and the target probabilities. The right figure shows the predicted probabilities with the box constraint $0.04 \le p_i \le 0.75$. Enforcing this box constraint mitigates the overconfidence of the top-label (argmax) prediction and the underconfidence of the other predictions. This mechanism directly enhances the reliability and trustworthiness of the prediction.}
        \label{fig:basic_idea}
    \end{figure}

    Modern machine learning methods, especially neural networks (NNs), have achieved high accuracy on various tasks, but it has been shown that they tend to be overconfident or underconfident in their predictions, i.e., their output probabilities do not match the true probabilities.
    This leads to issues in reliability and trustworthiness, e.g., it can result in inappropriate decisions~\citep{kull2017beta,zadrozny2002transforming} and cause information leakage from the training data~\citep{chen2023overconfidence,fu2024membership}.

    To address this issue, many studies have developed \emph{probability calibration} methods for classification models~\citep{gyusang2024tilt,ghosh2022adafocal,guo2017calibration,kull2019beyond,kumar2018trainable,ma2021meta,tao2025feature}.
    Among them, post-hoc calibration (recalibration) has attracted growing attention and has been widely used across various tasks due to its practical performance and ease of use~\citep{guo2017calibration,kull2019beyond,ma2021meta,tao2025feature,tomani2022parameterized}.
    Post-hoc calibration methods typically calibrate a trained model by fitting only a small number of additional parameters to a validation (calibration) set with negligible impact on classification accuracy.
    
    However, most existing post-hoc methods provide only \emph{soft control} over the output probabilities and cannot enforce \emph{hard constraints}.
    It is often desirable to impose explicit bounds on output probabilities to enable finer-grained and more reliable control in downstream decision-making.
    For example, in portfolio allocation, investment size is often determined as an explicit function of predicted success probabilities (e.g., Kelly-type criteria~\citep{kelly1956new}), which are known to be highly sensitive to probability estimation errors; overconfident probability estimates can therefore lead to excessively large allocations and catastrophic losses.
    This motivates the use of hard constraints on output probabilities, which can also be beneficial in other safety-critical applications such as medical diagnosis.

    Motivated by this limitation, in this study, we propose a novel post-hoc calibration method, \textbf{probability bounding} (\textbf{PB}).
    Our key idea of PB is to mitigate overconfident and underconfident predictions by enforcing box constraints on the output probabilities, as illustrated in~\cref{fig:basic_idea}.
    In general, it is difficult for practitioners to set appropriate upper and lower bounds.
    Thus, PB learns the upper and lower bounds of the output probability vector by minimizing the loss function on the validation dataset.
    Furthermore, we provide a theoretical analysis of PB, revealing the conditions under which it performs well and deriving an upper bound on its calibration error.
    \begin{figure}[t]
        \centering
        \includegraphics[width=0.7\linewidth]{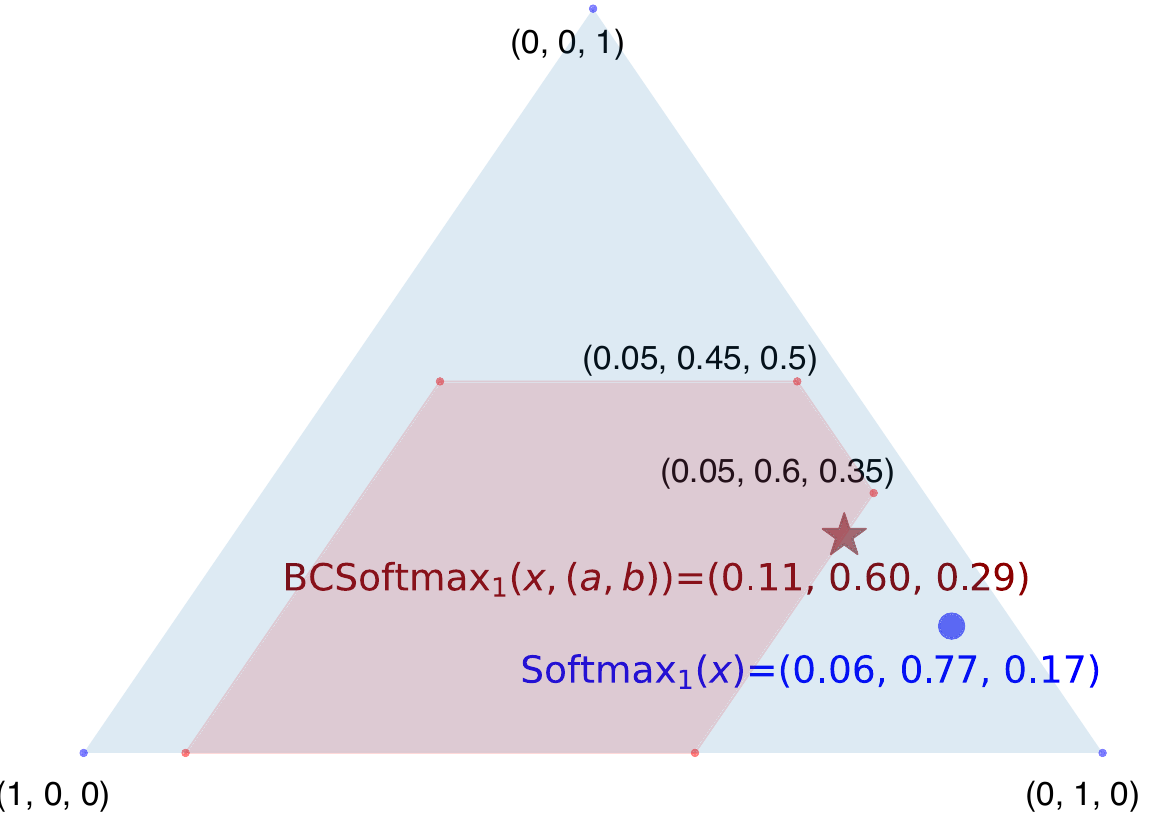}
        \caption{Comparison of the $\Softmax$ probabilities (blue point) with the $\BCSoftmax$ probabilities (red star point) for the logit vector $\bm{g}=(-1.5, 1, -0.5)^\top$, lower bound vector $\bm{a}=(0.05, 0.1, 0.0)^\top$, and upper bound vector $\bm{b} = (1.0, 0.6, 0.5)^\top$ with $\tau=1$. The blue region represents the three-dimensional probability simplex $\Delta^3$ and the red region is the box-constrained $\Delta^3$ induced by $\bm{a}$ and $\bm{b}$, that is, $\Delta^3 \cap [\bm{a}, \bm{b}]$. Due to the upper bound constraint $b_2 = 0.6$, the $\BCSoftmax$ probabilities are pushed into the red region, compared to the $\Softmax$ probabilities.}
        \label{fig:softmax_vs_bcsoftmax}
    \end{figure}
    
    To realize PB, we propose the \textbf{box-constrained softmax} ($\BCSoftmax$) function, an extension of the conventional $\Softmax$ function, by incorporating box constraints on the output probability vector.
    To achieve this in a principled way, we adopt a variational formulation of $\Softmax$, which characterizes it as the solution to a convex optimization problem~\citep{blondel2018smooth,blondel2020learning}.
    Although this formulation could pose computational challenges, we develop an \emph{exact and efficient} algorithm with a correctness guarantee for computing $\BCSoftmax$.
    It runs in expected $O(K)$ time if only upper or lower bounds are applied; otherwise, it runs in $O(K \log K)$ time.
    \Cref{fig:softmax_vs_bcsoftmax} illustrates the comparison between $\Softmax$ and $\BCSoftmax$ in a three-dimensional setting.

    Moreover, we propose other post-hoc calibration methods, \textbf{logit bounding} (LB) and \textbf{dual PB} (DPB), based on an analysis of PB.
    Our experimental results demonstrate that the proposed methods can improve calibration metrics, such as the expected calibration error, for NNs.

    The contributions of this paper are summarized as follows:
    \begin{itemize}
        \item We propose a novel post-hoc calibration method, PB, which mitigates both overconfidence and underconfidence by bounding output probabilities.
        \item We characterize the conditions under which PB performs well and derive an upper bound on its calibration error.
        \item To realize PB, we formulate $\BCSoftmax$ and develop an efficient and exact algorithm for computing it.
        \item We propose two other post-hoc calibration methods based on a mathematical analysis of PB. 
        \item We empirically demonstrate the effectiveness of the proposed methods on four real-world datasets.
    \end{itemize}
    
    \paragraph{Notation.}
    We denote the probability simplex by $\Delta^{K} \coloneqq \{\bm{p} \in \real^K_{\ge 0} \mid \sum_{i=1}^{K} p_i = 1\}$.
    We use $\bm{1}_K$ and $\bm{0}_K$ for the $K$-dimensional all-ones vector and all-zeros vector, respectively.
    For two vectors $\bm{a}, \bm{b} \in \real^m$, $\bm{a}\circ\bm{b} \in \real^m$ is their element-wise product.
    For two vectors $\bm{a} \in \real^m$ and $\bm{b}\in\real^n$, $\bm{a}\concat\bm{b} \in \real^{m+n}$ represents their concatenation.
    For a vector $\bm{a} \in \real^m$ and two integers $i, j \in [m]$, let $\bm{a}_{i:j}$ denote the contiguous subvector of $\bm{a}$ from the $i$-th element to the $j$-th element: $\bm{a}_{i:j} \coloneqq (a_i, a_{i+1}, \ldots, a_{j})^\top$.
    For $i > j$, we define $\bm{a}_{i:j}$ as the empty vector: $\bm{a}_{i:j} \coloneqq ()$ and $()\concat\bm{a}=\bm{a}\concat ()\coloneqq\bm{a}$.
    We use $L^K$ for the set of feasible lower bound vectors of $\Delta^K$, formally, $L^K \coloneqq \{\bm{a} \in \real^K \mid \bm{0}_K \preceq \bm{a} \preceq \bm{1}_K, \sum_{k=1}^{K} a_k \le 1\}$.
    Similarly, we define $U^K$ as the set of feasible upper bound vectors of $\Delta^K$, that is, $U^K \coloneqq \{\bm{b} \in \real^K \mid \bm{0}_K \preceq \bm{b} \preceq \bm{1}_K, \sum_{k=1}^{K} b_k \ge 1\}$.
    Subsequently, we denote the set of feasible box constraints, that is, feasible pairs of lower and upper bound vectors, of $\Delta^K$ by $B^K \coloneqq \{(\bm{u}, \bm{v}) \in L^K \times U^K: \bm{u} \preceq \bm{v}\}$.
    We use $\sigma: \real \to (0,1)$ for the logistic sigmoid function $\sigma(x)=1/(1+e^{-x})$.

\section{Problem Setting}
    \label{sec:calib_setup_def_existing}
    We begin by describing the problem setting, introducing definitions, and reviewing related work.
    
    \paragraph{Classification problem.} Consider a classification problem with $K$ classes.
    We denote $\mathcal{X}$ as the set of input feature vectors and $\mathcal{Y}=[K]$ as the set of output labels.
    We assume that a pair consisting of an input vector $\bm{x} \in \mathcal{X}$ and a label $y\in\mathcal{Y}$ is sampled from an unknown distribution $\mathcal{D}$.
    Let the training data $D_{\mathrm{tr}} = \{(\bm{x}_n, y_n)_{n=1}^{N_{\mathrm{tr}}}\}\sim\mathcal{D}^{N_{\mathrm{tr}}}$.
    Our goal is to train an accurate and well-calibrated (defined later) classification model $f: \mathcal{X} \to \Delta^K$ using $D_{\mathrm{tr}}$.
    Furthermore, assume that we can observe the validation data $D_{\mathrm{val}} \sim \mathcal{D}^{N_{\mathrm{val}}}$ in addition to $D_{\mathrm{tr}}$.
    The validation data are used to evaluate and calibrate models (definitions follow).
    
    \paragraph{Top-label calibration.}
    The model $f$ is said to be \textbf{top-label (argmax) calibrated} if
    \begin{align}
        \mathbb{P}(Y = \hat{y}(X) \mid \hat{p}(X)) = \hat{p}(X) \quad  \text{almost surely},
    \end{align}
    where $\hat{y}(X) = \argmax f(X)$ and $\hat{p}(X) = \max f(X)$.
    The degree of miscalibration is typically measured by the true calibration error ($\TCE$), which is defined as
    \begin{align}
        \TCE_{\mathcal{D}}(f) \coloneqq \mathbb{E}_{X}[\lvert \hat{p}(X) - \mathbb{P}(Y = \hat{y}(X) \mid \hat{p}(X))\rvert].
    \end{align}
    When the model is perfectly calibrated, its $\TCE$ is zero; thus, a lower $\TCE$ indicates a better calibration.
    In practice, $\TCE$ cannot be computed since the distribution $\mathcal{D}$ is unknown.
    As an estimator of $\TCE$, the expected calibration error ($\ECE$) is typically used.
    Given a test dataset $D_{\mathrm{te}} \sim \mathcal{D}^{N_{\mathrm{te}}}$ and a set of intervals (bins) $\mathcal{I}=\{I_m \subset [0,1]: m \in [M]\}$, which is a partition of $[0,1]$, $\ECE$ is defined as
    \begin{align}
        \ECE(f, D_{\mathrm{te}}, \mathcal{I}) \coloneqq \sum_{m=1}^M \frac{|\mathrm{B}_m|}{N_{\mathrm{te}}}\lvert \mathrm{conf}_m- \mathrm{acc}_m\rvert,
    \end{align}
    where $\mathrm{B}_m \coloneqq \{n \in [N_{\mathrm{te}}] \mid \hat{p}(\bm{x}_n) \in I_m\}$ is the set of instances of $D_{\mathrm{te}}$ in bin $m$, $\mathrm{conf}_m \coloneqq \sum_{n \in \mathrm{B}_m} \hat{p}(\bm{x}_n)/|\mathrm{B}_m|$ is the average confidence of bin $m$, and $\mathrm{acc}_m \coloneqq \lvert\{n \in \mathrm{B}_m \mid y_n=\hat{y}(\bm{x}_n)\}\rvert/ \lvert\mathrm{B}_m\rvert$ is the accuracy of bin $m$.
    The partition $\mathcal{I}$ may be defined arbitrarily, but in practice, equal-width ($|I_i| = |I_j|$ $\forall i,j$) or equal-mass ($|\mathrm{B}_i| = |\mathrm{B}_j|$ $\forall i,j$) binning is typically used.
    Some studies have proposed sophisticated estimates of $\TCE$~\citep{blasiok2024smooth,kumar2019verified,zhang2020mix} and alternative metrics of $\TCE$~\citep{blasiok2023does,blasiok2023unifying}.
    
    \paragraph{Post-hoc calibration methods.}
    Several machine learning classification models, especially modern NNs, achieve high classification accuracy but have calibration issues~\citep{guo2017calibration,dutch2024when,wang2021rethinking}.
    Consequently, post-hoc calibration (recalibration) methods have been proposed to calibrate the model using the validation data $D_{\mathrm{val}}$~\citep{gyusang2024tilt,ding2021local,kull2017beta,ma2021meta,platt1999probabilistic,rahimi2020intra,tao2025feature,wenger2020non,zadrozny2002transforming,zhang2020mix}.
    Typically, post-hoc calibration methods introduce a few additional parameters and fit only those parameters.
    Let us assume that the uncalibrated baseline model $f$ is defined as
    \begin{align}
        f(\bm{x}) = \Softmax_{1}(\bm{g}(\bm{x})),
    \end{align}
    where $g: \mathcal{X} \to \real^K$ outputs the logit vector of $\bm{x}$ and
    \begin{align}
         \Softmax_{\tau}(\bm{g}) \coloneqq \frac{\exp(\bm{g}/\tau)}{\sum_{k=1}^K \exp(g_k/\tau)} \quad (\tau > 0).
    \end{align}
    Then, for example, temperature scaling (TS)~\citep{guo2017calibration} introduces the temperature parameter $\tau$ and fits it by minimizing the empirical risk on $D_{\mathrm{val}}$:
    \begin{align}
        \min_{\tau > 0} \hat{R}_{\ell, D_{\mathrm{val}}}(\Softmax_\tau \circ g),
    \end{align}
    where $\hat{R}_{\ell, D}(h) \coloneqq \sum_{(\bm{x},y)\in D}\ell(y, h(\bm{x}))/|D|$ is the empirical $\ell$-risk of $h$ on $D$, $\ell: \mathcal{Y} \times \Delta^K \to \real_{\ge 0} \cup \{\infty\}$ is a loss function such as the cross-entropy (xent) loss $\ell_{\mathrm{xent}}(y, \bm{p})= -\log p_y$.
    Since TS only adds and fits the single scalar parameter $\tau$, it is computationally efficient and preserves model accuracy.
    Therefore, TS is widely used and performs well in various situations.
    Another approach for model calibration is to train a well-calibrated model directly via regularizers or loss functions ~\citep{ghosh2022adafocal,kumar2018trainable,mukhoti2020calibrating,tao2023dual,wang2021rethinking}.
    We mainly consider post-hoc calibration  because post-hoc methods are more widely applicable, easy to use, and empirically perform better than direct methods~\citep{wang2021rethinking}.

\section{Probability Bounding}
    \label{sec:pb}
    In this section, we propose probability bounding (PB), a novel post-hoc calibrator that enforces learned box constraints on output probabilities.
    Enforcing these bounds is reduced to solving an entropy-regularized, box-constrained optimization problem on the probability simplex; we refer to this operator as $\BCSoftmax$, and present its exact and efficient algorithm in~\cref{sec:algorithm_for_bcsoftmax}.
    
    The key idea behind PB is to learn lower and upper bounds on the output probabilities, which can mitigate both underconfidence and overconfidence.
    As described in~\cref{sec:intro}, introducing such constraints can be useful in certain applications that require reliable and trustworthy models.

    \paragraph{Box-constrained softmax.}
    To enforce the box constraints on output probabilities, for a feasible pair of lower and upper bounds $(\bm{a}, \bm{b}) \in B^K$, we define the \textbf{box-constrained softmax} ($\BCSoftmax$) function as
    \begin{align}
        \BCSoftmax_{\tau}(\bm{g}; (\bm{a}, \bm{b})) \coloneqq \argmax_{\bm{p} \in \Delta^K\color{blue}{\cap[\bm{a},\bm{b}]}} \bm{g}^\top\bm{p}+\tau H(\bm{p}),
        \label{eq:bcsoftmax}
    \end{align}
    where $H(\bm{p}) \coloneqq - \sum_{k=1}^K p_k \log p_k$ is the entropy.
    This definition is based on the variational formulation of $\Softmax$~\citep{blondel2018smooth,blondel2020learning,niculae2018sparsemap}:
    \begin{align}
        \Softmax_{\tau}(\bm{g}) = \argmax_{\bm{p} \in \Delta^K} \bm{g}^\top\bm{p}+\tau H(\bm{p}),
        \label{eq:softmax_opt}
    \end{align}
    and obviously $\BCSoftmax_\tau(\cdot; (\bm{0}_K, \bm{1}_K)) = \Softmax_{\tau}$. Thus,
    $\BCSoftmax$ is a natural generalization of $\Softmax$ with box constraints on the outputs.
    
    \paragraph{PB.}
    Then, PB replaces the prediction $f(\bm{x})$ with
    \begin{align}
        f_{\mathrm{PB}}(\bm{x}; a, b) \coloneqq \BCSoftmax_{1}(g(\bm{x}); (a\bm{1}_K, b\bm{1}_K)),
        \label{eq:pb}
    \end{align}
    where $a \in [0,1/K]$ is the lower bound on output probabilities and $b\in[1/K,1]$ is the upper bound on output probabilities.
    Both $a$ and $b$ are fitted by solving the following optimization problem:
    \begin{align}
        \min_{a\in [0, 1/K], b\in[1/K, 1]} \hat{R}_{\ell, D_{\mathrm{val}}}(f_{\mathrm{PB}}(\cdot;a,b)).
        \label{eq:pb_optimization}
    \end{align}
    
    \paragraph{Property of PB.} In PB, all class probabilities are bounded uniformly, i.e., constrained by the same lower and upper bounds.
    This design ensures the following desirable property of the post-hoc calibrated prediction $f_{\mathrm{PB}}$.
    \begin{propositionmd}
        The top-1 prediction(s) of $f(\bm{x})$ are always included in those of $f_{\mathrm{PB}}(\bm{x})$: 
        \begin{align}
            \argmax f(\bm{x}) \subseteq \argmax f_{\mathrm{PB}}(\bm{x})    
        \end{align}
        for all $\bm{x} \in \mathcal{X}$.
        Moreover, if $b > 1/(k+1)$, then PB exactly preserves the top-$k$ predictions.
        \label{prop:pb_preserves_accuracy}
    \end{propositionmd}
    \begin{remark}
        PB can change accuracy because there can be ties among the top-class probabilities in $f_{\mathrm{PB}}(\bm{x})$, even if $f(\bm{x})$ has no ties.
        However, our experimental results in~\cref{sec:experiments} show that such ties are rare in practice.
    \end{remark}
    \begin{remark}
        Because the top-label probability is always larger than the uniform lower bound ($a \le 1/K \le \max f(\bm{x})$ for all $\bm{x}$), the uniform lower bound cannot mitigate the underconfidence of the top-label prediction.
        Instead, it mitigates the overconfidence of the top-label prediction and the underconfidence of the other predictions.
        Non-uniform bounds can mitigate the underconfidence of the top-label prediction, but they no longer preserves the top-1 predictions and model accuracy. 
    \end{remark}

\subsection{Theoretical Analysis of PB}
    We next provide a theoretical analysis of PB.
    Throughout this section, we focus on the binary classification setting, i.e., $\mathcal{Y}=\{0,1\}$, to simplify the discussion.
    We assume that $g$ is a scalar-output function: $g(\bm{x}) \in \real$.
    Then, we redefine $f$ and $f_{\mathrm{PB}}$ as functions that output the probability of $Y=1$, i.e.,
    $f(\bm{x})\coloneqq\Softmax_1((g(\bm{x}), 0)^\top)[1]=\sigma(g(\bm{x}))$ and
    $f_{\mathrm{PB}}(\bm{x}; a, b)\coloneqq\BCSoftmax_1((g(\bm{x}), 0)^\top; (a\bm{1}_2,b\bm{1}_2))[1]$.
    We also redefine $\TCE$ as $\TCE_{\mathcal{D}}(f) \coloneqq \mathbb{E}_{X}[\lvert f(X) - \mathbb{P}(Y=1 \mid f(X))\rvert]$.

\begin{figure}
        \centering
        \includegraphics[width=0.7\linewidth]{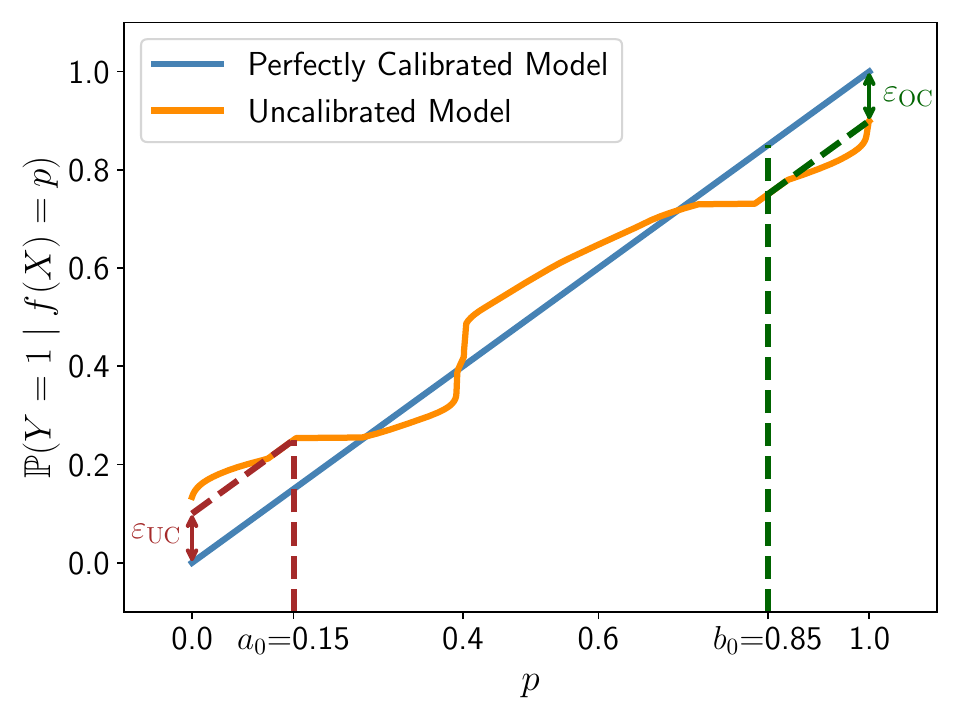}
        \caption{An illustration of the uniform underconfidence (overconfidence) on a low- (high-) probability region property.}
        \label{fig:uniform_uc_oc}
    \end{figure}

    \paragraph{Conditions for PB Effectiveness.}
    We analyze under what conditions the proposed PB can be useful.
    We first define the \emph{uniform underconfidence (overconfidence) on a low-probability (high-probability) region} property, which is illustrated in~\cref{fig:uniform_uc_oc}.
    \begin{definitionmd}
        We say the (measurable) function $f: \mathcal{X} \to [0,1]$ is uniformly underconfident on a low-probability region if there exist $a_0 \in (0,1/2)$ and $\varepsilon_{\mathrm{UC}}>0$ such that
        \begin{align}
             \mathbb{P}[Y=1\mid f(X)] \ge f(X)+\varepsilon_{\mathrm{UC}} \quad \text{almost surely on } \{f(X)\le a_0\}.
        \end{align}
        We also say that $f$ is uniformly overconfident on a high-probability region if there exist $b_0 \in (1/2,1)$ and $\varepsilon_{\mathrm{OC}}>0$ such that 
        \begin{align}
            \mathbb{P}[Y=1\mid f(X)] \le f(X) -\varepsilon_{\mathrm{OC}} \quad \text{almost surely on } \{f(X)\ge b_0\}.
        \end{align}
    \end{definitionmd}
    Then, $\TCE$ of $f_{\mathrm{PB}}$ with appropriate bounds is no greater than that of $f$ (and can be strictly smaller) when $f$ satisfies these properties.
    \begin{theoremmd}
        Assume $\mathcal{Y}=\{0,1\}$ and $f$ is uniformly underconfident on a low-probability region and overconfident on a high-probability region.
        Then, $\exists (a^*, b^*) \in (0,1/2) \times (1/2,1)$ such that for all $a\le a^*$ and $b\ge b^*$,
        \begin{align}
           \mathrm{TCE}_{\mathcal{D}}(f_{\mathrm{PB}}(\cdot; a, b)) \le \mathrm{TCE}_{\mathcal{D}}(f).
            \label{eq:pb_tce_is_not_worse_than_baseline}
        \end{align}
        Moreover,~\cref{eq:pb_tce_is_not_worse_than_baseline} is strict if at least one of the bounds is non-vacuous, i.e., $\mathbb{P}(f(X)<\tilde a)>0$ or $\mathbb{P}(f(X)>\tilde b)>0$, where $\tilde a\coloneqq\max\{a,1-b\}$ and $\tilde{b}\coloneqq 1-\tilde{a}$.
        
        Furthermore, for a strictly proper convex loss $\ell$~\citep{buja2005loss}, PB can also reduce the expected $\ell$-risk under the same conditions, namely,
        \begin{align}
           R_{\ell, \mathcal{D}}(f_{\mathrm{PB}}(\cdot; a, b)) \le R_{\ell, \mathcal{D}}(f) \quad \forall a \le a^*, b\ge b^*,
            \label{eq:pb_risk_is_not_worse_than_baseline}
        \end{align}
        where $R_{\ell, \mathcal{D}}(h) \coloneqq \mathbb{E}_{(X,Y)\sim\mathcal{D}}[\ell(Y, h(X))]$ is the expected $\ell$-risk  of $h$ on $\mathcal{D}$.
        \label{thm:when_does_pb_work}
    \end{theoremmd}
    
\paragraph{Calibration error analysis.}
    Next, we establish an upper bound on the smooth calibration error~\citep{blasiok2023does,blasiok2023unifying} of $f_{\mathrm{PB}}$.
    The smooth calibration error ($\smCE$) and the true distance to calibration ($\dCE$) of $f$ on $\mathcal{D}$, an alternative metric of $\TCE$, are defined as
    \begin{align}
        \smCE_{\mathcal{D}}(f) &\coloneqq \sup_{\eta \in \mathcal{H}} \mathbb{E}_{(X,Y)}[\eta(f(X))(Y - f(X))], 
        \label{eq:smCE}
        \\
        \dCE_{\mathcal{D}}(f) &\coloneqq \inf_{f' \in \mathsf{cal}(\mathcal{D})} \mathbb{E}_{X}[\lvert f(X) - f'(X)\rvert], \label{eq:dCE}
    \end{align}
    where $\mathcal{H}$ is the set of all $1$-Lipschitz functions $\eta: [0,1] \to [-1,1]$ and $\mathsf{cal}(\mathcal{D})$ is the set of all perfectly-calibrated models.
    Because $\smCE$ is consistent with $\dCE$~\citep{blasiok2023unifying}, its upper bound guarantees good calibration.
    \begin{theoremmd}
        Assume $\mathcal{Y}=\{0,1\}$.
        Given $g$ and $D_{\mathrm{val}}$, let $(\hat{a}, \hat{b})$ be the solution to~\cref{eq:pb_optimization} with the xent loss: $\ell(y, p)=-y\log p -(1-y)\log (1-p)$ and let $\mathcal{K}$ be the family of all functions $\kappa:\real \to \real$ such that $\eta(v)=\kappa(v)-v$ is 1-Lipschitz and bounded $\lvert\eta(v)\rvert\le 4$ for all $v \in \real$.
        Then, for any $\delta \in (0,1)$, with probability at least $1-\delta$, 
        \begin{align}
            \begin{split}
            \smCE_{\mathcal{D}}(\hat{f}_{\mathrm{PB}})^2
            &\le \frac{1}{2}\left(\hat{R}_{\ell, D_{\mathrm{val}}}(\hat{f}_{\mathrm{PB}})-\inf_{\kappa \in \mathcal{K}}\hat{R}_{\ell,D_{\mathrm{val}}}(\hat{f}_{\mathrm{PB},\kappa})\right) \\
            &\quad +\frac{1}{\sqrt{N_{\mathrm{val}}}}\left(42B_{\hat{k}} + 144 + \left(B_{\hat{k}}+2\right) \sqrt{\frac{\log (2^{\hat{k}+2}/\delta)}{2}} \right),
            \label{eq:smooth_ce_bound} 
            \end{split}
        \end{align}
        where $\hat{f}_{\mathrm{PB}}=\hat{f}_{\mathrm{PB}}(\cdot; \hat{a},\hat{b})$, $\hat{f}_{\mathrm{PB},\kappa} \coloneqq \sigma\circ\kappa\circ\sigma^{-1}\circ \hat{f}_{\mathrm{PB}}$, $B_k \coloneqq \sigma^{-1}(1-2^{-k})$, and $\hat{k} \coloneqq \min\{k \in \mathbb{N}_{>0}:\max(\hat{a}, 1-\hat{b})\ge 2^{-k}\}$.
        \label{thm:smooth_ce_bound_pb}
    \end{theoremmd}
    The first term in~\cref{eq:smooth_ce_bound} is the empirical post-processing gap~\citep{blasiok2023does}.
    The second term depends on the trained bounds through $B_{\hat{k}}\ge 0$ and decreases as the bounds become tighter.

\subsection{Exact and Efficient Algorithm for \texorpdfstring{$\BCSoftmax$}{BCSoftmax}}
    \label{sec:algorithm_for_bcsoftmax}
    Evaluation of $\BCSoftmax$ in~\cref{eq:bcsoftmax} requires solving a box-constrained convex optimization problem, which can be computationally costly.
    Moreover, when $\BCSoftmax$ is used in machine learning models trained by gradient-based optimization methods, we also need its Jacobian.
    Recently, differentiation methods for optimization problems have attracted attention and have been actively developed~\citep{agrawal2019differentiable,amos2017optnet,blondel2022efficient,bolte2023one}, but they remain computationally expensive.
    \citet{martins2017learning} and~\citet{parra2025deep} proposed special cases of $\BCSoftmax$: with only upper- and lower-bound constraints, respectively, based on~\cref{eq:softmax_opt}.
    Unfortunately, theoretical guarantees for their algorithms are missing, and one of them can yield incorrect solutions\footnote{We show a counterexample in~\cref{sec:counterexample_for_lb}.}.
    Therefore, we need to develop an efficient and exact algorithm for $\BCSoftmax$.

\subsubsection{Upper-Bounded Softmax}
    \label{sec:ubsoftmax}
    For simplicity, we first derive an efficient and exact algorithm for computing $\BCSoftmax$ with only upper-bound constraints.
    The upper-bounded softmax ($\UBSoftmax$) function, which is called constrained softmax in~\citet{martins2017learning}, is defined as
    \begin{align}
        \UBSoftmax_{\tau}(\bm{g}, \bm{b}) \coloneqq \BCSoftmax_{\tau}(\bm{g}; (\bm{0}_K, \bm{b})).
    \end{align}
    To derive an exact algorithm for computing $\UBSoftmax$, we present a theorem that states a relationship between $\UBSoftmax$ and $\Softmax$.
    \begin{theoremmd}[Proposition 1 in~\citet{martins2017learning}]
        For all $\tau>0$, $\bm{g} \in \real^K$, and $\bm{b} \in U^K$, there exists a nonnegative vector $\bm{\beta} \in \real^K_{\ge 0}$ such that
        \begin{align}
            \UBSoftmax_{\tau}(\bm{g}, \bm{b})[i] &= \Softmax_{\tau}(\bm{g}-\bm{\beta})[i]
            \label{eq:relationship_ub_soft}\\
            &=
            \begin{cases}
                b_i, & \beta_i > 0 \\
                {\exp(g_i / \tau)}/{Z}, & \beta_i = 0
            \end{cases}
            \label{eq:ub_beta_indices}
        \end{align}
        where $Z \coloneqq r/s,$ $r \coloneqq \sum_{\beta_i =0}e^{g_i/\tau}$, and $s \coloneqq 1 - \sum_{\beta_i > 0}b_i$.
        \label{thm:relationship_ub_soft}
    \end{theoremmd}
    By~\cref{eq:relationship_ub_soft}, $\UBSoftmax$ can be evaluated exactly if we can obtain the nonnegative vector $\bm{\beta}$.
    Furthermore,~\cref{eq:ub_beta_indices} indicates that for the evaluation of $\UBSoftmax$, it is unnecessary to compute $\bm{\beta}$ itself, but it is sufficient to identify the indices of nonzero elements of $\bm{\beta}$.
    
    How do we compute the indices of nonzero elements of $\bm{\beta}$?
    The following theorem, which is not proven in~\citet{martins2017learning}, states that the indices of the nonzero elements of $\bm{\beta}$ can be found by basic calculations when the vectors $\bm{g}$ and $\bm{b}$ are sorted by $b_i / \exp(g_i/\tau)$.
    \begin{theoremmd}
        For all $\tau>0$, $\bm{g} \in \real^K$, and $\bm{b} \in U^K$ such that $b_1/e^{g_1/\tau} \le \cdots \le b_K / e^{g_K/\tau}$, we have
        \begin{align}
            \UBSoftmax_{\tau}(\bm{g}, \bm{b}) = \bm{b}_{1:\rho}\concat s_\rho \cdot \Softmax_{\tau}(\bm{g}_{\rho+1:K}),
            \label{eq:ub_sorted}
        \end{align}
        where $\rho \coloneqq \min \{k \in \{0, \ldots, K-1\} \mid e^{ g_{k+1}/\tau}/Z_k \le b_{k+1}\}$, $Z_k \coloneqq r_k/s_k$, $r_k \coloneqq \sum_{i=k+1}^{K} e^{g_i/\tau}$, $
        s_k \coloneqq 1-\sum_{i=1}^{k} b_{i}$.
        \label{thm:relathinship_ub_soft_sorted}
    \end{theoremmd}
    Thus, $\UBSoftmax$ can be evaluated in $O(K \log K)$ time by~\cref{alg:ubsoftmax}.
    Furthermore, it can be improved to $O(K)$ time based on a quickselect-like procedure~\citep{cormen2022introduction,duchi2008efficient} (please see~\cref{sec:ubsoftmax_linear}).

    \begin{algorithm}[t]
        \caption{$O(K \log K)$ algorithm for computing $\UBSoftmax_\tau$}
        \label{alg:ubsoftmax}
        \begin{algorithmic}[1]
            \Input $\bm{g} \in \real^K, \bm{b} \in U^K$
            \State $\bm{g} \gets \bm{g}/\tau$
            \State Sort $\bm{g}$ and $\bm{b}$ as $b_{1}/e^{g_{1}} \le \cdots \le b_{K}/e^{g_{K}}$ 
            \State $r_k \gets \sum_{i=k+1}^{K} e^{g_i}$, $s_k \gets 1-\sum_{i=1}^{k} b_{i}$, and $Z_k \gets r_k/s_k$ for all $k \in [K]$
            \State $\rho \gets \min \{k \in \{0, \ldots, K-1\} \mid e^{g_{k+1}} / Z_k \le b_{k+1}\}$
            \State $p_i \gets b_i \ \forall i \le \rho$ and $p_i \gets e^{g_i} / Z_\rho$ $\forall i > \rho$ 
            \State Undo sorting $\bm{p}$
            \Output $\bm{p}$
        \end{algorithmic}
    \end{algorithm}

\subsubsection{From \texorpdfstring{$\UBSoftmax$}{UBSoftmax} to \texorpdfstring{$\BCSoftmax$}{BCSoftmax}}
    \label{sec:from_ub_to_bc}
    We next derive an efficient and exact algorithm for computing $\BCSoftmax$.
    First, we extend~\cref{thm:relationship_ub_soft} and~\cref{thm:relathinship_ub_soft_sorted} for $\BCSoftmax$.
    \begin{theoremmd}
        For all $\tau>0$, $\bm{g} \in \real^K$, and $(\bm{a}, \bm{b}) \in B^K$, there exists a vector $\bm{\gamma} \in \real^K$ such that
        \begin{align}
            \BCSoftmax_{\tau}(\bm{g}; (\bm{a}, \bm{b}))[i] &= \Softmax_{\tau}(\bm{g}-\bm{\gamma})[i]
            \label{eq:relationship_bc_soft}\\
            &=
            \begin{cases}
                a_i & \gamma_i < 0 \\
                b_i & \gamma_i > 0 \\
                {\exp(g_i / \tau)}/{Z} & \gamma_i = 0
            \end{cases},
            \label{eq:bc_gamma_indices}
        \end{align}
        where $Z \coloneqq r/s$, $r \coloneqq \sum_{\gamma_i = 0}e^{g_i/\tau}$, and $s \coloneqq 1 - \sum_{\gamma_i < 0}a_i - \sum_{\gamma_i > 0}b_i$.
        \label{thm:relationship_bc_soft}
    \end{theoremmd}
    \begin{theoremmd}
        For all $\tau>0$, $\bm{g} \in \real^K$, and $(\bm{a}, \bm{b}) \in B^K$ such that $a_1/e^{g_1/\tau} \ge \cdots \ge a_K / e^{g_K/\tau}$, we have
        \begin{align}
            \BCSoftmax_{\tau}(\bm{g}; (\bm{a}, \bm{b})) = p(\rho),
            \label{eq:bc_sorted}
        \end{align}
        where $\rho \coloneqq \min \{k \in \{0, \ldots, K^\prime\} \mid \bm{a} \preceq p(k) \preceq \bm{b}\}$, $K^\prime\coloneqq \max \{k \mid \sum_{i>k}b_i \ge s_k\}$, $s_k \coloneqq 1-\sum_{i=1}^{k} a_{i}$, and
        \begin{align}
            p(k) \coloneqq \bm{a}_{1:k}\concat s_k \cdot \UBSoftmax_{\tau}(\bm{g}_{k+1:K}, \bm{b}_{k+1:K}/s_k).
            \label{eq:p_of_k}
        \end{align}
        \label{thm:relathinship_bc_soft_sorted}
    \end{theoremmd}
    Based on~\cref{thm:relathinship_bc_soft_sorted}, we introduce~\cref{alg:bcsoftmax}, which computes $\BCSoftmax$ exactly in $O(K \log K)$ time\footnote{In~\cref{alg:bcsoftmax}, we assume that the algorithm for computing $\UBSoftmax$ runs in $O(K)$ time.
    As described in the previous section, it is achieved by a quickselect-like procedure.}.

    \begin{algorithm}[t]
        \caption{$O(K \log K)$ algorithm for computing $\BCSoftmax_\tau$}
        \label{alg:bcsoftmax}
        \begin{algorithmic}[1]
            \Input $\bm{g} \in \real^K, (\bm{a}, \bm{b}) \in B^K$
            \State $\bm{g} \gets \bm{g}/\tau$
            \State Sort $\bm{g}$, $\bm{a}$, and $\bm{b}$ as $a_{1}/e^{g_{1}} \ge \cdots \ge a_{K}/e^{g_{K}}$
            \State $L \gets 0$, and $R \gets K^\prime=\max \{k:\sum_{i>k}b_i \ge s_k\}$
            \While{$L < R$}
                \State $\rho \gets \lfloor(L+R) / 2\rfloor, s_\rho \gets 1-\sum_{i=1}^{\rho} a_i$
                \State $\bm{p} \gets p(\rho)$ in~\cref{thm:relathinship_bc_soft_sorted} (\cref{eq:p_of_k})
                \State $R \gets \rho$ if $\bm{a} \preceq \bm{p} \preceq \bm{b}$; otherwise $L\gets\rho+1$
            \EndWhile
            \State $\rho \gets \lfloor(L+R) / 2\rfloor, s_\rho \gets 1-\sum_{i=1}^{\rho} a_i$
            \State $\bm{p} \gets p(\rho)$ 
            \State Undo sorting $\bm{p}$
            \Output $\bm{p}$
        \end{algorithmic}
    \end{algorithm}
 
    \Cref{thm:relationship_bc_soft}, especially~\cref{eq:bc_gamma_indices}, states that each output probability of the $\BCSoftmax$ function is its lower bound, upper bound, or softmax-like probability.
    Some basic properties of $\BCSoftmax$ and its Jacobians are derived in~\cref{sec:basic_properties}.
    Moreover, our experimental results demonstrate that the proposed algorithm for $\BCSoftmax$ is 150--400$\times$ faster than the existing general algorithm~\citep{agrawal2019differentiable} (\cref{sec:runtime_comparison}).

\subsection{Extension of PB}
    \label{sec:extension_of_pb}
    \paragraph{Instance-dependent bounds.}
    While~\cref{eq:pb} uses constant bounds, it can be beneficial to use instance-dependent bounds when prediction uncertainty varies across inputs.
    For such cases, it may be effective to parameterize the bounds as $a(\bm{x}; \Theta_a)$ and $b(\bm{x}; \Theta_b)$, e.g., using neural networks, and learn $\Theta_a$ and $\Theta_b$ on the validation dataset.
    
    \paragraph{Combination of PB and existing methods.}
    Since many post-hoc calibrators rely on $\Softmax$~\citep{ding2021local,kull2019beyond,mozafari2018attended,rahimi2020intra,zhang2020mix}, PB can be combined with them by replacing $\Softmax$ with $\BCSoftmax$.
    For example, the combination of PB and TS is defined as
    \begin{align}
        f_{\mathrm{TSPB}}(\bm{x}; \tau, a,b) = \BCSoftmax_{\tau}(g(\bm{x}); (a\bm{1}_K,b\bm{1}_K)),
    \end{align}
    and the parameters $\tau$, $a$, and $b$ are fit by minimizing the empirical risk on $D_{\mathrm{val}}$.
    
\section{Logit Bounding and Dual PB}
    \label{sec:lb_and_dpb}
    In this section, we propose two additional post-hoc calibration methods based on PB.
\subsection{Logit Bounding}
    \label{sec:lb}
    We introduce \textbf{logit bounding} (LB), which can be a computationally simpler surrogate for PB.
    Our idea, which is based on~\cref{thm:relationship_bc_soft}, is that we can bound the probabilities by learning $\bm{\gamma}$ instead of $\bm{a}$ and $\bm{b}$.
    The following theorem states that, for scalar lower and upper bounds of probabilities, there exist scalar lower and upper bounds of logits.
    \begin{propositionmd}
        Given $\tau>0$, $\bm{g} \in \real^K$, and $(a\bm{1}_K, b\bm{1}_K) \in B^K$, there exist two scalars: a lower bound on logits $c \in \real$ and an upper bound on logits $C \in \real$ such that $c \le C$ and
        \begin{align}
            \BCSoftmax_{\tau}(\bm{g}; (a\bm{1}_K, b\bm{1}_K)) &= \Softmax_{\tau}(\tilde{\bm{g}}),
            \label{eq:relationship_bc_soft_scalar}
        \end{align}
        where $\tilde{\bm{g}} \coloneqq \clip(\bm{g}, c, C)=\max(c \bm{1}_K, \min(\bm{g}, C \bm{1}_K))$, and $\max(\cdot,\cdot)$ and $\min(\cdot, \cdot)$ denote the element-wise maximum and minimum, respectively.
        \label{thm:relationship_bc_soft_scalar}
    \end{propositionmd}
    Therefore, our LB modifies the prediction function as
    \begin{align}
        f_{\mathrm{LB}}(\bm{x}; c, C) \coloneqq \Softmax_{1}(\clip(g(\bm{x}), c, C)).
        \label{eq:lb}
    \end{align}
    As with $a$ and $b$ in PB, $c$ and $C$ are fit by empirical risk minimization on the validation dataset.
    Note that even if $a$ and $b$ are constant, $c$ and $C$ in~\cref{thm:relationship_bc_soft_scalar} are not constant; they depend on $a$, $b$, and $\bm{g}$.
    In this sense, it is reasonable to define $c$ and $C$ as functions of $g(\bm{x})$.
    The advantage of LB compared to PB is its simplicity of implementation.
    \cref{eq:lb} can be evaluated in $O(K)$ because it consists of canonical $\Softmax$ and element-wise $\max$ and $\min$.

\subsection{Relationship between PB/LB and FC}
    \label{sec:relationship_fc}
    As a similar method to LB,~\citet{tao2025feature} proposed feature clipping (FC) for post-hoc calibration of NNs.
    Assume that $f$ is an NN and the logit function is defined as $g(\bm{x}) = \bm{W}_{\mathrm{linear}}z(\bm{x})+\bm{w}_{\mathrm{bias}}$, where $z(\bm{x})\in\real^D$ is the output of the penultimate layer and $\bm{W}_{\mathrm{linear}} \in \real^{K \times D}, \bm{w}_{\mathrm{bias}} \in \real^{K}$ are the parameters of the output layer.
    Then, FC clips $z(\bm{x})$ as $\clip(\bm{z}(\bm{x}), -c, c)$ while LB clips $g(\bm{x})$, that is, 
    \begin{align}
        f_{\mathrm{FC}}(\bm{x}; c) \coloneqq \Softmax_1(\bm{W}_{\mathrm{linear}}\clip(z(\bm{x}), -c, c)+\bm{w}_{\mathrm{bias}}).
    \end{align}
    
    The next proposition establishes that FC admits the representation in~\cref{eq:relationship_bc_soft} with instance-dependent $\bm{\gamma}$ and LB does so with instance-dependent and potentially sparse $\bm{\gamma}$.
    \begin{propositionmd}
        Given $\bm{z}\in\real^D, \bm{W} \in \real^{K\times D}, \bm{w}\in\real^K$ and $c>0$, let $\bm{g} \coloneqq \bm{W}\bm{z}+\bm{w}$, $\bm{g}_{\mathrm{FC}} \coloneqq \bm{W}\clip(\bm{z}, -c, c)+\bm{w}$, and $\bm{g}_{\mathrm{LB}} \coloneqq \clip(\bm{g}, -c, c)$.
        Then, 
        \begin{align}
            \bm{g}_{\mathrm{FC}} = \bm{g} - \bm{\gamma}_{\mathrm{FC}}, \quad \bm{g}_{\mathrm{LB}} = \bm{g} - \bm{\gamma}_{\mathrm{LB}},
        \end{align}
        where $\bm{\gamma}_{\mathrm{FC}} \coloneqq \bm{W}S_{c}(\bm{z)}$, $\bm{\gamma}_{\mathrm{LB}} \coloneqq S_{c}(\bm{g})$, and $S_{c}: \bm{u} \mapsto\argmin_{\bm{v}} \lVert \bm{v}-\bm{u}\rVert_2^2/2 + c\lVert \bm{v}\rVert_1$ is the soft-thresholding operator~\citep{beck2009fast}:
        \begin{align}
            S_c(\bm{u})[i] = \begin{cases}
                u_i - c &\ \ \ c < u_i, \\
                0 & -c < u_i < c, \\
                u_i + c &\ \ \ \ \ \ \ \ \ \ u_i < -c.
            \end{cases}
        \end{align}
        \label{prop:lb_and_fc}
    \end{propositionmd}
    
\subsection{Dual Probability Bounding}
    \label{sec:dpb}
    Based on~\cref{thm:relationship_bc_soft} and~\cref{prop:lb_and_fc}, we propose \textbf{dual PB} (\textbf{DPB}), which learns the dual variable $\bm{\gamma}$ more directly:
    \begin{align}
        f_{\mathrm{DPB}}(\bm{x}; \bm{w}, d) \coloneqq \Softmax_{1}(g(\bm{x}) - \bm{\gamma}), \quad \bm{\gamma} \coloneqq S_d(\bm{w}\circ g(\bm{x})),
        \label{eq:dpb}
    \end{align}
    where $\bm{w} \in \real_{\ge 0}^{K}$ and $d > 0$ are parameters to be optimized.
    In DPB, $\bm{\gamma}$ can be sparse because $S_d$ is applied to $\bm{w}\circ g(\bm{x})$ and $\bm{w}$ is optimized, whereas $S_c$ is applied to $\bm{z}$ and $\bm{W}$ is fixed in FC.
    From the perspective of its correspondence with PB, it is desirable for $\bm{\gamma}$ to be potentially sparse: \cref{thm:relationship_bc_soft} states that $\gamma_k = 0$ if the $k$-th probability is not bounded.
    To prevent overfitting, we define $\bm{w}$ as a nonnegative vector rather than as a real-valued matrix~\citep{guo2017calibration,kull2019beyond}.

\section{Related Work}
\label{sec:related_work}
\subsection{Extensions of \texorpdfstring{$\Softmax$}{Softmax}}
    Based on the variational formulation of $\Softmax$ in~\cref{eq:softmax_opt}, several researchers proposed sparse extensions of $\Softmax$.
    \citet{martins2016softmax} proposed $\Sparsemax$ by employing a squared $L_2$ norm instead of entropy:
    \begin{align}
        \Sparsemax_{\tau}(\bm{x}) \coloneqq \argmax_{\bm{y} \in \Delta^K} \bm{x}^\top \bm{y} - \frac{\tau}{2}\lVert \bm{y} \rVert_2^2 = \argmin_{\bm{y} \in \Delta^K} \lVert \bm{x}/\tau - \bm{y} \rVert_2^2.
    \end{align}
    Although the outputs of $\Softmax$ are always dense, \citet{laha2018controllable} introduced $\Sparsegen$, which is a generalization of $\Softmax$ and $\Sparsemax$:
    \begin{align}
        \Sparsegen_\tau(\bm{x}; g, \lambda) \coloneqq \argmin_{\bm{y} \in \Delta^K} \lVert \bm{y} - g(\bm{x}/\tau) \rVert_2^2 - \lambda \lVert \bm{y} \rVert_2^2,
    \end{align}
    where $\lambda <1$ and $g: \real^K \to \real^K$ is a component-wise transformation function.
    \citet{balazy2023r} proposed a sparse extension of the $\Softmax$ function, $r$-$\Softmax$, using an approach distinct from $\Sparsemax$ and $\Sparsegen$.
    First, they defined the weighted softmax function, $\mathrm{w\text{-}Softmax}$, as 
    \begin{align}
        \mathrm{w\text{-}Softmax}_{\tau}(\bm{x}, \bm{w})[i] \coloneqq \frac{w_i \cdot \exp (x_i / \tau)}{\sum_{k=1}^{K}w_k \cdot \exp (x_k / \tau)},
    \end{align}
    where $\bm{w} \in \{\bm{u} \in \real^K \mid \bm{0}_K \preceq \bm{u}, \sum_k u_k > 0\}$ is a weight vector.
    The $i$-th probability of $\mathrm{w\text{-}Softmax}$ is zero when $w_i = 0$, thus $\mathrm{w\text{-}Softmax}$ can produce sparse probabilities.
    Subsequently, they proposed $\mathrm{t\text{-}Softmax}$ as $\mathrm{t\text{-}Softmax}_{\tau}(\bm{x}, t > 0) \coloneqq \mathrm{w\text{-}Softmax}_{\tau}(\bm{x}, w(t, \bm{x}))$, where $w(t, \bm{x})[i] \coloneqq \max(0, x_i + t - \max(\bm{x}))$.
    Based on $\mathrm{t\text{-}Softmax}$, they proposed $\mathrm{r\text{-}Softmax}$ as $\mathrm{r\text{-}Softmax}(\bm{x}, r\in [0,1]) \coloneqq \mathrm{t\text{-}Softmax}_{\tau}(\bm{x}, -\mathrm{quantile}(\bm{x}, r) + \max(\bm{x}))$.
    The parameter $r$ represents the sparsity rate; $\mathrm{r\text{-}Softmax}$ with $r=k/K$ outputs a probability vector with $k$ zero values.
    \citet{wang2024softmax} proposed $\epsilon$-$\Softmax$ to learn an accurate classifier from noisy labels.
    Their proposed $\epsilon$-$\Softmax$ outputs probability vectors that approximate one-hot vectors.

    As described in~\cref{sec:algorithm_for_bcsoftmax}, \citet{martins2017learning} and~\citet{parra2025deep} proposed an extension of $\Softmax$ with upper and lower bound constraints, respectively, based on~\cref{eq:softmax_opt}.
    Our $\BCSoftmax$ generalizes these constrained variants.
    
\subsection{Controlling Output Probabilities of \texorpdfstring{$\Softmax$}{Softmax}}
    The temperature $\tau$ softly controls the output of $\Softmax$ and plays a crucial role not only in probability calibration~\citep{guo2017calibration} but also in several applications.
    Since $\BCSoftmax$ enables us to control the output probabilities via hard constraints, it is also useful for such applications.

    In reinforcement learning, when a policy function employs the $\Softmax$ function, a trade-off between exploration and exploitation can be balanced by tuning $\tau$.
    A higher $\tau$ yields a more exploratory policy, while a lower $\tau$ results in a more exploitative policy function.
    However, determining an appropriate value remains a challenge~\citep{sutton2018reinforcement}.
    \citet{he2018determining} proposed a metric for evaluating the effectiveness of $\tau$ and a procedure for selecting $\tau$ based on it.
    
    Recently, large language models (LLMs)  have achieved remarkable performance in various natural language processing tasks.
    Fundamentally, LLMs primarily solve the next-token prediction problem by using $\Softmax$.
    \citet{peeperkorn-etal-2024} demonstrated that the trade-off between novelty and coherence in generated text can be controlled by adjusting $\tau$.
    \citet{renze2024effect} found that the performance of LLMs began to drop rapidly when $\tau > 1$ and the generated text became incoherent when $\tau=1.6$.
    
\section{Experiments}
    \label{sec:experiments}
    \begin{table}[t]
        \centering
        \caption{Summary of datasets.}
        \begin{tabular}{cccccc}
            \multirow{2}{*}{Dataset}&\multirow{2}{*}{Input space $\mathcal{X}$} & \multirow{2}{*}{$K$} & \multicolumn{3}{c}{The number of samples}\\
            &&&$N_{\mathrm{tr}}$ & $N_{\mathrm{val}}$ & $N_{\mathrm{te}}$\\\midrule
            TImageNet&3$\times$64$\times$64 images&200&90,000&10,000&10,000\\\midrule
            CIFAR-100&3$\times$32$\times$32 images&100&45,000&5,000&10,000\\\midrule
            \multirow{2}{*}{20News}&1K-word texts with&\multirow{2}{*}{20}&\multirow{2}{*}{10,182}&\multirow{2}{*}{1,132}&\multirow{2}{*}{7,532}\\
            &20K-word vocab&&\\\midrule
            ImageNet-1K&3$\times$224$\times$224 images&1,000&N/A&10,000&40,000
        \end{tabular}
        \label{tab:datasets}
    \end{table}

    \begin{table*}[t]
        \caption{Comparison of Error Rate, smECE, emECE, and ewECE (\%, lower is better) among the baseline (\textbf{Uncal}) and the existing methods (\textbf{TS}, \textbf{IBTS/PTS}, \textbf{ETS}, \textbf{Dir}, \textbf{NC}, and \textbf{T\&A}) without (Base) and with the proposed methods (+\textbf{PB} or \textbf{DPB}). We report average scores and standard errors (in round brackets) over 10 runs with different random seeds. The symbols \greendown\, and \redup\, indicate improvement and degradation, respectively.}
        \centering
        \tiny
        \begin{tabular}{lcgcgcgcgcgcgcgcg}
            \multirow{2}{*}{{\footnotesize \textbf{Metric}}}&\multicolumn{2}{c}{{\footnotesize \textbf{Uncal}}}&\multicolumn{2}{c}{{\footnotesize \textbf{TS}}}&\multicolumn{2}{c}{{\footnotesize \textbf{IBTS/PTS}}}&\multicolumn{2}{c}{{\footnotesize \textbf{ETS}}}&\multicolumn{2}{c}{{\footnotesize \textbf{Dir}}}&\multicolumn{2}{c}{{\footnotesize \textbf{NC}}}&\multicolumn{2}{c}{{\footnotesize \textbf{T\&A}}}&{{\footnotesize \textbf{FC}}}&{\footnotesize \textbf{DPB}}\\
            &{\scriptsize Base}&{\scriptsize +\textbf{PB}}&{\scriptsize Base}&{\scriptsize +\textbf{PB}}&{\scriptsize Base}&{\scriptsize +\textbf{PB}}&{\scriptsize Base}&{\scriptsize +\textbf{PB}}&{\scriptsize Base}&{\scriptsize +\textbf{PB}}&{\scriptsize Base}&{\scriptsize +\textbf{PB}}&{\scriptsize Base}&{\scriptsize +\textbf{PB}}&{}&{}\\\midrule
            \multicolumn{16}{c}{{\footnotesize Dataset: \textbf{TImageNet} \quad Model: \textbf{ResNet-50}}}\\\midrule
            \multirow{2}{*}{{\scriptsize Error Rate$\downarrow$}}&38.92&38.92&38.92&38.92&38.92&38.92&38.92&38.92&40.15&40.08\greendown&39.05&39.03\greendown&39.66&39.67\redup&41.69&39.15\greendown\\
            &(0.09)&(0.09)&(0.09)&(0.09)&(0.09)&(0.09)&(0.09)&(0.09)&(0.08)&(0.16)&(0.08)&(0.10)&(0.11)&(0.10)&(0.11)&(0.11)\\
            \multirow{2}{*}{{\scriptsize smECE$\downarrow$}}&24.26&7.01\greendown&2.58&2.36\greendown&1.54&\textbf{1.45}\greendown&2.30&2.18\greendown&3.40&2.99\greendown&2.75&2.55\greendown&2.90&2.63\greendown&3.27&2.57\greendown\\
            &(0.13)&(0.09)&(0.05)&(0.04)&(0.11)&(0.03)&(0.05)&(0.04)&(0.09)&(0.09)&(0.07)&(0.05)&(0.07)&(0.10)&(0.17)&(0.06)\\
            \multirow{2}{*}{{\scriptsize emECE$\downarrow$}}&21.50&10.16\greendown&2.55&2.41\greendown&1.40&\textbf{1.30}\greendown&2.28&2.23\greendown&3.37&3.29\greendown&2.74&2.58\greendown&2.91&2.71\greendown&3.30&2.53\greendown\\
            &(0.11)&(0.12)&(0.06)&(0.07)&(0.12)&(0.07)&(0.06)&(0.07)&(0.09)&(0.07)&(0.07)&(0.06)&(0.09)&(0.09)&(0.17)&(0.06)\\
            \multirow{2}{*}{{\scriptsize ewECE$\downarrow$}}&21.51&8.56\greendown&2.60&2.38\greendown&1.45&\textbf{1.37}\greendown&2.37&2.28\greendown&3.49&3.12\greendown&2.79&2.66\greendown&2.99&2.75\greendown&3.28&2.59\greendown\\
            &(0.11)&(0.23)&(0.05)&(0.05)&(0.13)&(0.07)&(0.06)&(0.05)&(0.10)&(0.09)&(0.08)&(0.06)&(0.06)&(0.10)&(0.17)&(0.07)\\\midrule
            \multicolumn{16}{c}{{\footnotesize Dataset: \textbf{CIFAR-100} \quad Model: \textbf{DenseNet-121}}}\\\midrule
            \multirow{2}{*}{{\scriptsize Error Rate$\downarrow$}}&22.96&22.96&22.96&22.96&22.96&22.96&22.96&22.96&22.86&22.89\redup&23.33&23.32\greendown&23.36&23.33\greendown&23.63&23.07\greendown\\
            &(0.11)&(0.11)&(0.11)&(0.11)&(0.11)&(0.11)&(0.11)&(0.11)&(0.11)&(0.11)&(0.08)&(0.09)&(0.08)&(0.09)&(0.10)&(0.10)\\
            \multirow{2}{*}{{\scriptsize smECE$\downarrow$}}&8.76&5.60\greendown&1.59&1.53\greendown&1.39&1.49\redup&1.64&1.53\greendown&1.80&1.71\greendown&2.19&1.94\greendown&3.40&2.63\greendown&4.43&\textbf{1.32}\greendown\\
            &(0.12)&(0.14)&(0.05)&(0.06)&(0.06)&(0.06)&(0.07)&(0.06)&(0.05)&(0.07)&(0.07)&(0.10)&(0.27)&(0.17)&(0.11)&(0.06)\\
            \multirow{2}{*}{{\scriptsize emECE$\downarrow$}}&7.84&6.60\greendown&1.35&1.46\redup&1.27&1.36\redup&1.48&1.45\greendown&1.59&1.79\redup&1.98&2.07\redup&3.17&3.05\greendown&4.29&\textbf{1.04}\greendown\\
            &(0.10)&(0.09)&(0.04)&(0.06)&(0.08)&(0.08)&(0.08)&(0.06)&(0.07)&(0.08)&(0.09)&(0.14)&(0.27)&(0.20)&(0.11)&(0.07)\\
            \multirow{2}{*}{{\scriptsize ewECE$\downarrow$}}&7.84&5.98\greendown&1.51&1.50\greendown&1.28&1.40\redup&1.62&1.50\greendown&1.71&1.61\greendown&2.06&1.95\greendown&3.24&2.57\greendown&4.34&\textbf{1.18}\greendown\\
            &(0.10)&(0.11)&(0.08)&(0.10)&(0.06)&(0.06)&(0.10)&(0.10)&(0.06)&(0.09)&(0.09)&(0.13)&(0.25)&(0.20)&(0.10)&(0.08)\\\midrule
            \multicolumn{16}{c}{{\footnotesize Dataset: \textbf{20News} \quad Model: \textbf{GPCNN}}}\\\midrule
            \multirow{2}{*}{{\scriptsize Error Rate$\downarrow$}}&36.79&36.79&36.79&36.79&36.79&36.79&36.79&36.79&36.86&36.84\greendown&36.90&36.66\greendown&43.30&43.00\greendown&37.10&37.31\redup\\
            &(0.22)&(0.22)&(0.22)&(0.22)&(0.22)&(0.22)&(0.22)&(0.22)&(0.22)&(0.21)&(0.19)&(0.19)&(1.29)&(1.37)&(0.19)&(0.25)\\
            \multirow{2}{*}{{\scriptsize smECE$\downarrow$}}&33.50&11.99\greendown&6.34&5.56\greendown&5.76&5.52\greendown&7.34&6.61\greendown&6.18&7.07\redup&9.31&7.86\greendown&28.68&11.11\greendown&6.10&\textbf{4.98}\greendown\\
            &(1.01)&(0.45)&(0.49)&(0.29)&(0.57)&(0.54)&(0.27)&(0.24)&(0.42)&(0.97)&(0.38)&(0.26)&(2.64)&(0.87)&(0.43)&(0.31)\\
            \multirow{2}{*}{{\scriptsize emECE$\downarrow$}}&28.90&12.06\greendown&6.23&5.75\greendown&5.66&5.91\redup&7.36&6.65\greendown&6.04&7.29\redup&8.98&8.12\greendown&25.79&11.13\greendown&6.16&\textbf{4.88}\greendown\\
            &(0.91)&(0.43)&(0.46)&(0.27)&(0.56)&(0.51)&(0.28)&(0.25)&(0.39)&(0.35)&(0.26)&(0.38)&(2.64)&(0.47)&(0.43)&(0.29)\\
            \multirow{2}{*}{{\scriptsize ewECE$\downarrow$}}&28.95&12.04\greendown&6.23&5.75\greendown&5.83&5.77\greendown&7.42&6.70\greendown&6.15&7.06\redup&9.02&8.04\greendown&25.78&12.23\greendown&6.18&\textbf{5.03}\greendown\\
            &(0.90)&(0.46)&(0.46)&(0.30)&(0.52)&(0.55)&(0.27)&(0.25)&(0.40)&(0.96)&(0.25)&(0.39)&(3.01)&(0.47)&(0.43)&(0.30)\\\midrule
            \multicolumn{16}{c}{{\footnotesize Dataset: \textbf{ImageNet-1K} \quad Model: \textbf{ViT-B/16}}}\\\midrule
            \multirow{2}{*}{{\scriptsize Error Rate$\downarrow$}}&18.94&18.94&18.94&18.94&18.94&18.94&18.94&18.94&19.17&19.18\redup&18.94&18.94&18.93&18.93&18.94&19.03\redup\\
            &(0.02)&(0.02)&(0.02)&(0.02)&(0.02)&(0.02)&(0.02)&(0.02)&(0.03)&(0.03)&(0.02)&(0.02)&(0.02)&(0.02)&(0.02)&(0.03)\\
            \multirow{2}{*}{{\scriptsize smECE$\downarrow$}}&5.53&5.55\redup&3.72&3.70\greendown&3.52&\textbf{3.27}\greendown&3.72&3.69\greendown&5.00&4.38\greendown&3.70&3.44\greendown&5.54&5.54&5.54&7.69\redup\\
            &(0.02)&(0.01)&(0.04)&(0.04)&(0.05)&(0.12)&(0.04)&(0.04)&(0.08)&(0.11)&(0.04)&(0.03)&(0.01)&(0.01)&(0.01)&(0.05)\\
            \multirow{2}{*}{{\scriptsize emECE$\downarrow$}}&5.54&5.54&4.16&4.14\greendown&3.57&\textbf{3.45}\greendown&4.16&4.13\greendown&5.10&5.28\redup&4.16&4.13\greendown&5.61&5.61&5.55&7.69\redup\\
            &(0.01)&(0.01)&(0.03)&(0.03)&(0.05)&(0.16)&(0.03)&(0.03)&(0.08)&(0.12)&(0.04)&(0.05)&(0.01)&(0.01)&(0.01)&(0.05)\\
            \multirow{2}{*}{{\scriptsize ewECE$\downarrow$}}&5.60&5.60&3.77&3.75\greendown&3.56&\textbf{3.31}\greendown&3.77&3.73\greendown&5.07&4.74\greendown&3.76&3.72\greendown&5.54&5.55\redup&5.61&7.69\redup\\
            &(0.02)&(0.02)&(0.05)&(0.05)&(0.05)&(0.12)&(0.05)&(0.05)&(0.09)&(0.10)&(0.04)&(0.06)&(0.01)&(0.01)&(0.02)&(0.05)
        \end{tabular}
        \label{tab:calibration_comparison_with_existing_methods}
    \end{table*}
    We conducted experiments to evaluate the effectiveness of the proposed methods using several real-world datasets.
    Detailed experimental settings are provided in~\cref{sec:detailed_experimental_settings}.

    \paragraph{Datasets, baselines, and evaluation metrics.}
    We used four datasets: TinyImageNet~\citep{deng2009imagenet} (TImageNet), ImageNet-1K~\citep{deng2009imagenet}, and CIFAR-100~\citep{cifar} for image classification, and 20NewsGroups~\citep{twentynewsgroups} (20News) for text classification.
    The preprocessing procedures followed prior work~\citep{aaron2024roadless,kumar2018trainable,paszke2019pytorch}.
    Since the proposed methods are post-hoc calibration techniques, they require uncalibrated baseline models.
    We used ResNet-50~\citep{he2016deep} for TImageNet, DenseNet-121~\citep{huang2017densely} for CIFAR-100, ViT-B/16~\citep{dosovitskiy2021image} for ImageNet-1K, and global-pooling 1D convolutional neural networks (GPCNNs)~\citep{kumar2018trainable} with 100-dimensional GloVe embeddings~\citep{pennington-etal-2014-glove} for 20News.
    We optimized the baselines using schedule-free optimizers~\citep{aaron2024roadless} for ResNet-50, DenseNet-121, and GPCNNs.
    For these models, we evaluated the error rate on a validation set at each epoch and selected the checkpoint that achieved the lowest validation error rate as the final model.
    For ViT-B/16, we used the pretrained weights provided by \texttt{torchvision}~\citep{paszke2019pytorch} instead of training models.
    We evaluated the models using classification error rate, smooth ECE (smECE)~\citep{blasiok2024smooth}, equal-mass ECE (emECE, called adaptive ECE in~\citet{mukhoti2020calibrating}), and equal-width ECE (ewECE) with 15 bins.
    We trained and evaluated the models 10 times with different seeds, including different dataset splits.
    \Cref{tab:datasets} summarizes the datasets.
    
    \paragraph{Comparison methods.}
    As described in~\cref{sec:extension_of_pb}, our proposed methods not only calibrate baseline models but can also be combined with existing calibration methods.
    Therefore, we compared the following methods with and without PB/LB: the uncalibrated baseline model described in the previous paragraph (\textbf{Uncal}), \textbf{TS}~\citep{guo2017calibration}, instance-based/parameterized TS (\textbf{IBTS/PTS})~\citep{ding2021local,tomani2022parameterized}, ensemble TS (\textbf{ETS})~\citep{zhang2020mix}, Dirichlet calibration (\textbf{Dir})~\citep{kull2019beyond}, neural clamping (\textbf{NC})~\citep{tang2024neural}, and tilt and average (\textbf{T\&A})~\citep{gyusang2024tilt}.
    Based on the discussion in~\cref{sec:lb_and_dpb}, we also compared \textbf{FC}~\citep{tao2025feature} with DPB defined in ~\cref{eq:dpb}.

    \paragraph{The effectiveness of \textbf{PB}.}
    \Cref{tab:calibration_comparison_with_existing_methods} shows the results for \textbf{PB}.
    In most settings, \textbf{PB} substantially improved the calibration of \textbf{Uncal} and provided additional improvements when combined with existing post-hoc calibration methods, while preserving predictive accuracy.
    On TImageNet, CIFAR-100, and 20News, \textbf{PB} improved \textbf{Uncal} by a large margin.
    On the other hand, \textbf{PB} (and \textbf{FC}/\textbf{T\&A}) did not improve \textbf{Uncal} on ImageNet-1K.
    This is because the ViT-B/16 model is not overconfident but slightly underconfident~\citep{le2024confidence}.
    Nevertheless, even on ImageNet-1K, PB improved the performance of several existing methods.
    Indeed, \textbf{IBTS/PTS+PB} achieved the best performance on both ImageNet-1K and TImageNet.
    \textbf{DPB} improved \textbf{FC} on all datasets except ImageNet-1K and achieved the best overall calibration on CIFAR-100 and 20News.
    These results demonstrate the effectiveness of the proposed methods and highlight the importance of both soft and hard control.

    \begin{table*}[t]
        \caption{Comparison of Error Rate, smECE, emECE, and ewECE (\%, lower is better) among the baseline (\textbf{Uncal}) and the existing methods (\textbf{TS}, \textbf{IBTS/PTS}, \textbf{ETS}, \textbf{Dir}, \textbf{NC}, and \textbf{T\&A}) without (Base) and with LB (+\textbf{LB}). We report average scores and standard errors (in round brackets) over 10 runs with different random seeds. The symbols \greendown\, and \redup\, indicate improvement and degradation, respectively. For reference, the results of \textbf{FC} and \textbf{DPB} are also shown.}
        \centering
        \tiny
        \begin{tabular}{lcgcgcgcgcgcgcgcg}
            \multirow{2}{*}{{\footnotesize \textbf{Metric}}}&\multicolumn{2}{c}{{\footnotesize \textbf{Uncal}}}&\multicolumn{2}{c}{{\footnotesize \textbf{TS}}}&\multicolumn{2}{c}{{\footnotesize \textbf{IBTS/PTS}}}&\multicolumn{2}{c}{{\footnotesize \textbf{ETS}}}&\multicolumn{2}{c}{{\footnotesize \textbf{Dir}}}&\multicolumn{2}{c}{{\footnotesize \textbf{NC}}}&\multicolumn{2}{c}{{\footnotesize \textbf{T\&A}}}&{{\footnotesize \textbf{FC}}}&{{\footnotesize \textbf{DPB}}}\\
            &{\scriptsize Base}&{\scriptsize +\textbf{LB}}&{\scriptsize Base}&{\scriptsize +\textbf{LB}}&{\scriptsize Base}&{\scriptsize +\textbf{LB}}&{\scriptsize Base}&{\scriptsize +\textbf{LB}}&{\scriptsize Base}&{\scriptsize +\textbf{LB}}&{\scriptsize Base}&{\scriptsize +\textbf{LB}}&{\scriptsize Base}&{\scriptsize +\textbf{LB}}&&\\\midrule 
            \multicolumn{16}{c}{{\footnotesize Dataset: \textbf{TImageNet} \quad Model: \textbf{ResNet-50}}}\\\midrule
            \multirow{2}{*}{{\scriptsize Error Rate$\downarrow$}}&38.92&44.25\redup&38.92&38.91\greendown&38.92&38.91\greendown&38.92&38.91\greendown&40.15&39.84\greendown&39.05&39.02\greendown&39.66&39.68\redup&41.69&39.15\greendown\\
            &(0.09)&(4.44)&(0.09)&(0.09)&(0.09)&(0.09)&(0.09)&(0.09)&(0.08)&(0.13)&(0.08)&(0.08)&(0.11)&(0.10)&(0.11)&(0.11)\\
            \multirow{2}{*}{{\scriptsize smECE$\downarrow$}}&24.26&6.49\greendown&2.58&2.28\greendown&\textbf{1.54}&1.57\redup&2.30&2.14\redup&3.40&2.33\greendown&2.75&2.44\greendown&2.90&2.48\greendown&3.27&2.57\greendown\\
            &(0.13)&(0.44)&(0.05)&(0.06)&(0.11)&(0.12)&(0.05)&(0.05)&(0.09)&(0.13)&(0.07)&(0.04)&(0.07)&(0.11)&(0.17)&(0.06)\\
            \multirow{2}{*}{{\scriptsize emECE$\downarrow$}}&21.50&6.48\greendown&2.55&2.28\greendown&\textbf{1.40}&\textbf{1.40}&2.28&2.12&3.37&2.34\greendown&2.74&2.42\greendown&2.91&2.50\greendown&3.30&2.53\greendown\\
            &(0.11)&(0.45)&(0.06)&(0.09)&(0.12)&(0.14)&(0.06)&(0.07)&(0.09)&(0.14)&(0.07)&(0.06)&(0.09)&(0.10)&(0.17)&(0.06)\\
            \multirow{2}{*}{{\scriptsize ewECE$\downarrow$}}&21.51&6.54\greendown&2.60&2.29\greendown&\textbf{1.45}&1.55\redup&2.37&2.21\greendown&3.49&2.39\greendown&2.79&2.51\greendown&2.99&2.58\greendown&3.28&2.59\greendown\\
            &(0.11)&(0.43)&(0.05)&(0.07)&(0.13)&(0.14)&(0.06)&(0.06)&(0.10)&(0.14)&(0.08)&(0.05)&(0.06)&(0.11)&(0.17)&(0.07)\\\midrule
            \multicolumn{16}{c}{{\footnotesize Dataset: \textbf{CIFAR-100} \quad Model: \textbf{DenseNet-121}}}\\\midrule
            \multirow{2}{*}{{\scriptsize Error Rate$\downarrow$}}&22.96&23.24\redup&22.96&22.96&22.96&22.97\redup&22.96&23.14\redup&22.86&22.92\redup&23.33&23.27\greendown&23.36&23.34\greendown&23.63&23.07\greendown\\
            &(0.11)&(0.12)&(0.11)&(0.11)&(0.11)&(0.11)&(0.11)&(0.13)&(0.11)&(0.10)&(0.08)&(0.09)&(0.08)&(0.09)&(0.10)&(0.10)\\
            \multirow{2}{*}{{\scriptsize smECE$\downarrow$}}&8.76&6.77\greendown&1.59&1.49\greendown&1.39&\textbf{1.29}\greendown&1.64&6.21\redup&1.80&1.76\greendown&2.19&2.04\greendown&3.40&2.89\greendown&4.43&1.32\greendown\\
            &(0.12)&(0.14)&(0.05)&(0.05)&(0.06)&(0.06)&(0.07)&(0.13)&(0.05)&(0.08)&(0.07)&(0.10)&(0.27)&(0.23)&(0.11)&(0.06)\\
            \multirow{2}{*}{{\scriptsize emECE$\downarrow$}}&7.84&6.58\greendown&1.35&1.22\greendown&1.27&1.17\greendown&1.48&6.37\redup&1.59&1.57\greendown&1.98&1.90\greendown&3.17&2.83\greendown&4.29&\textbf{1.04}\greendown\\
            &(0.10)&(0.12)&(0.04)&(0.06)&(0.08)&(0.10)&(0.08)&(0.13)&(0.07)&(0.07)&(0.09)&(0.11)&(0.27)&(0.24)&(0.11)&(0.07)\\
            \multirow{2}{*}{{\scriptsize ewECE$\downarrow$}}&7.84&6.58\greendown&1.51&1.46\greendown&1.28&\textbf{1.16}\greendown&1.62&6.27\redup&1.71&1.71&2.06&2.02\greendown&3.24&2.87\greendown&4.34&1.18\greendown\\
            &(0.10)&(0.12)&(0.08)&(0.08)&(0.07)&(0.09)&(0.10)&(0.13)&(0.06)&(0.09)&(0.09)&(0.13)&(0.25)&(0.23)&(0.10)&(0.08)\\\midrule
            \multicolumn{16}{c}{{\footnotesize Dataset: \textbf{20News} \quad Model: \textbf{GPCNN}}}\\\midrule
            \multirow{2}{*}{{\scriptsize Error Rate$\downarrow$}}&36.79&44.94\redup&36.79&36.86\redup&36.79&36.86\redup&36.79&36.80\redup&36.86&36.88\redup&36.90&36.78\greendown&43.30&51.61\redup&37.10&37.31\redup\\
            &(0.22)&(2.41)&(0.21)&(0.23)&(0.21)&(0.22)&(0.21)&(0.21)&(0.22)&(0.21)&(0.19)&(0.19)&(1.29)&(2.75)&(0.19)&(0.25)\\
            \multirow{2}{*}{{\scriptsize smECE$\downarrow$}}&33.50&12.39\greendown&6.34&6.39\redup&5.76&5.88\redup&7.34&7.31\greendown&6.18&7.99\redup&9.31&7.20\greendown&28.68&7.23\greendown&6.10&\textbf{4.98}\greendown\\
            &(1.01)&(1.06)&(0.49)&(0.48)&(0.57)&(0.54)&(0.27)&(0.29)&(0.42)&(0.66)&(0.26)&(0.54)&(3.01)&(0.84)&(0.43)&(0.31)\\
            \multirow{2}{*}{{\scriptsize emECE$\downarrow$}}&28.90&12.31\greendown&6.23&6.28\redup&5.66&5.84\redup&7.36&7.32\greendown&6.04&7.83\redup&8.98&7.01\greendown&25.78&7.39\greendown&6.16&\textbf{4.88}\greendown\\
            &(0.91)&(1.06)&(0.46)&(0.44)&(0.56)&(0.52)&(0.28)&(0.30)&(0.39)&(0.62)&(0.26)&(0.50)&(2.64)&(0.80)&(0.43)&(0.29)\\
            \multirow{2}{*}{{\scriptsize ewECE$\downarrow$}}&28.95&12.50\greendown&6.23&6.30\redup&5.83&5.95\redup&7.42&7.39\greendown&6.15&7.89\redup&9.02&7.05\greendown&25.79&7.44\greendown&6.18&\textbf{5.03}\greendown\\
            &(0.90)&(1.00)&(0.46)&(0.44)&(0.52)&(0.51)&(0.27)&(0.29)&(0.40)&(0.62)&(0.25)&(0.51)&(2.64)&(0.83)&(0.43)&(0.30)\\\midrule
            \multicolumn{16}{c}{{\footnotesize Dataset: \textbf{ImageNet-1K} \quad Model: \textbf{ViT-B/16}}}\\\midrule
            \multirow{2}{*}{{\scriptsize Error Rate$\downarrow$}}&18.94&18.93\greendown&18.94&18.93\greendown&18.93&19.40\redup&18.96&18.93\greendown&19.17&19.18\redup&18.94&18.94&18.93&18.93&18.94&19.03\redup\\
            &(0.02)&(0.02)&(0.02)&(0.02)&(0.02)&(0.04)&(0.02)&(0.02)&(0.03)&(0.03)&(0.02)&(0.02)&(0.02)&(0.02)&(0.02)&(0.03)\\
            \multirow{2}{*}{{\scriptsize smECE$\downarrow$}}&5.53&5.54\redup&3.72&3.68\greendown&3.52&\textbf{2.61}\greendown&3.72&3.68\greendown&5.00&5.01\redup&3.70&3.69\greendown&5.54&5.54&5.54&7.69\redup\\
            &(0.02)&(0.01)&(0.04)&(0.04)&(0.05)&(0.07)&(0.04)&(0.04)&(0.08)&(0.08)&(0.04)&(0.04)&(0.01)&(0.01)&(0.01)&(0.05)\\
            \multirow{2}{*}{{\scriptsize emECE$\downarrow$}}&5.54&5.55\redup&4.16&4.15\greendown&3.57&\textbf{2.64}\greendown&4.16&4.15\greendown&5.10&5.09\greendown&4.16&4.08\greendown&5.54&5.54&5.55&7.69\redup\\
            &(0.01)&(0.01)&(0.03)&(0.03)&(0.05)&(0.07)&(0.03)&(0.03)&(0.08)&(0.07)&(0.04)&(0.03)&(0.01)&(0.01)&(0.01)&(0.05)\\
            \multirow{2}{*}{{\scriptsize ewECE$\downarrow$}}&5.60&5.61\redup&3.77&3.74\greendown&3.56&\textbf{2.64}\greendown&3.77&3.74\greendown&5.07&5.09\redup&3.76&3.75\greendown&5.61&5.61&5.61&7.69\redup\\
            &(0.02)&(0.02)&(0.05)&(0.05)&(0.05)&(0.07)&(0.05)&(0.05)&(0.09)&(0.08)&(0.04)&(0.04)&(0.01)&(0.01)&(0.02)&(0.05)
        \end{tabular}
        \label{tab:ablation_lb}
    \end{table*}
    
    \paragraph{The effectiveness of LB.}
    The results for \textbf{LB} are shown in~\cref{tab:ablation_lb}.
    Similar to PB, LB substantially improved the uncalibrated baseline (\textbf{Uncal}) on TImageNet, CIFAR-100, and 20News.
    Combining existing post-hoc calibration methods with LB (+\textbf{LB}) was often effective; however, these variants were less stable than their +\textbf{PB} (shown in~\cref{tab:calibration_comparison_with_existing_methods}).
    Moreover, compared to PB, LB more frequently degraded both classification error and calibration metrics.
    Overall, these results suggest that PB is a more reliable and effective post-hoc calibrator than LB.
    
    \begin{table*}[t]
        \centering
        \caption{Comparison of Error Rate, smECE, emECE, and ewECE (\%, lower is better) among the baseline (\textbf{Uncal}) and the existing methods (\textbf{TS}, \textbf{IBTS/PTS}, \textbf{ETS}, \textbf{Dir}, \textbf{NC}, and \textbf{T\&A}) with \textbf{PB} employing constant bounds (Con) and employing instance-dependent bounds (\textbf{MLP}). We report average scores and standard errors (in round brackets) over 10 runs with different random seeds. The symbols \greendown\, and \redup\, indicate improvement and degradation, respectively.}
        \tiny
        \begin{tabular}{lcgcgcgcgcgcgcg}
            \multirow{2}{*}{{\footnotesize \textbf{Metric}}}&\multicolumn{2}{c}{{\footnotesize \textbf{Uncal+PB}}}&\multicolumn{2}{c}{{\footnotesize \textbf{TS+PB}}}&\multicolumn{2}{c}{{\footnotesize \textbf{IBTS/PTS+PB}}}&\multicolumn{2}{c}{{\footnotesize \textbf{ETS+PB}}}&\multicolumn{2}{c}{{\footnotesize \textbf{Dir+PB}}}&\multicolumn{2}{c}{{\footnotesize \textbf{NC+PB}}}&\multicolumn{2}{c}{{\footnotesize \textbf{T\&A+PB}}}\\
            &{\scriptsize Con}&{\scriptsize \textbf{MLP}}&{\scriptsize Con}&{\scriptsize \textbf{MLP}}&{\scriptsize Con}&{\scriptsize \textbf{MLP}}&{\scriptsize Con}&{\scriptsize \textbf{MLP}}&{\scriptsize Con}&{\scriptsize \textbf{MLP}}&{\scriptsize Con}&{\scriptsize \textbf{MLP}}&{\scriptsize Con}&{\scriptsize \textbf{MLP}}\\\midrule 
            \multicolumn{14}{c}{{\footnotesize Dataset: \textbf{TImageNet} \quad Model: \textbf{ResNet-50}}}\\\midrule
            \multirow{2}{*}{{\scriptsize Error Rate$\downarrow$}}&38.92&38.92&38.92&38.93\redup&38.92&38.91\greendown&38.90&38.92\redup&40.08&39.95\greendown&39.03&39.13\redup&39.67&39.73\redup\\
            &(0.09)&(0.09)&(0.09)&(0.09)&(0.09)&(0.10)&(0.10)&(0.09)&(0.16)&(0.11)&(0.10)&(0.09)&(0.10)&(0.09)\\ 
            \multirow{2}{*}{{\scriptsize smECE$\downarrow$}}&7.01&7.24\redup&2.36&1.74\greendown&\textbf{1.45}&1.54\redup&1.66&2.18\redup&2.99&1.86\greendown&2.55&1.77\greendown&2.63&1.95\greendown\\
            &(0.09)&(0.86)&(0.04)&(0.13)&(0.03)&(0.09)&(0.11)&(0.04)&(0.09)&(0.22)&(0.05)&(0.15)&(0.10)&(0.25)\\
            \multirow{2}{*}{{\scriptsize emECE$\downarrow$}}&10.16&7.90\greendown&2.41&1.68\greendown&\textbf{1.30}&1.43\redup&1.53&2.23\redup&3.29&1.75\greendown&2.58&1.71\greendown&2.71&1.84\greendown\\
            &(0.12)&(0.81)&(0.07)&(0.16)&(0.07)&(0.13)&(0.16)&(0.07)&(0.07)&(0.26)&(0.06)&(0.18)&(0.09)&(0.31)\\
            \multirow{2}{*}{{\scriptsize ewECE$\downarrow$}}&8.56&7.62\greendown&2.38&1.70\greendown&\textbf{1.37}&1.41\redup&1.58&2.28\redup&3.12&1.85\greendown&2.66&1.75\greendown&2.75&1.82\greendown\\
            &(0.23)&(0.82)&(0.05)&(0.15)&(0.07)&(0.12)&(0.13)&(0.05)&(0.09)&(0.25)&(0.06)&(0.16)&(0.10)&(0.31)\\\midrule
            \multicolumn{14}{c}{{\footnotesize Dataset: \textbf{CIFAR-100} \quad Model: \textbf{DenseNet-121}}}\\\midrule
            \multirow{2}{*}{{\scriptsize Error Rate$\downarrow$}}&22.96&22.98\redup&22.96&22.97\redup&22.96&22.97\redup&22.96&22.97\redup&22.87&22.90\redup&23.32&23.28\redup&23.33&23.39\redup\\
            &(0.11)&(0.12)&(0.11)&(0.11)&(0.11)&(0.11)&(0.11)&(0.11)&(0.07)&(0.10)&(0.09)&(0.10)&(0.09)&(0.11)\\
            \multirow{2}{*}{{\scriptsize smECE$\downarrow$}}&5.60&3.85\greendown&1.53&1.53&1.49&\textbf{1.36}\greendown&1.53&1.57\redup&2.87&1.57\greendown&1.94&2.09\redup&2.63&2.28\greendown\\
            &(0.14)&(0.91)&(0.06)&(0.06)&(0.05)&(0.06)&(0.06)&(0.06)&(0.28)&(0.06)&(0.10)&(0.06)&(0.17)&(0.19)\\
            \multirow{2}{*}{{\scriptsize emECE$\downarrow$}}&6.60&3.87\greendown&1.46&1.37\greendown&1.36&\textbf{1.23}\greendown&1.45&1.42\greendown&3.34&1.49\greendown&2.07&1.98\greendown&3.05&2.24\greendown\\
            &(0.09)&(0.92)&(0.06)&(0.08)&(0.08)&(0.11)&(0.06)&(0.08)&(0.38)&(0.07)&(0.14)&(0.06)&(0.20)&(0.20)\\
            \multirow{2}{*}{{\scriptsize ewECE$\downarrow$}}&5.98&3.79\greendown&1.50&1.51\redup&1.40&\textbf{1.22}\greendown&1.50&1.58\redup&3.02&1.49\greendown&1.95&2.07\redup&2.57&2.32\greendown\\
            &(0.11)&(0.91)&(0.10)&(0.08)&(0.06)&(0.11)&(0.10)&(0.08)&(0.34)&(0.11)&(0.13)&(0.09)&(0.20)&(0.18)\\\midrule         
            \multicolumn{14}{c}{{\footnotesize Dataset: \textbf{20News} \quad Model: \textbf{GPCNN}}}\\\midrule
            \multirow{2}{*}{{\scriptsize Error Rate$\downarrow$}}&36.79&36.79&36.79&36.98\redup&36.79&36.81\redup&36.79&36.86\redup&36.84&37.23\redup&36.66&36.69\redup&43.00&42.82\greendown\\
            &(0.21)&(0.21)&(0.21)&(0.19)&(0.21)&(0.20)&(0.21)&(0.21)&(0.21)&(0.20)&(0.19)&(0.20)&(1.37)&(0.90)\\
            \multirow{2}{*}{{\scriptsize smECE$\downarrow$}}&12.78&11.99\greendown&5.56&6.58\redup&\textbf{5.52}&6.92\redup&6.61&7.19\redup&7.07&8.13\redup&7.86&9.10\redup&11.11&7.40\greendown\\
            &(0.52)&(0.45)&(0.29)&(0.33)&(0.54)&(0.44)&(0.24)&(0.31)&(0.97)&(0.66)&(0.38)&(0.75)&(0.47)&(0.81)\\
            \multirow{2}{*}{{\scriptsize emECE$\downarrow$}}&14.46&12.06\greendown&\textbf{5.75}&6.58\redup&5.91&6.90\redup&6.65&7.21\redup&7.29&8.20\redup&8.12&9.18\redup&12.23&7.39\greendown\\
            &(0.49)&(0.43)&(0.27)&(0.32)&(0.51)&(0.43)&(0.25)&(0.32)&(0.94)&(0.72)&(0.38)&(0.72)&(0.87)&(0.78)\\
            \multirow{2}{*}{{\scriptsize ewECE$\downarrow$}}&12.85&12.04\greendown&5.75&6.68\redup&\textbf{5.77}&6.98\redup&6.70&7.31\redup&7.16&8.13\redup&8.04&9.25\redup&11.13&7.51\greendown\\
            &(0.53)&(0.46)&(0.30)&(0.30)&(0.55)&(0.41)&(0.25)&(0.32)&(0.95)&(0.66)&(0.39)&(0.72)&(0.47)&(0.77)\\\midrule
            \multicolumn{14}{c}{{\footnotesize Dataset: \textbf{ImageNet-1K} \quad Model: \textbf{ViT-B/16}}}\\\midrule
            \multirow{2}{*}{{\scriptsize Error Rate$\downarrow$}}&18.94&18.94&18.94&19.00\redup&18.93&19.08\redup&18.93&19.01\redup&19.28&19.70\redup&18.94&18.99\redup&18.93&18.93\\
            &(0.02)&(0.02)&(0.02)&(0.07)&(0.02)&(0.05)&(0.02)&(0.07)&(0.06)&(0.05)&(0.02)&(0.05)&(0.02)&(0.02)\\            
            \multirow{2}{*}{{\scriptsize smECE$\downarrow$}}&5.55&5.55&3.70&3.42\greendown&3.27&1.69\greendown&3.69&3.41\greendown&4.09&\textbf{1.30}\greendown&3.44&3.40\greendown&5.55&5.54\greendown\\
            &(0.01)&(0.01)&(0.04)&(0.29)&(0.12)&(0.37)&(0.04)&(0.29)&(0.13)&(0.13)&(0.03)&(0.29)&(0.01)&(0.01)\\
            \multirow{2}{*}{{\scriptsize emECE$\downarrow$}}&5.54&5.54&4.14&3.81\greendown&3.45&1.67\greendown&4.15&3.81\greendown&4.81&\textbf{1.22}\greendown&4.13&3.80\greendown&5.55&5.54\greendown\\
            &(0.01)&(0.01)&(0.03)&(0.34)&(0.16)&(0.43)&(0.03)&(0.34)&(0.17)&(0.13)&(0.05)&(0.34)&(0.01)&(0.01)\\
            \multirow{2}{*}{{\scriptsize ewECE$\downarrow$}}&5.60&5.60&3.75&3.45\greendown&3.31&1.65\greendown&3.73&3.44\greendown&4.72&\textbf{1.23}\greendown&3.72&3.41\greendown&5.61&5.61\\
            &(0.02)&(0.01)&(0.05)&(0.31)&(0.12)&(0.40)&(0.05)&(0.31)&(0.19)&(0.13)&(0.06)&(0.32)&(0.02)&(0.01)\\
        \end{tabular}
        \label{tab:ablation_instance_dependent_bound}
    \end{table*}
    
    \paragraph{The effectiveness of instance-dependent bounds.}
    Next, we compare the constant-bound approach (Con), defined in~\cref{eq:pb}, with an instance-dependent bound approach (\textbf{MLP}), described in~\cref{sec:extension_of_pb}.
    For \textbf{MLP}, we parameterized $a'(\bm{x})$ and $b'(\bm{x})$ as a two-layer neural network whose inputs are the top-10 elements of the logit vector $g(\bm{x})$, following the temperature network modeling in \textbf{IBTS/PTS}~\citep{tomani2022parameterized} in~\cref{eq:ibtspts}.
    The results are shown in~\cref{tab:ablation_instance_dependent_bound}.
    Using instance-dependent bounds further reduced calibration errors.
    In particular, on ImageNet-1K, \textbf{IBTS/PTS} and \textbf{Dir} were enhanced substantially.
    The calibration errors of \textbf{Dir} were also substantially reduced on CIFAR-100 and TImageNet.
    However, there were cases where calibration errors worsened; notably, on 20News, calibration errors increased in many settings.
    This may be because the validation dataset for 20News was small.
    We also observed that instance-dependent bounds can introduce ties in the predicted probabilities, which more frequently lead to slight drops in accuracy than constant bounds.
    Nonetheless, the magnitude of the accuracy degradation was very small; it was less than 0.1\% in most cases.
    
    \begin{table*}[t]
        \centering
        \caption{Comparison of Error Rate, smECE, emECE, and ewECE (\%, lower is better) among the baseline (\textbf{Uncal}) without \textbf{PB} (Base) and with \textbf{PB} under three constraint types: box constraints (+\textbf{Box}), only lower-bound constraints (+\textbf{Lower}), and only upper-bound constraints (+\textbf{Upper}). We report average scores and standard errors (in round brackets) over 10 runs with different random seeds. The symbols \greendown\, and \redup\, indicate improvement and degradation, respectively.}
        \tiny
        \begin{tabular}{lcggg}
            \multirow{2}{*}{{\footnotesize \textbf{Metric}}}&\multicolumn{4}{c}{\footnotesize {\textbf{Uncal}}}\\
            &{\scriptsize Base}&{\scriptsize +\textbf{Box}}&{\scriptsize +\textbf{Lower}}&{\scriptsize +\textbf{Upper}}\\\midrule
            \multicolumn{5}{c}{{\footnotesize Dataset: \textbf{TImageNet} \quad Model: \textbf{ResNet-50}}}\\\midrule
            \multirow{2}{*}{{\scriptsize Error Rate$\downarrow$}}&38.92&38.92&38.92&40.07\redup\\
            &(0.09)&(0.09)&(0.09)&(0.64)\\\midrule
            \multirow{2}{*}{{\scriptsize smECE$\downarrow$}}&24.26&7.01\greendown&12.13\greendown&8.47\greendown\\
            &(0.13)&(0.09)&(0.17)&(0.48)\\\midrule
            \multirow{2}{*}{{\scriptsize emECE$\downarrow$}}&21.50&10.16\greendown&16.71\greendown&10.16\greendown\\
            &(0.11)&(0.12)&(0.11)&(0.90)\\\midrule
            \multirow{2}{*}{{\scriptsize ewECE$\downarrow$}}&21.51&8.56\greendown&13.67\greendown&9.72\greendown\\
            &(0.11)&(0.23)&(0.25)&(0.91)\\\midrule
            \multicolumn{5}{c}{{\footnotesize Dataset: \textbf{CIFAR-100} \quad Model: \textbf{DenseNet-121}}}\\\midrule
            \multirow{2}{*}{{\scriptsize Error Rate$\downarrow$}}&22.96&22.96&22.96&22.96\\
            &(0.11)&(0.11)&(0.11)&(0.11)\\\midrule
            \multirow{2}{*}{{\scriptsize smECE$\downarrow$}}&8.76&5.60\greendown&7.09\greendown&6.05\greendown\\
            &(0.12)&(0.14)&(0.09)&(0.13)\\\midrule
            \multirow{2}{*}{{\scriptsize emECE$\downarrow$}}&7.84&6.60\greendown&7.40\greendown&6.53\greendown\\
            &(0.10)&(0.09)&(0.09)&(0.08)\\\midrule
            \multirow{2}{*}{{\scriptsize ewECE$\downarrow$}}&7.84&5.98\greendown&7.15\greendown&6.12\greendown\\
            &(0.10)&(0.11)&(0.09)&(0.13)\\\midrule
            \multicolumn{5}{c}{{\footnotesize Dataset: \textbf{20News} \quad Model: \textbf{GPCNN}}}\\\midrule
            \multirow{2}{*}{{Error Rate$\downarrow$}}&36.79&36.79&36.79&43.35\redup\\
            &(0.22)&(0.22)&(0.21)&(4.55)\\\midrule
            \multirow{2}{*}{{\scriptsize smECE$\downarrow$}}&33.50&11.99\greendown&12.78\greendown&14.49\greendown\\
            &(1.01)&(0.45)&(0.52)&(1.20)\\\midrule
            \multirow{2}{*}{{\scriptsize emECE$\downarrow$}}&28.90&12.06\greendown&14.46\greendown&14.50\greendown\\
            &(0.91)&(0.43)&(0.49)&(1.20)\\\midrule
            \multirow{2}{*}{{\scriptsize ewECE$\downarrow$}}&28.95&12.04\greendown&12.85\greendown&14.55\greendown\\
            &(0.90)&(0.46)&(0.53)&(1.20)\\\midrule
            \multicolumn{5}{c}{{\footnotesize Dataset: \textbf{ImageNet-1K} \quad Model: \textbf{ViT-B/16}}}\\\midrule
            \multirow{2}{*}{{\scriptsize Error Rate$\downarrow$}}&18.94&18.94&18.94&18.94\\
            &(0.02)&(0.02)&(0.02)&(0.02)\\\midrule
            \multirow{2}{*}{{\scriptsize smECE$\downarrow$}}&5.53&5.55\redup&5.54\redup&5.55\redup\\
            &(0.02)&(0.01)&(0.01)&(0.01)\\\midrule
            \multirow{2}{*}{{\scriptsize emECE$\downarrow$}}&5.54&5.54&5.54&5.55\redup\\
            &(0.01)&(0.01)&(0.01)&(0.01)\\\midrule
            \multirow{2}{*}{{\scriptsize ewECE$\downarrow$}}&5.60&5.60&5.60&5.61\redup\\
            &(0.02)&(0.02)&(0.01)&(0.01)\\\midrule
        \end{tabular}
        \label{tab:ablation_bounds}
    \end{table*}

    \paragraph{Box constraints, lower-bound constraints, or upper-bound constraints?}
    We compare the uncalibrated baseline (\textbf{Uncal}) with three variants of \textbf{PB}: with box constraints (+\textbf{Box}), with only a lower bound (+\textbf{Lower}), and with only an upper bound (+\textbf{Upper}).
    The results are summarized in~\cref{tab:ablation_bounds}.
    Using only lower bounds or only upper bounds also performed well on TImageNet, CIFAR-100, and 20News, while providing little to no improvement on ImageNet-1K.
    Overall, \textbf{PB} with box constraints (+\textbf{Box}) performed best in most settings.
    Compared to +\textbf{Box}, +\textbf{Lower} yielded smaller calibration gains, whereas +\textbf{Upper} more frequently degraded accuracy.
    These results suggest that imposing both lower and upper bounds is important for effective post-hoc calibration.
    
\section{Conclusion}
    We proposed a novel post-hoc calibration method, probability bounding (PB), which mitigates overconfidence and underconfidence by enforcing box constraints on output probabilities.
    To implement PB, we introduced the box-constrained softmax ($\BCSoftmax$) function, a generalization of $\Softmax$ with box constraints on the output probability vector, and presented an efficient and exact algorithm for computing $\BCSoftmax$.
    We provided theoretical guarantees for PB and proposed two variants of PB.
    Our experimental results indicated that the proposed methods can calibrate the baseline models and enhance the existing post-hoc calibration methods.
    We believe that $\BCSoftmax$ has broad applicability across various domains.

\section*{Acknowledgements}
    This work was supported by Japan Science and Technology Agency (JST), Core Research for Evolutionary Science and Technology CREST Program, Grant Number JPMJCR21D1.

\bibliographystyle{plainnat}
\bibliography{ref}

\newpage
\appendix
\input{appendix_additional_analysis.tex}
\input{appendix_detailed_settings}
\input{appendix_experiments}

\newpage
\input{appendix_proofs}
\newpage
\input{appendix_algorithms}
\end{document}

%% file: appendix_additional_analysis.tex
\section{Additional Analysis}
\subsection{Basic Properties and Jacobians of \texorpdfstring{$\BCSoftmax$}{BCSoftmax}}
\label{sec:basic_properties}
We show basic properties and compute the Jacobian matrices of $\BCSoftmax$.

    \paragraph{Basic Properties.}
    Like $\Softmax$, changing the temperature is equivalent to changing the scale of the logits.
    Moreover, $\BCSoftmax$ is invariant to constant offsets.
    \begin{propositionmd}
        For all $\tau > 0$, $\bm{g} \in \real^K$, $(\bm{a}, \bm{b}) \in B^K$, and $z \in \real$,
        \begin{align}
            \BCSoftmax_{\tau}(\bm{g}; (\bm{a}, \bm{b})) &= \BCSoftmax_{1}(\bm{g}/\tau; (\bm{a}, \bm{b})) \quad and
            \label{eq:bc_temp_is_scale_of_logits}\\
            \BCSoftmax_{\tau}(\bm{g}; (\bm{a}, \bm{b})) &= \BCSoftmax_{\tau}(\bm{g} - z \cdot \bm{1}_K; (\bm{a}, \bm{b})).
            \label{eq:bc_offset_invariant}
        \end{align}
        \label{prop:temp_offset}
    \end{propositionmd}
    \begin{proof}
        Since $\argmax_x f(x) = \argmax_x f(x)/\tau$ for all $\tau>0$, we have
        \begin{align}
            \BCSoftmax_{\tau}(\bm{g}; (\bm{a}, \bm{b})) &\coloneqq \argmax_{\hm{p} \in \Delta^K, \bm{a}\preceq \hm{p} \preceq\bm{b}} \bm{g}^\top \hm{p} + \tau H(\bm{p}) = \argmax_{\hm{p} \in \Delta^K, \bm{a}\preceq \hm{p} \preceq\bm{b}} \bm{g}^\top \hm{p} / \tau + H(\bm{p})\\
            &= \BCSoftmax_{1}(\bm{g}/\tau; (\bm{a}, \bm{b})). \tag{\ref{eq:bc_temp_is_scale_of_logits}}
        \end{align}
        Since the sum of elements of the probability vector $\hm{p}$ is 1, we have
        \begin{align}
            \BCSoftmax_{\tau}(\bm{g} - z \cdot \bm{1}_K; (\bm{a}, \bm{b})) &\coloneqq \argmax_{\hm{p} \in \Delta^K, \bm{a}\preceq \hm{p} \preceq\bm{b}} (\bm{g} - z \cdot \bm{1}_K)^\top \hm{p} + \tau H(\bm{p})  \\
            &= \argmax_{\hm{p} \in \Delta^K,\bm{a}\preceq \hm{p} \preceq\bm{b}} \bm{g}^\top \hm{p}  - z + \tau H(\bm{p})\\
            &= \argmax_{\hm{p} \in \Delta^K, \bm{a}\preceq \hm{p} \preceq\bm{b}} \bm{g}^\top \hm{p}  + \tau H(\bm{p})\\
            &= \BCSoftmax_{\tau}(\bm{g}; (\bm{a}, \bm{b})). \tag{\ref{eq:bc_offset_invariant}}
        \end{align}
    \end{proof}    
    
    \paragraph{Jacobians.} When machine learning models employing $\BCSoftmax$ are optimized using gradient-based methods, the Jacobian matrices of $\BCSoftmax$ play a crucial role.
    From~\cref{eq:bc_temp_is_scale_of_logits}, we can set $\tau=1$ without loss of generality, which simplifies the derivation of the Jacobian.
    The $\BCSoftmax$ function is differentiable everywhere except at boundary points $(\bm{g}', (\bm{a}', \bm{b}'))$, where the indices of lower- or upper-bounded probabilities change.
    For differentiable $(\bm{g}, (\bm{a}, \bm{b}))$, Jacobian matrices are derived directly from~\cref{thm:relationship_bc_soft}:
    \begin{align}
        \frac{\partial \BCSoftmax_{1}(\bm{g}; (\bm{a}, \bm{b}))[i]}{\partial g_j} &= \left[\mathrm{Diag}(\bm{q}) - \bm{q} \bm{q}^\top/s \right]_{i,j},\\
        \frac{\partial \BCSoftmax_{1}(\bm{g}; (\bm{a}, \bm{b}))[i]}{\partial a_j} &= \left[\mathrm{Diag}(\bm{m}) - \bm{q}\bm{m}^\top/s\right]_{i,j},\\
        \frac{\partial \BCSoftmax_{1}(\bm{g}; (\bm{a}, \bm{b}))[i]}{\partial b_j} &= \left[\mathrm{Diag}(\bm{m}^\prime) - \bm{q}(\bm{m}^\prime)^\top/s\right]_{i,j},
    \end{align}
    where $\bm{q} = \bm{p} \circ (\bm{1}_K - \bm{m}) \circ (\bm{1}_K - \bm{m}^\prime)$, $\bm{p}=\BCSoftmax_1(\bm{g}; (\bm{a}, \bm{b}))$, $\bm{m} \in \{0, 1\}^K$ is the flag (mask) vector such that $m_i = 1$ if $p_i = a_i$; otherwise $m_i=0$.
    Similarly, $\bm{m}^\prime$ is the $K$-dimensional boolean vector such that $m^\prime_i=1$ if $p_i = b_i$; otherwise $m^\prime_i=0$.
    Since all Jacobians are in the form of the difference between a diagonal matrix and a matrix of rank 1, both vector-Jacobian and Jacobian-vector products can be performed in $O(K)$ time.
    Therefore, our proposed $\BCSoftmax$ function can be used easily and efficiently in various deep neural networks.

%% file: appendix_detailed_settings.tex
\section{Detailed Experimental Settings}
    \label{sec:detailed_experimental_settings}
    In this section, we describe the detailed experimental settings.
    We ran all experiments on a server with two AMD EPYC 7413 CPUs, six NVIDIA RTX A6000 GPUs, and 512GB RAM.

\subsection{Detailed Settings of Datasets}
    We obtained TImageNet from \url{http://cs231n.stanford.edu/tiny-imagenet-200.zip}.
    We used \texttt{datasets.CIFAR100} module in the \texttt{torchvision} library\footnote{\url{https://docs.pytorch.org/vision/0.24/generated/torchvision.datasets.CIFAR100.html}} and \texttt{datasets.fetch\_20newsgroups} in the \texttt{sklearn} library\footnote{\url{https://scikit-learn.org/1.7/modules/generated/sklearn.datasets.fetch_20newsgroups.html}} to get the CIFAR-100 and 20News datasets, respectively.
    We got ImageNet-1K from the Hugging Face repository\footnote{\url{https://huggingface.co/datasets/ILSVRC/imagenet-1k}}.
    Note that we obtained only 50,000 validation images from this repository since we used the pretrained model \texttt{ViT\_B\_16\_Weights.IMAGENET1K\_V1} in the \texttt{torchvision} library\footnote{\url{https://docs.pytorch.org/vision/0.24/models/generated/torchvision.models.vit_b_16.html}} as the uncalibrated baseline model, and there are no labels for the 100,000 test images.
    The preprocessing procedures followed prior work~\cite{aaron2024roadless,kumar2018trainable,paszke2019pytorch}, with dataset-specific details provided below.  
    \begin{itemize}  
        \item \textbf{TImageNet and CIFAR-100}. Following~\cite{aaron2024roadless}, we applied online data augmentation during training: random horizontal flipping, reflection padding of 4 pixels on each side, random cropping (to 32×32 for CIFAR-100 and 64×64 for TImageNet), and normalization to zero mean and unit variance.
        At evaluation time (i.e., testing), we applied only the same normalization used during training.
        \item \textbf{20News}. We tokenized the texts using \texttt{TextVectorization} in the \texttt{Keras} library\footnote{\url{https://keras.io/api/layers/preprocessing_layers/text/text_vectorization/}}.
        Following~\cite{kumar2018trainable}, we set the vocabulary size to 20,000 and the maximum sequence length to 1,000, and removed the headers.
        \item \textbf{ImageNet-1K}. We transformed the input images using \texttt{ViT\_B\_16\_Weights.IMAGENET1K\_V1.transforms} in the \texttt{torchvision} library.
    \end{itemize}

\subsection{Detailed Settings of Baselines}
    All models except ViT-B/16 for ImageNet-1K were trained with xent loss.  
    We optimized ResNet-50 and DenseNet-121 using schedule-free stochastic gradient descent (SF-SGD)~\cite{aaron2024roadless}, and GPCNN (for 20News) using SF-AdamW~\cite{aaron2024roadless}.
    Following~\cite{aaron2024roadless}, we set the learning rate to $5.0$ for the DenseNet-121 on CIFAR-100.
    For the other cases, we tuned the learning rate based on validation accuracy.
    Other hyperparameters were selected according to previous work~\cite{aaron2024roadless,kumar2018trainable} or library defaults\footnote{\url{https://github.com/facebookresearch/schedule_free} (version: 1.4.1)}.
    The ResNet-50 model was based on \texttt{models.resnet50} from the \texttt{torchvision} library\footnote{\url{https://pytorch.org/vision/0.24/models/generated/torchvision.models.resnet50.html}}.
    We modified the kernel size, stride, and padding of the input convolutional layer to 3, 1, and 1, respectively.  
    The number of output units was set to $K = 200$.  
    The DenseNet-121 model was adapted from a tutorial implementation in the \texttt{lightning} library\footnote{\url{https://lightning.ai/docs/pytorch/2.5.1/notebooks/course_UvA-DL/04-inception-resnet-DenseNet.html}}.  
    \Cref{tab:baselines} summarizes the baseline configurations.
    \begin{table*}[t]
        \centering
        \caption{Summary of the baselines.}
        \begin{tabular}{cccccc}
            Dataset & Model & Optimizer & Step Size & Batch Size & Epochs \\ \midrule
            TImageNet&ResNet-50&SF-SGD&$\{1.5, 5.0, 15.0\}$&256&100\\\midrule
            CIFAR-100&DenseNet-121&SF-SGD&5.0&64&300\\\midrule
            20News&GPCNN&SF-AdamW&$\{10^{-3}, 10^{-2}, 10^{-1}\}$&128&100\\ \midrule
            ImageNet-1K&ViT-B/16&\multicolumn{4}{c}{N/A (Using the pretrained weights)}
        \end{tabular}
        \label{tab:baselines}
    \end{table*}

\subsection{Detailed Settings of Post-hoc Methods}
    For all post-hoc methods except \textbf{T\&A} without \textbf{PB}, we used the Adam optimizer~\cite{kingma2015adam} with default hyperparameters.
    The additional parameter introduced in \textbf{T\&A} was determined by Algorithm 1 in~\cite{gyusang2024tilt} (see~\cref{eq:tilt_and_average}).
    We set the number of epochs to 500 except for \textbf{NC} on ImageNet-1K.
    Since \textbf{NC} on ImageNet-1K had a high computational cost, we set the number of epochs to 200.
    The batch size was set to 256 for TImageNet, CIFAR-100, and ImageNet-1K, and 64 for 20News.

    \paragraph{PB, LB, and DPB.} For PB, to avoid the constrained optimization problem, we reparameterized $a$ and $b$ as $a=\sigma(a^\prime)/K$ and $b=1/K + (1-1/K)\sigma(b^\prime)$, and optimized $a^\prime$ and $b^\prime$ in $\real$ instead of $a$ and $b$.
    Similarly, for LB, we reparameterized $c = \lVert g(\bm{x})\rVert_2 \mathrm{tanh}(c')$ and $C=\lVert g(\bm{x})\rVert_2 \mathrm{tanh}(c'+\mathrm{softplus}(C'))$, and optimized $c^\prime$ and $C^\prime$ in $\real$.
    For DPB, we reparameterized $\bm{w}$ and $d$ as $\bm{w} =\exp(\bm{w}^\prime)$ and $d=\exp(d^\prime)$, and optimized $\bm{w}^\prime \in \real^{K}$ and $d^\prime\in \real$.
    
    \paragraph{TS~\citep{guo2017calibration} and ETS~\citep{zhang2020mix}.} 
    As described in~\cref{sec:calib_setup_def_existing}, the calibration map of \textbf{TS} was defined as
    \begin{align}
        f_{\mathrm{TS}}(\bm{x}; \tau) \coloneqq \Softmax_{\tau}(g(\bm{x})), \quad \tau > 0.
    \end{align}
    \textbf{ETS} calibrated the baseline model as
    \begin{align}
        f_{\mathrm{ETS}}(\bm{x}; \tau, \bm{w}) \coloneqq w_1 f_{\mathrm{TS}}(\bm{x}; \tau) + w_2 f(\bm{x}) + w_3 \frac{1}{K} , \quad \tau > 0, \ \bm{w}\in\Delta^3.
    \end{align}
    We reparameterized $\tau$ as $\tau = \exp(\tau')$ and optimized $\tau' \in \real$.
    We applied the projection onto the simplex~\cite{duchi2008efficient} to $\bm{w}$ after each gradient descent step.
    \paragraph{IBTS/PTS~\cite{ding2021local,tomani2022parameterized}.}
    \textbf{IBTS/PTS} was defined as
    \begin{align}
        f_{\mathrm{IBTS/PTS}}(\bm{x}; \theta) \coloneqq \Softmax_{\tau(\bm{x}; \theta)}(g(\bm{x})),
        \label{eq:ibtspts}
    \end{align}
    where $\tau(\dot; \theta):\mathcal{X}\to\real_{>0}$ outputs an instance-dependent temperature and $\theta$ is its parameter.
    The definition of $\tau(\cdot; \theta)$ followed~\citet{tomani2022parameterized}: it is a neural network with two fully connected hidden layers with five nodes, whose inputs are the top-10 elements of $g(\bm{x})$.
    
    \paragraph{Dir~\cite{kull2019beyond}.}
    We defined \textbf{Dir} using the linear parameterization model in~\cite{kull2019beyond}:
    \begin{align}
        f_{\mathrm{Dir}}(\bm{x}; \bm{W}, \bm{w}^{\prime}) \coloneqq \Softmax_1( \bm{W}\log f(\bm{x}) + \bm{w}^\prime),
    \end{align}
    where $\bm{W} \in \real^{K \times K}$ and $\bm{w}^\prime \in \real^K$.
    When optimizing $\bm{W}$ and $\bm{w}^\prime$ by minimizing the empirical risk on the validation dataset, we also introduced off-diagonal and intercept regularization~\citep{kull2019beyond}:
    \begin{align}
        \frac{\lambda}{K(K-1)} \sum_{i\neq j}w_{ij}^2 + \frac{\mu}{K}\sum_{i}(w^{\prime}_i)^2,
    \end{align}
    where $\lambda>0$ and $\mu > 0$ are hyperparameters of regularization strength.
    We tuned these parameters on the validation set to preserve accuracy as much as possible after post-hoc calibration.
    Then, $\lambda$ and $\mu$ were set to $10^{7}$ for TImageNet, CIFAR-100, and 20News, and to $10^9$ for ImageNet-1K.

    \paragraph{NC~\citep{tang2024neural}.}
    We defined \textbf{NC} as
    \begin{align}
        f_{\mathrm{NC}}(\bm{x}; \tau, \bm{\delta}) = \Softmax_{\tau}(g(\bm{x}+\bm{\delta})),
    \end{align}
    where $\tau > 0$ is the temperature parameter and $\bm{\delta} \in \mathcal{X}$ is the perturbation to be optimized.
    Note that $\mathcal{X} = \real^{3 \times d\times d}$ for TImageNet ($d=64$), CIFAR-100 ($d=32$), and ImageNet-1K ($d=224$) and $\mathcal{X} = \real^{1000 \times 100}$ for 20News.
    We optimized $\tau$ (with reparameterization $\tau=\exp (\tau^\prime)$) and $\bm{\delta}$ by minimizing the empirical risk on the validation dataset with the regularization $0.1 \lVert \bm{\delta} \rVert_2^2$.
    
    \paragraph{T\&A~\citep{gyusang2024tilt}.}
    Similar to \textbf{FC}, assume that the logit function is defined as $g(\bm{x}) = \bm{W}_{\mathrm{linear}}z(\bm{x}) + \bm{w}_{\mathrm{bias}}$, where $\bm{W}_{\mathrm{linear}} \in \real^{K \times D}$ and $\bm{w}_{\mathrm{bias}}\in \real^K$.
    Then, \textbf{T\&A}~\cite{gyusang2024tilt} tilted the weight matrix $\bm{W}_{\mathrm{linear}}$ by $\bm{R} \in \real^{D\times D}$:
    \begin{align}
        f_{\mathrm{T\&A}}(\bm{x}; \bm{R}) = \Softmax_{1}((\bm{W}_{\mathrm{linear}}\bm{R})z(\bm{x}) + \bm{w}_{\mathrm{bias}}).
        \label{eq:tilt_and_average}
    \end{align}
    We determined $\bm{R}$ by the algorithm proposed by~\citet{gyusang2024tilt}.

%% file: appendix_experiments.tex
\section{Runtime Comparison of \texorpdfstring{$\BCSoftmax$}{BCSoftmax} with Existing Method}
    \label{sec:runtime_comparison}
    We compare the execution time of the forward computation of the proposed algorithm for $\BCSoftmax$ with that of the existing general algorithm for convex optimization layers~\citep{agrawal2019differentiable} on synthetic data.
    We ran the experiments on a server with AMD EPYC 7413 CPUs, NVIDIA RTX A6000 GPUs, and 512GB RAM.

    \paragraph{Setup.}
    We generated a minibatch of logits $\{\bm{g}_n\in \real^{K}\}_{n=1}^{N_{\mathrm{batch}}}$ according to $\mathcal{N}(0, 3)$ with $N_{\mathrm{batch}}=128$ and $K = 2^5, 2^6, \ldots, 2^{10}$.
    A batch of upper bound vectors $\{\bm{b}_n\in U^{K}\}_{n=1}^{N_{\mathrm{batch}}}$ was created by sampling each element independently from the uniform distribution on $[0,1]$, i.e., $b_{n, k} \sim \mathrm{Uniform}(0,1)$ and normalizing as $b_{n, k} \gets b_{n, k} / \min \{1, \sum_{j=1}^K b_{n,j}\}$.
    A batch of lower bound vectors $\{\bm{a}_n\in L^{K}\}_{n=1}^{N_{\mathrm{batch}}}$ was created by sampling in a similar way to $b_{n, k}$: $a_{n, k} \sim \mathrm{Uniform}(0, 1/K)$ and $a_{n, k} \gets \min \{a_{n,k}, b_{n,k}\}$.
    Although~\cref{alg:bcsoftmax} runs in $O(K \log K)$ time, it is not GPU-friendly due to the non-trivial binary search.
    Thus, we implemented an algorithm that runs in $O(K^2 \log K)$ but is more GPU-friendly in PyTorch~\citep{paszke2019pytorch}.
    It computes $y(k)$ for all $k$ and then determines $\rho$ na\"ively; see~\cref{alg:bcsoftmax_gpu_friendly} for more details.
    For the existing method~\citep{agrawal2019differentiable}, we used the \texttt{cvxpylayers} library provided by the authors.

    \paragraph{Result.}
    \begin{figure}
        \centering
        \includegraphics[width=0.70\linewidth]{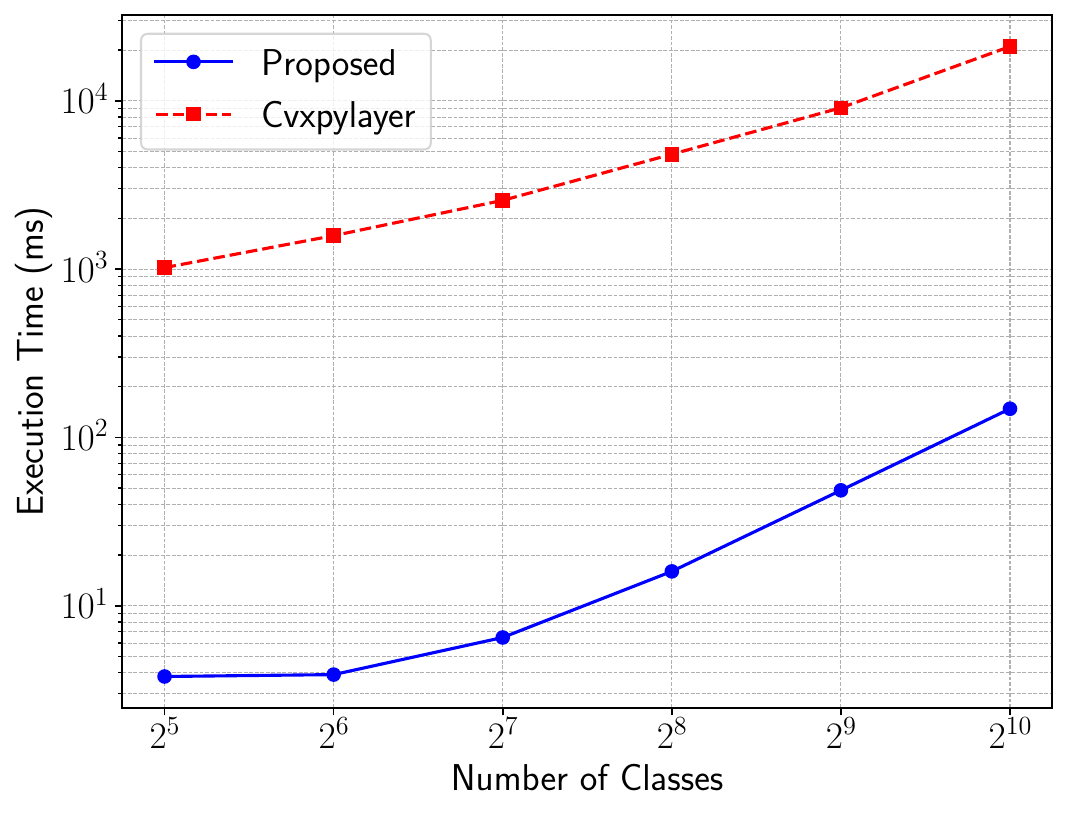}
        \caption{Runtime comparison of the proposed exact algorithm with the existing method. The proposed method is 150-400$\times$ faster than the existing method.}
        \label{fig:runtime_comparison}
    \end{figure}
    \Cref{fig:runtime_comparison} shows the results.
    Our algorithm is 150-400$\times$ faster than the existing algorithm.
    Moreover, although the existing algorithm~\cite{agrawal2019differentiable} outputs an approximate solution to~\cref{eq:bcsoftmax}, our algorithm outputs the exact solution.
    Therefore, our algorithm is more efficient and useful than the existing one.

%% file: appendix_proofs.tex
\section{Proofs}
\label{sec:proofs}

\subsection{Proof of \texorpdfstring{\cref{thm:relationship_ub_soft}}{Theorem \ref{thm:relationship_ub_soft}}}
    \begin{proof}
        By~\cref{prop:temp_offset}, we may assume $\tau = 1$ without loss of generality.
        Consider the optimization problem in~\cref{eq:bcsoftmax} with $\bm{a}=\bm{0}_K$.
        Its Lagrangian is
        \begin{align}
            L(\bm{p}, \lambda, \bm{\beta}) = \bm{g}^\top \bm{p} - \sum_{k=1}^K p_k\log p_k - \lambda \left(\sum_{k=1}^K p_k - 1\right) - \bm{\beta}^\top (\bm{p}-\bm{b}),
            \label{eq:ubsoft_lagrangian}
        \end{align}
        where $\lambda \in \real$ and $\bm{\beta} \in \real^K_{\ge 0}$ are Lagrange multipliers.
        Differentiating with respect to $p_k$ yields the stationarity condition
        \begin{align}
            \frac{\partial L}{\partial p_k} = g_k - (1 + \log p_k) - \lambda - \beta_k = 0 \\
            \implies p_k = \exp(g_k - \beta_k) / Z, \quad \text{where} \ Z = \exp(\lambda + 1).
        \end{align}
        From the simplex constraint $\sum_k p_k=1$, we have $Z = \sum_{k} \exp (g_k - \beta_k)$, which proves~\cref{eq:relationship_ub_soft}:
        \begin{align}
            p_k = \frac{\exp(g_k - \beta_k)}{\sum_{i=1}^K \exp(g_i - \beta_i)} = \Softmax_1(\bm{g}-\bm{\beta})[k]. \tag{\ref{eq:relationship_ub_soft}}
        \end{align}
        By the complementary slackness KKT condition, we have
        \begin{align}
            \beta_k (p_k - b_k) = 0 \implies \beta_k = 0 \quad \text{or} \quad p_k = b_k,
        \end{align}
        and this implies~\cref{eq:ub_beta_indices}.
        The normalization term $Z$ is derived by a simple calculation:
        \begin{align}
            \sum_{i} p_i &= \sum_{i: \beta_i > 0} \exp(g_i - \beta_i)/Z + \sum_{i: \beta_i = 0}\exp(g_i)/Z \\
            &= \sum_{i: \beta_i > 0} b_i + \sum_{i: \beta_i = 0}\exp(g_i)/Z = 1\\
            \implies Z &= \frac{\sum_{i:\beta_i = 0} \exp(g_i)}{1- \sum_{i: \beta_i>0}b_i} \coloneqq \frac{r}{s}.
        \end{align}
        Strictly speaking, $Z$ is not defined when $s=0$.
        However, $s=0$ means $\beta_k > 0$ and $p_k = b_k$ for all $k \in [K]$.
        Therefore, $Z$ does not appear in~\cref{eq:ub_beta_indices} in this case.
    \end{proof}

\subsection{Proof of \texorpdfstring{\cref{thm:relathinship_ub_soft_sorted}}{Theorem \ref{thm:relathinship_ub_soft_sorted}}}
    For the proof of~\cref{thm:relathinship_ub_soft_sorted}, we provide additional lemmas.
    First, we provide the following lemma that provides a recursive structure of $\BCSoftmax$ in a special case.
    \begin{lemmamd}
        For given $\bm{g} \in \real^K$ and $(\bm{a}, \bm{b}) \in B^K$ such that $a_1 = b_1, \ldots, a_k=b_k$ and $\sum_{i=1}^k a_k < 1$ for some $k \in [K-1]$, the following holds:
        \begin{align}
            &\BCSoftmax_1(\bm{g}; (\bm{a}, \bm{b})) \nonumber \\
            &= \bm{a}_{1:k}\concat s_k \cdot \BCSoftmax_1(\bm{g}_{k+1:K}; (\bm{a}_{k+1:K}/s_k, \bm{b}_{k+1:K}/s_k)) \\
            &=\bm{b}_{1:k} \concat s_k \cdot \BCSoftmax_1(\bm{g}_{k+1:K}; (\bm{a}_{k+1:K}/s_k, \bm{b}_{k+1:K}/s_k)), \\
            \text{where} \quad
            &s_k \coloneqq 1 - \sum_{i=1}^{k}a_i = 1 - \sum_{i=1}^{k}b_i.
        \end{align}
        \label{lemma:bcsoftmax_fixed}
    \end{lemmamd}
    \begin{proof}
        Since $a_1 = b_1, \ldots, a_k=b_k$ and $s_k = 1 - \sum_{i=1}^k a_k > 0$, we have
        \begin{align}
            \BCSoftmax_1(\bm{g}; (\bm{a}, \bm{b})) &=\argmax_{\bm{p} \in \Delta^K, \bm{a}\preceq \bm{p} \preceq\bm{b}} \bm{g}^\top \bm{p} - \sum_{k=1}^K p_k \log p_k \tag{\ref{eq:bcsoftmax}}\\
            &= \bm{a}_{1:k} \concat \bm{q}, \\
            \text{where} \quad  \bm{q} &= \argmax_{\bm{a}_{k+1:K} \preceq \bm{p} \preceq \bm{b}_{k+1:K}, \bm{p} \in s_k \cdot \Delta^{K-k}} \bm{g}_{k+1:K}^\top\bm{p} - \sum_{i=1}^{K-k} p_i \log p_i.
        \end{align}
        Moreover, $(\bm{a}_{k+1:K}/s_k, \bm{b}_{k+1:K}/s_k) \in B^{K-k}$ because
        \begin{align}
            (\bm{a} \preceq \bm{b})\land (s_k >0)\ \implies \frac{\bm{a}_{k+1:K}}{s_k} \preceq\frac{\bm{b}_{k+1:K}}{s_k}, \\
            \sum_{i=k+1}^K a_i \le 1-\sum_{i=1}^k a_i\implies \sum_{i=k+1}^K a_{i}/s_k \le 1, \\
            \sum_{i=k+1}^{K} b_i \ge 1 - \sum_{i=1}^k b_i \implies \sum_{i=k+1}^K b_i/s_k \ge 1.
        \end{align}
        Therefore, by defining $\bm{p}^\prime = \bm{p}/s_k$, we have
        \begin{align}
            \bm{q} &= \argmax_{\bm{a}_{k+1:K}/s_k \preceq \bm{p}/s_k \preceq \bm{b}_{k+1:K}/s_k, \bm{p}/s_k \in \cdot \Delta^{K-k}} \bm{g}_{k+1:K}^\top\bm{p} - \sum_{i=1}^{K-k} p_i \log p_i \\
            &= s_k \cdot \argmax_{\bm{a}_{k+1:K}/s_k \preceq \bm{p}' \preceq \bm{b}_{k+1:K}/s_k, \bm{p}' \in \cdot \Delta^{K-k}} \bm{g}_{k+1:K}^\top\bm{p}' - \sum_{i=1}^{K-k} p^\prime_i \log p^\prime_i\\
            &=  s_k \cdot \BCSoftmax_1(\bm{g}_{k+1:K}; (\bm{a}_{k+1:K}/s_k, \bm{b}_{k+1:K}/s_k)).
        \end{align}
    \end{proof}
    
    The following lemma suggests the importance of the ratio of $b_k$ to $\exp(g_k)$, i.e., $b_k / \exp(g_k)$.
    \begin{lemmamd}
        Let $\bm{g} \in \mathbb{R}^K$ be a logit vector and $\bm{b} \in U^K$ be an upper bound vector.
        Define $\bm{p} \coloneqq \UBSoftmax_1(\bm{g}, \bm{b})$ and $V \coloneqq \{k \in [K] \mid p_k = b_k\}$. 
        If $i \in V$ and $b_j / \exp(g_j) \le b_i /\exp(g_i)$, then $j \in V$.
        \label{lemma:ratio_of_upper_bound_and_exp_logit}
    \end{lemmamd}
    \begin{proof}
        By~\cref{thm:relationship_ub_soft}, there exist $\beta_i, \beta_j \ge 0$ such that $p_i = \exp(g_i - \beta_i)/Z \le b_i$ and $p_j = \exp(g_j - \beta_j)/Z \le b_j$, where $Z > 0$ is the normalization term.
        In addition, if $i \in V$ and $b_j / \exp(g_j) \le b_i /\exp(g_i)$, we have
        \begin{align}
            i \in V &\implies \ \beta_i > 0 \quad\text{and}\\
            \frac{b_j}{\exp(g_j)} \le \frac{b_i}{\exp(g_i)} &\implies b_j \le b_i \frac{\exp(g_j)}{\exp(g_i)}.
        \end{align}
        Then, we obtain the following inequality
        \begin{align}
            \frac{\exp(g_j-\beta_j)}{Z} \le b_j \le b_i \frac{\exp(g_j)}{\exp(g_i)} \implies& \exp(-\beta_j) \le b_i\frac{Z}{\exp(g_i)} = \exp(-\beta_i) < 1\\
            \implies & \beta_j \ge \beta_i > 0,
        \end{align}
        and this implies $j \in V$.
    \end{proof}
    
    Then, by~\cref{lemma:ratio_of_upper_bound_and_exp_logit} and~\cref{thm:relationship_ub_soft}, we immediately obtain the following corollary.
    \begin{corollarymd}
        For all $\bm{g} \in \real^K$, and $\bm{b} \in U^K$ such that $b_1/\exp(g_1) \le \cdots \le b_K / \exp(g_K)$, there exists $0 \le \rho \le K$ such that
        \begin{align}
            k \le \rho &\implies p_k = b_k, \\
            k > \rho &\implies p_k < b_k \implies p_k \propto \exp(g_k),
        \end{align}
        where $\bm{p} = \UBSoftmax_\tau(\bm{g}, \bm{b})$.
        Therefore,
        \begin{align}
            \bm{p} \in & \{ p(k): p(k) \preceq \bm{b}, p(k) \in \Delta^K, k \in \{0, \ldots, K-1 \mid s_k > 0\}\,
        \end{align}
        where
        \begin{align}
            \label{eq:cand_of_y}
            p(k) \coloneqq & \bm{b}_{1:k} \concat (\exp(g_{k+1}), \ldots, \exp(g_K))/Z_k
             = \bm{b}_{1:k} \concat s_k\cdot \Softmax_1(\bm{g}_{k+1:K}), \\
             r_k \coloneqq & \sum_{i=1}^{K} \exp (g_i / \tau) - \sum_{i=1}^{k}\exp(g_{i}/\tau)=\sum_{i=k+1}^{K}\exp(g_i/\tau),  \\
             r_0 \coloneqq & \sum_{i=1}^{K} \exp (g_i / \tau), s_k \coloneqq  1 - \sum_{i=1}^{k}b_i, \ s_0 \coloneqq 1, \ \text{and}, \ Z_k \coloneqq \frac{r_k}{s_k}.
        \end{align}
        \label{coro:relathinship_ub_soft_sorted_weak}
    \end{corollarymd}
    
    \begin{algorithm}[ht]
        \caption{$O(K^2)$ algorithm for $\UBSoftmax_\tau$}
        \label{alg:ubsoftmax_square}
        \begin{algorithmic}[1]
            \Input $\bm{g} \in \real^K, \bm{b} \in U^K$
            \State $\bm{g} \gets \bm{g}/\tau$
            \State Sort $\bm{g}$ and $\bm{b}$ as $b_{1}/\exp (g_{1}) \le \cdots \le b_{K}/\exp (g_{K})$
            \State $p(k) \gets \bm{b}_{1:k} \concat s_k\cdot \Softmax_1(\bm{g}_{k+1:K})$ for all $k \in \{0, \ldots, K-1\}$ such that $s_k > 0$
            \State $S \gets \{k \in \{0, \ldots, K-1\}\mid p(k) \preceq \bm{b}, p(k) \in \Delta^K\}$
            \State $\rho \gets \argmax_{k \in S} \bm{g}^\top \bm{p}(k) + H(p(k))$
            \State $\bm{p} \gets p(\rho)$
            \State Undo sorting $\bm{p}$
            \Output $\bm{p}$
        \end{algorithmic}
    \end{algorithm}
    In~\cref{coro:relathinship_ub_soft_sorted_weak}, $Z_k$ is the normalization constant when assuming $p_1=b_1, \ldots, p_k=b_k$, $p_{k+1}\propto \exp(g_{k+1}), \ldots, p_{K}\propto \exp(g_K)$.
    Note that~\cref{coro:relathinship_ub_soft_sorted_weak} also holds for the edge case $\sum_{k=1}^K b_k = 1$: in this case, $s_{K-1} > 0$ and $p(K-1) = \bm{b}_{1:K-1} \concat s_{K-1} \cdot \exp(g_K) / \exp(g_K) = \bm{b}$.
    By~\cref{coro:relathinship_ub_soft_sorted_weak}, we can derive the $O(K^2)$ time computation algorithm for $\UBSoftmax$, as shown in~\cref{alg:ubsoftmax_square}.
    
    We investigate additional properties of $\UBSoftmax_1(\bm{g}, \bm{b})$ to derive a more efficient computation algorithm.
    Then, the next lemma characterizes the threshold index $\rho$ in~\cref{coro:relathinship_ub_soft_sorted_weak}.
    \begin{lemmamd}
        Let $\bm{g}, \bm{b}, \bm{p}$, $\rho$, and $Z_k$ be defined in~\cref{coro:relathinship_ub_soft_sorted_weak}.
        Then,
        \begin{align}
            \forall k \in \{0, \ldots, K-1 \mid Z_k > 0\}, \quad \frac{\exp(g_{k+1})}{Z_k} &> b_{k+1} \implies \rho \neq k.
        \end{align}
        \label{lemma:condition_of_rho}
    \end{lemmamd}
    \begin{proof}
        We prove it by contradiction.
        Assume that $\rho=k$.
        Then, by~\cref{thm:relationship_ub_soft} and~\cref{coro:relathinship_ub_soft_sorted_weak}, $p_{k+1}=\exp(g_{k+1})/Z_k \le b_{k+1}$, which contradicts the assumption $\exp(g_{k+1})/Z_k > b_{k+1}$.
    \end{proof}
    Thus,~\cref{lemma:condition_of_rho} states that $p_{k+1}=\exp(g_{k+1})/Z_k$ is less than or equal to $b_{k+1}$ if we assume that $\rho=k$.
    
    By~\cref{lemma:condition_of_rho} and the following lemma, we obtain the following \emph{monotone} candidate set of $\rho$:
    \begin{align}
        \mathsf{Cand}(\bm{g}, \bm{b}) \coloneqq \left\{k \in \{0, \ldots, K-1\} \mid Z_k > 0, \frac{\exp (g_{k+1})}{Z_k} \le b_{k+1}\right\}.
    \end{align}
    \begin{lemmamd}
        Let $\bm{g}, \bm{b}$, and $Z_k$ be defined in~\cref{lemma:condition_of_rho}.
        Then,
        \begin{align}
             \forall k \in \{0, \ldots, K-1 \mid Z_k > 0\},\quad \frac{\exp(g_{k+1})}{Z_k} \le b_{k+1} \implies \frac{\exp(g_{i+1})}{Z_k} \le b_{i+1} \ \forall i > k.
        \end{align}
        Therefore, if $\exp(g_{k+1})/Z_k \le b_{k+1}$, $p(k)$ is a feasible (but not necessarily optimal) solution to the optimization problem of the $\UBSoftmax$ function, that is, $\bm{0}_K \preceq p(k) \preceq \bm{b}$ and $p(k) \in \Delta^K$.

        Moreover, define $K^\prime\coloneqq \max \{k: Z_k > 0, \exp(g_{k+1})/Z_k \le b_{k+1}\}$.
        Then, for $k \in \{0, \ldots, K^\prime-1\}$, $k \in \mathsf{Cand}(\bm{g}, \bm{b}) \implies i \in \mathsf{Cand}(\bm{g}, \bm{b})$ for all $i \in \{k+1, \ldots, K^\prime\}$, that is, there exists $k_0 \in\{0, \ldots, K^\prime\}$ such that $\mathsf{Cand}(\bm{g}, \bm{b}) = \{k_0, \ldots, K^\prime\}$.
        \label{lemma:feasibility_of_p_of_k}
    \end{lemmamd}
    \begin{proof}
        Since $\bm{g}$ and $\bm{b}$ are sorted as $b_1/\exp(g_1) \le \cdots \le b_K / \exp(g_K)$, we have
        \begin{align}
            \frac{b_{k+1}}{\exp(g_{k+1})} \le \frac{b_{i+1}}{\exp (g_{i+1})} \ \forall i > k.
        \end{align}
        From this inequality and the sorting assumption, we have
        \begin{align}
            \frac{1}{Z_k} \le \frac{b_{k+1}}{\exp(g_{k+1})} \le \frac{b_{i+1}}{\exp (g_{i+1})} \implies \frac{\exp(g_{i+1})}{Z_k} \le b_{i+1},
        \end{align}
        which concludes the proof of the first part.

        We next prove the second part of the lemma.
        Assume that $k \in \{0, \ldots, K^\prime-1\}$ and $k \in \mathsf{Cand}(\bm{g}, \bm{b})$.
        It is sufficient to prove that $k+1\in \mathsf{Cand}(\bm{g},\bm{b})$, i.e., $Z_{k+1} > 0$ and $\exp(g_{k+2})/Z_{k+1} \le b_{k+2}$.
        For all $i \in \{k+1, \ldots, K^\prime\}$,
        \begin{align}
            r_k\ge r_i \ge r_{K^\prime} > 0 \quad \text{and} \quad s_k\ge s_i \ge s_{K^\prime} > 0 \implies Z_{i} > 0,
        \end{align}
        thus we have
        \begin{align}
            Z_{k+1} > 0.
        \end{align}
        By the definitions of $r_{k}$ and $s_{k}$, we have
        \begin{align}
            r_k = \exp(g_{k+1}) + r_{k+1}, \quad s_k = b_{k+1} + s_{k+1}
        \end{align}
        for all $k \in \{0, \ldots, K-1\}$.
        Since $k\in \mathsf{Cand}(\bm{g}, \bm{b})$, we obtain
        \begin{align}
            &\frac{\exp(g_{k+1})}{Z_k} \le b_{k+1} \\
            \implies &\exp(g_{k+1})(b_{k+1}+s_{k+1}) \le b_{k+1}(\exp(g_{k+1})+r_{k+1})\\
            \implies &\exp(g_{k+1})s_{k+1} \le b_{k+1}r_{k+1} \\
            \implies &\frac{1}{Z_{k+1}} \le \frac{b_{k+1}}{\exp(g_{k+1})}.
        \end{align}
        From this inequality and the sorting assumption, we have
        \begin{align}
            \frac{1}{Z_{k+1}} \le \frac{b_{k+1}}{\exp(g_{k+1})} \le \frac{b_{k+2}}{\exp(g_{k+2})}\implies \frac{\exp(g_{k+2})}{Z_{k+1}} \le b_{k+2}.
        \end{align}
        This concludes the proof of the second part.
    \end{proof}
    
    The next lemma states that $\rho$ is the minimum value in $\mathsf{Cand}$; thus enabling us to efficiently determine $\rho$ without computing each $p(k)$ and verifying whether $p(k) \in \Delta^K \cap [\bm{0}_K, \bm{b}]$ for all $k \in \{0, \ldots, K\}$. 
    \begin{lemmamd}
        Let $\bm{g}, \bm{b}, \bm{p}, s_k, p(k)$, and $\rho$ be defined in~\cref{coro:relathinship_ub_soft_sorted_weak}.
        Then,
        \begin{align}
            \rho = \min \mathsf{Cand}(\bm{g}, \bm{b}).
        \end{align}
        \label{lemma:rho_is_min_of_cand}
    \end{lemmamd}
    \begin{proof}
        By~\cref{coro:relathinship_ub_soft_sorted_weak},~\cref{lemma:condition_of_rho}, and~\cref{lemma:feasibility_of_p_of_k},
        $\bm{p}=\UBSoftmax_1(\bm{g}, \bm{b}) \in \{p(k)\mid k\in\mathsf{Cand}(\bm{g}, \bm{b})\}$.
        Moreover, by the definition of $\UBSoftmax$, we have
        \begin{align}
            \bm{p} = \argmax_{\bm{p}' \in \{p(k)\mid k \in \mathsf{Cand}(\bm{g}, \bm{b})\}} \bm{g}^\top\bm{p}^\prime + H(\bm{p}^\prime)
        \end{align}
        On the other hand, by~\cref{lemma:bcsoftmax_fixed}, for all $k\in \mathsf{Cand}(\bm{g}, \bm{b})$, $p(k)$ is the solution to the following optimization problem:
        \begin{align}
            p(k) &= \bm{b}_{1:k} \concat s_k\cdot \Softmax_1(\bm{g}_{k+1:K}) \\
            & = \bm{b}_{1:k} \concat s_k\cdot \BCSoftmax_1(\bm{g}_{k+1:K}; (\bm{0}_{K-k}, \bm{1}_{K-k}))\\
            &= \argmax_{\bm{p} \in P(k)} \bm{g}^\top\bm{p} + H(\bm{p}), \\
            \text{where} \quad & P(k) \coloneqq \{\bm{q} \in \Delta^K \mid q_1=b_1, \ldots, q_k=b_k\}.
        \end{align}
        Since $P(k+1) \subset P(k)$, we have
        \begin{align}
            \bm{g}^\top p(k) + H(p(k)) \ge \bm{g}^\top p(k+1) + H(p(k+1)).
        \end{align}
        Therefore, $\bm{p}= p(\min \mathsf{Cand}(\bm{g}, \bm{b})) \implies \rho=\min \mathsf{Cand}(\bm{g}, \bm{b})$.
    \end{proof}

    We conclude this section with the proof of~\cref{thm:relathinship_ub_soft_sorted}.
    \begin{proof}
        \Cref{thm:relathinship_ub_soft_sorted} follows directly from~\cref{coro:relathinship_ub_soft_sorted_weak} and~\cref{lemma:rho_is_min_of_cand}.
    \end{proof}

\subsection{Proof of \texorpdfstring{\cref{thm:relationship_bc_soft}}{Theorem \ref{thm:relationship_bc_soft}}}
    \begin{proof}
        The proof of~\cref{thm:relationship_bc_soft} is similar to that of~\cref{thm:relationship_ub_soft}.
        The Lagrangian of the optimization problem in~\cref{eq:bcsoftmax} is
        \begin{align}
            L(\bm{p}, \lambda, \bm{\alpha}, \bm{\beta}) = \bm{g}^\top \bm{p} - \sum_{k=1}^K p_k\log p_k - \lambda \left(\sum_{k=1}^K p_k - 1\right) - \alpha^\top(\bm{a}-\bm{p})-\bm{\beta}^\top (\bm{p}-\bm{b}),
            \label{eq:bcsoft_lagrangian}
        \end{align}
        where $\lambda \in \real$ and $\bm{\alpha}, \bm{\beta} \in \real^K_{\ge 0}$ are Lagrange multipliers.
        We obtain the optimality condition by differentiating~\cref{eq:bcsoft_lagrangian} with respect to $p_k$:
        \begin{align}
            \frac{\partial L}{\partial p_k} = g_k - (1 + \log p_k) - \lambda + \alpha_k - \beta_k = 0 \\
            \implies p_k = \exp(g_k + \alpha_k - \beta_k) / Z, \quad \text{where} \ Z = \exp(\lambda + 1).
        \end{align}
        From the summation condition $\sum_k p_k=1$, $Z = \sum_{k} \exp (g_k + \alpha_k - \beta_k)$, thus, by defining $\bm{\gamma} \coloneqq -\bm{\alpha}+\bm{\beta}$,~\cref{eq:relationship_bc_soft} holds:
        \begin{align}
            p_k = \frac{\exp(g_k + \alpha_k - \beta_k)}{\sum_{i=1}^K \exp(g_i +\alpha_i- \beta_i)} = \Softmax_1(\bm{g}-\bm{\gamma})[k].
        \end{align}
        By the complementary slackness KKT condition, we have
        \begin{align}
            \alpha_k (a_k - p_k) = 0 &\implies \alpha_k = 0 \quad \text{or} \quad p_k = a_k,\\
            \beta_k (p_k - b_k) = 0 &\implies \beta_k = 0 \quad \text{or} \quad p_k = b_k,
        \end{align}
        and this implies~\cref{eq:bc_gamma_indices}.
        
        The normalization term $Z$ is derived by a simple calculation:
        \begin{align}
            \sum_{i} p_i &= \sum_{i: \gamma_i < 0} \exp(g_i - \gamma_i)/Z + \sum_{i: \gamma_i > 0} \exp(g_i - \gamma_i)/Z + \sum_{i: \gamma_i = 0}\exp(g_i)/Z \\
            &= \sum_{i: \gamma_i < 0} a_i + \sum_{i: \gamma_i > 0} b_i + \sum_{i: \gamma_i = 0}\exp(g_i)/Z = 1\\
            \implies Z &= \frac{\sum_{i:\gamma_i = 0} \exp(g_i)}{1-\sum_{i: \gamma_i<0}a_i-\sum_{i: \gamma_i>0}b_i} \coloneqq \frac{r}{s}.
        \end{align}
        Strictly speaking, $Z$ is not defined when $s=0$.
        However, $s=0$ means $\gamma_k \neq 0$ and $p_k = b_k$ or $p_k=a_k$ for all $k \in [K]$.
        Therefore, $Z$ does not appear in~\cref{eq:ub_beta_indices} in this case.
    \end{proof}
    
\subsection{Proof of \texorpdfstring{\cref{thm:relathinship_bc_soft_sorted}}{Theorem \ref{thm:relathinship_bc_soft_sorted}}}
    For the proof of~\cref{thm:relathinship_bc_soft_sorted}, we provide additional lemmas.

     \begin{lemmamd}
        Let $\bm{g} \in \real^K$ be a logit vector and $(\bm{a}, \bm{b}) \in B^K$ be a pair of lower- and upper-bound vectors.
        Define $\bm{p}=\BCSoftmax_1(\bm{g}; (\bm{a}, \bm{b}))$ and $V_a = \{k \in [K]\mid p_k = a_k\}$. 
        If $i \in V_a$ and $a_j / \exp(g_j) \ge a_i /\exp(g_i)$, then $j \in V_a$.
        \label{lemma:ratio_of_lower_bound_and_exp_logit}
    \end{lemmamd}
    \begin{proof}
        By~\cref{thm:relationship_bc_soft}, there exist $\gamma_i, \gamma_j \in \real$ such that $p_i = \exp(g_i - \gamma_i) / Z \ge a_i$ and $p_j = \exp(g_j - \gamma_j) / Z \ge a_j$, where $Z > 0$ is the normalization term.
        In addition, if $i \in V_a$ and $a_j / \exp(g_j) \ge a_i /\exp(g_i)$, we have
        \begin{align}
            i \in V_a &\implies \gamma_i < 0 \quad \text{and}\\
            \frac{a_j}{\exp(g_j)}\ge \frac{a_i}{\exp(g_i)} &\implies a_j \ge a_i \frac{\exp(g_j)}{\exp(g_i)}.
        \end{align}
        Then, we obtain the following inequality, which implies $j \in V_a$.
        \begin{align}
            \frac{\exp(g_j-\gamma_j)}{Z}\ge a_j \ge a_i \frac{\exp(g_j)}{\exp(g_i)} &\implies \exp(-\gamma_j) \ge a_i \frac{Z}{\exp(g_i)}  = \exp(-\gamma_i) > 1\\
            &\implies \gamma_j \le \gamma_i < 0.
        \end{align}
    \end{proof}
    
    Then, by~\cref{lemma:ratio_of_lower_bound_and_exp_logit},~\cref{thm:relationship_bc_soft}, and~\cref{lemma:bcsoftmax_fixed}, we immediately obtain the following corollary.
    \begin{corollarymd}
        For all $\bm{g} \in \real^K$ and $(\bm{a}, \bm{b}) \in B^K$ such that $a_1/\exp(g_1) \ge \cdots \ge a_K / \exp(g_K)$, there exists $0 \le \rho_a \le K$ such that
        \begin{align}
            k \le \rho_a &\implies p_k = a_k, \\
            k > \rho_a &\implies p_k > a_k \implies p_k \propto \exp(g_k) \quad \text{or}\quad  p_k = b_k,
        \end{align}
        where $\bm{p} = \BCSoftmax_1(\bm{g}; (\bm{a}, \bm{b}))$.
        Therefore,
        \begin{align}
            \bm{p} \in & \left\{ p(k):\bm{a}  \preceq p(k) \preceq \bm{b}, p(k) \in \Delta^K, k \in \{0, \ldots, K^\prime\}\right\}, \quad \text{where}\\
            \label{eq:cand_of_p_bc}
            p(k) &\coloneqq \argmax_{\bm{p}' \in \Delta^K, \bm{a} \preceq \bm{p}' \preceq \bm{b}, p^{\prime}_1=a_1, \ldots, p^{\prime}_k=a_k} \bm{g}^\top \bm{p}' - \sum_{i}p^{\prime}_i \log p^{\prime}_i\\
            &= \bm{a}_{1:k} \concat s_k\cdot \UBSoftmax_1(\bm{g}_{k+1:K}, \bm{b}_{k+1:K}/s_k),\\
             s_k \coloneqq& 1 - \sum_{i=1}^{k}a_i, \ s_0 \coloneqq 1, K^\prime \coloneqq \max\left\{k \mid \sum_{i=k+1}^{K}b_i \ge s_k \right\}.
        \end{align}
        \label{coro:relathinship_bc_soft_sorted_weak}
    \end{corollarymd}

    We next extend~\cref{lemma:feasibility_of_p_of_k,lemma:rho_is_min_of_cand} to $\BCSoftmax$.
    Note that the second part of the following lemma guarantees the correctness of~\cref{alg:bcsoftmax}.
    Since $\mathsf{Cand}_{a}(\bm{g}, (\bm{a}, \bm{b}))$ is monotone, $\rho_a$ can be found by a quickselect-like procedure.
    \begin{lemmamd}
        Let $\mathsf{Cand}_a(\bm{g}, (\bm{a}, \bm{b})) \subset \{0, \ldots, K\}$ be defined as
        \begin{align}
            \mathsf{Cand}_a(\bm{g}, (\bm{a}, \bm{b})) \coloneqq \{k:\bm{a}  \preceq p(k) \preceq \bm{b}, p(k) \in \Delta^K, k \in \{0, \ldots, K^\prime\}\},
        \end{align}
        where $p(k)$ and $K^\prime$ are defined in~\cref{coro:relathinship_bc_soft_sorted_weak}.
        Then,
        \begin{align}
            \rho_a = \min \mathsf{Cand}_a(\bm{g}, (\bm{a}, \bm{b})).
        \end{align}

        Moreover, for $k \in \{0, \ldots, K^\prime-1\}$, $k \in \mathsf{Cand}_a(\bm{g}, (\bm{a},\bm{b})) \implies i \in \mathsf{Cand}_a(\bm{g}, (\bm{a},\bm{b}))$ for all $i \in \{k+1, \ldots, K^\prime\}$, that is, there exists $k_0 \in\{0, \ldots, K^\prime\}$ such that $\mathsf{Cand}_a(\bm{g}, (\bm{a}, \bm{b})) = \{k_0, \ldots, K^\prime\}$.
    \label{lemma:rho_a_is_min_of_cand}
    \end{lemmamd}
    \begin{proof}
        For all $k\in \mathsf{Cand}_a(\bm{g}, (\bm{a}, \bm{b}))$, $p(k)$ is the solution to the following optimization problem:
        \begin{align}
            p(k) &= \argmax_{\bm{p} \in P(k) } \bm{g}^\top\bm{p} - \sum_{i=1}^K p_i \log p_i, \\
            \text{s.t.} \quad \bm{p} &\in P(k) \coloneqq \{\bm{q} \in \Delta^K \mid q_1=a_1, \ldots, q_k=a_k, a_{k+1} \le q_{k+1} \le b_{k+1}, \ldots, a_K \le q_K \le b_K\}.
        \end{align}
        Since $P(k+1) \subset P(k)$, we have
        \begin{align}
            \bm{g}^\top p(k) + H(p(k)) \ge \bm{g}^\top p(k+1) + H(p(k+1))
        \end{align}
        Therefore, $\rho_a=\min \mathsf{Cand}_a(\bm{g}, (\bm{a}, \bm{b}))$.

        We next prove the second part.
        Assume that $k \in \mathsf{Cand}_a(\bm{g},(\bm{a}, \bm{b}))$, namely, $\bm{a} \preceq p(k) \preceq \bm{b}$ for some $k \in \{0, \ldots, K^\prime-1\}$.
        It is sufficient to prove that $\bm{a}\preceq p(k+1)\preceq\bm{b}$.
        For all $j \in \{0, \ldots, K^\prime\}$, we have
        \begin{align}
            &\sum_{i=j+1}^K b_i - s_{j}  = \sum_{i=K^\prime+1}^K b_i + \sum_{i=j+1}^{K^\prime} b_i - s_{K^\prime} - \sum_{i=j+1}^{K^\prime}a_i \\
            =& \sum_{i=K^\prime+1}^K b_i - s_{K^\prime} + \sum_{i=j+1}^{K^\prime}(b_i-a_i)
            \ge \sum_{i=K^\prime+1}^K b_i - s_{K^\prime} \ge 0
            \\\implies& p(j) \preceq \bm{b}.
        \end{align}
        This indicates that $p(k+1) \preceq \bm{b}$.
 Next, we show that $\bm{a}\preceq p(k+1)$.
        By~\cref{thm:relationship_bc_soft} and the definition of $p(k)$, for each $k\in \{0,\ldots,K^\prime\}$, there exists $Z_k > 0$ such that
        \begin{align}
            p_{i}(k) = \min \left(b_i, \frac{\exp(g_i)}{Z_k}\right) \quad \forall i > k.
        \end{align}
        Define $q_{k+1} \coloneqq p_{k+1}(k)$.
        Then, we can rewrite $p(k)$ as
        \begin{align}
            p(k) &\coloneqq \argmax_{\bm{p}' \in \Delta^K, \bm{a} \preceq \bm{p}' \preceq \bm{b}, p^{\prime}_1=a_1, \ldots, p^{\prime}_k=a_k} \bm{g}^\top \bm{p}' - \sum_{i}p^{\prime}_i \log p^{\prime}_i\\
            &= \argmax_{\bm{p}' \in \Delta^K, \bm{a} \preceq \bm{p}' \preceq \bm{b}, p^{\prime}_1=a_1, \ldots, p^{\prime}_k=a_k, p^\prime_{k+1}=q_{k+1}} \bm{g}^\top \bm{p}' - \sum_{i}p^{\prime}_i \log p^{\prime}_i.
        \end{align}
        By this equation and~\cref{thm:relationship_bc_soft}, we have $Z_{k+1}\le Z_{k}$, which implies
        \begin{align}
            p_{i}(k+1) \ge p_{i}(k) \quad \text{for all } i > k+1.
        \end{align}
        Since we assume that $\bm{a}\preceq p(k)$, we have $\bm{a}_{k+2:K}\preceq p_{k+2:K}(k+1)$.
        By the definition of $p(k+1)$, we have $\bm{a}_{1:k+1}=p_{1:k+1}(k+1)$.
        These imply $\bm{a} \preceq p(k+1)$, which concludes the proof.
    \end{proof}
    
    We conclude this section with the proof of~\cref{thm:relathinship_bc_soft_sorted}.
    \begin{proof}
        \Cref{thm:relathinship_bc_soft_sorted} follows directly from~\cref{coro:relathinship_bc_soft_sorted_weak} and~\cref{lemma:rho_a_is_min_of_cand}.
    \end{proof}
\subsection{Proof of \texorpdfstring{\cref{prop:pb_preserves_accuracy}}{Proposition \ref{prop:pb_preserves_accuracy}}}
    \begin{proof}
        Without loss of generality, we may assume that $f(\bm{g})[1] \le \cdots \le  f(\bm{g})[K]$.
        Since the lower and upper bounds are the same in all classes and the lower bound $a < 1/K$, by~\cref{thm:relathinship_bc_soft_sorted}, we have
        \begin{align}
            \exists\rho_a, \rho_b \quad \text{such that}\quad 0 \le \rho_a < \rho_b \le K+1, \quad f_{\mathrm{PB}}(\bm{g})[i] &= \begin{cases}
                a & i \le \rho_a\\
                \frac{\exp(g_i/\tau)}{Z} & \rho_a < i < \rho_b \\
                b & \rho_b \le i
            \end{cases}, 
            \label{eq:pb_indices}\\
            \text{where}\quad Z = r/s, r = \sum_{\rho_a < i < \rho_b} \exp(g_i/\tau), s = 1-\rho_a\cdot a &- (K+1-\rho_b)\cdot b.
        \end{align}
        Since $a < f_{\mathrm{PB}}(\bm{g})[i] < b$ for all $\rho_a < i < \rho_b$, we have
        \begin{align}
            f_{\mathrm{PB}}(\bm{g})[1] \le \cdots \le f_{\mathrm{PB}}(\bm{g})[K],
        \end{align}
        namely, $f_{\mathrm{PB}}$ preserves the order (ranking) of the original probabilities, and this indicates that $\argmax f(\bm{g}) \subseteq \argmax f_{\mathrm{PB}}(\bm{g})$.
        This concludes the first part of~\cref{prop:pb_preserves_accuracy}.

        We next prove the second part: if $b > 1/(k+1)$, then PB exactly preserves the top-$k$ predictions.
        For simplicity, we additionally assume that $g_{K-k} < g_{K-k+1}$, that is, the indices of the top-$k$ predictions of $f(\bm{g})$ are unique:
        \begin{align}
            \{K-k+1, K-k+2, \ldots, K-1, K\}
        \end{align}
        Since $b > 1 / (1+k)$, the number of upper-bounded samples, $n = (K-\rho_b+1)$, is less than or equal to $k$, indicating that
        \begin{align}
            K - \rho_b +1 \le k \implies K - k+1 \le \rho_b.
        \end{align}
        Therefore, the $(k+1)$-th largest probability of $f_{\mathrm{PB}}(\bm{g})$ is not upper-bounded.
        Combining the assumption $g_{K-k} < g_{K-k+1}$, we have
        \begin{align}
            f_{\mathrm{PB}}(\bm{g})[K-k] < f_{\mathrm{PB}}(\bm{g})[K-k+1] \le \cdots \le f_{\mathrm{PB}}(\bm{g})[K] \le b,
        \end{align}
       so the top-$k$ indices, i.e., top-$k$ predictions, are preserved.
    \end{proof}

\subsection{Proof of \texorpdfstring{\cref{thm:when_does_pb_work}}{Theorem \ref{thm:when_does_pb_work}}}
\label{sec:proof_of_when_does_pb_work}
\subsubsection{Preliminary}
    We assume that $K=2$ and $\mathcal{Y}=\{0,1\}$, that is, we consider the binary classification problem.
    When $K=2$, $\Softmax$ can be reduced to the sigmoid function $\sigma$:
    \begin{align}
        \Softmax_1(\bm{g})[1] = \frac{\exp (g_1)}{\exp(g_1) + \exp(g_2)} = \frac{\exp(g_1-g_2)}{1+\exp(g_1-g_2)} = \sigma(g_1-g_2).
    \end{align}
    Therefore, we can assume that the logit function $g$ outputs a scalar, i.e., $g: \mathcal{X}\to\real$ and $f=\sigma \circ g$ outputs the probability of $y=1$.
    
    When $K=2$, $\BCSoftmax$ can be written as the sigmoid function with clipping.
    \begin{corollarymd}
        For all $\bm{g} \in \real^K$ and $(\bm{a}, \bm{b}) \in B^K$,
        \begin{align}
            \BCSoftmax_1(\bm{g}; (\bm{a}, \bm{b}))[1] &= \clip(\sigma(g_1-g_2), \tilde{a}(a,b), \tilde{b}(a,b)) \\
            &= \sigma(\clip(g_1-g_2, \tilde{c}(a,b), \tilde{C}(a,b)))
        \end{align}
        where
        \begin{align}
            \tilde{a}(a,b) &= \max \{a, 1-b\}, \ &\tilde{b}(a,b) = \min\{1-a, b\}=1-\tilde{a}(a,b), \\
            \tilde{c}(a,b) &= \sigma^{-1}(\tilde{a}),\ &\tilde{C}(a,b) = \sigma^{-1}(\tilde{b})=-\tilde{c}(a,b).
        \end{align}
        \label{coro:bcsoftmax_is_clipped_sigmoid}
    \end{corollarymd}
    
\subsubsection{Proof of \texorpdfstring{\cref{thm:when_does_pb_work}}{Theorem~\ref{thm:when_does_pb_work}}}
    \begin{proof}[Proof of~\cref{thm:when_does_pb_work}]
        By~\cref{coro:bcsoftmax_is_clipped_sigmoid}, we can write $f$ and $f_{\mathrm{PB}}$ as
        \begin{align}
            f(\bm{x}) &= \sigma(g(\bm{x})), \\
            f_{\mathrm{PB}}(\bm{x}; a, b) &= \clip(f(\bm{x}), \tilde{a}(a,b), \tilde{b}(a,b)).
        \end{align}
        We assume that $f$ is uniformly underconfident with the parameters $(a_0, \varepsilon_{\mathrm{UC}})$ and uniformly overconfident with the parameters $(b_0, \varepsilon_{\mathrm{OC}})$.
        Define $a^*$ and $b^*$ as
        \begin{align}
            a^* \coloneqq \min(a_0, 1-b_0, \varepsilon_{\mathrm{UC}}, \varepsilon_{\mathrm{OC}}),\,
            b^*\coloneqq 1-a^*,
        \end{align}
        and fix any $a\le a^*$ and $b \ge b^*$.
        Then, because $a \le a^* \le \min(a_0, 1-b_0)$ and $b \ge b^* \ge 1-\min(a_0, 1-b_0) \implies \min(a_0, 1-b_0) \ge 1-b$, we have
        \begin{align}
            \tilde{a}(a,b) &= \max(a, 1-b) \le \max(\min(a_0, 1-b_0), \min(a_0, 1-b_0)) = \min(a_0, 1-b_0) \\
            &\le a_0, \\
            \tilde{b}(a, b) &= 1 - \max(a, 1-b) \ge 1 - \max(\min(a_0, 1-b_0), \min(a_0, 1-b_0)) = 1 - \min(a_0, 1-b_0) \\
            &\ge  b_0.
        \end{align}
        For simplicity, we denote $\tilde{a}(a,b)$ and $\tilde{b}(a,b)$ as $\tilde{a}$ and $\tilde{b}$.
        We partition the input space $\mathcal{X}$ into the following three regions:
        \begin{align}
            A &\coloneqq \{\bm{x}\mid f(\bm{x}) \le \tilde{a}\} = \{\bm{x} \mid f_{\mathrm{PB}}(\bm{x}) = \tilde{a}\}, \\
            M &\coloneqq \{\bm{x}\mid \tilde{a}<f(\bm{x})<\tilde{b}\} = \{\bm{x}\mid \tilde{a}<f_{\mathrm{PB}}(\bm{x})<\tilde{b}\},\\
            B &\coloneqq \{\bm{x}\mid f(\bm{x}) \ge \tilde{b}\} =  \{\bm{x}\mid f_{\mathrm{PB}}(\bm{x}) = \tilde{b}\}.
        \end{align}
        
        \paragraph{Proof of the $\TCE$ inequality.}
        Then, $\TCE$ for $f_{\mathrm{PB}}$ can be written as
        \begin{align}
            \TCE_{\mathcal{D}}(f_{\mathrm{PB}}) &= \mathbb{E}_X[\lvert f_{\mathrm{PB}}(X) -\mathbb{P}(Y=1 \mid f_{\mathrm{PB}}(X))\rvert] \\
            \begin{split}
                &= P(A)\mathbb{E}_X[\lvert f_{\mathrm{PB}}(X) -\mathbb{P}(Y=1 \mid f_{\mathrm{PB}}(X))\rvert \mid A] \\
                &\quad +P(M)\mathbb{E}_X[\lvert f_{\mathrm{PB}}(X) - \mathbb{P}(Y=1 \mid f_{\mathrm{PB}}(X))\rvert \mid M] \\
                &\quad +P(B)\mathbb{E}_X[\lvert f_{\mathrm{PB}}(X) -\mathbb{P}(Y=1 \mid f_{\mathrm{PB}}(X))\rvert \mid B].
                \label{eq:tce_decomposition}
            \end{split}
        \end{align}
        We first consider the expectation in the first term, $\mathbb{E}_X[\lvert f_{\mathrm{PB}}(X) - \mathbb{P}(Y=1 \mid f_{\mathrm{PB}}(X))\rvert \mid A]$.
        For all $\bm{x} \in A$, $f(\bm{x}) \le f_{\mathrm{PB}}(\bm{x}) = \tilde{a} \le a_0$.
        Since $f$ is uniformly underconfident on a low probability region with the parameters $a_0$ and $\varepsilon_{\mathrm{UC}}$, we obtain
        \begin{align}
            \mathbb{P}(Y=1\mid f(X)) &\ge f(X) + \varepsilon_{\mathrm{UC}} \ge \varepsilon_{\mathrm{UC}} \ge \tilde{a}
        \end{align}
        almost surely on $A$.
        This implies the following inequality:
        \begin{align}
            \mathbb{E}_{X}[\mathbb{P}(Y=1 \mid f(X)) \mid A] \ge \tilde{a}.
        \end{align}
        Therefore, since $f_{\mathrm{PB}}(\bm{x})=\tilde{a}$ (i.e., constant) on $A$, we have
        \begin{align}
            \mathbb{E}_X[\lvert f_{\mathrm{PB}}(X) - \mathbb{P}(Y=1 \mid f_{\mathrm{PB}}(X))\rvert \mid A] &= \lvert \tilde{a}- \mathbb{E}_{X}[\mathbb{P}(Y=1 \mid f(X)) \mid A] \rvert \\
            &= \mathbb{E}_{X}[\mathbb{P}(Y=1 \mid f(X)) \mid A] - \tilde{a},
            \label{eq:conditional_tce_on_A_pb}
        \end{align}
        where the last equality uses $\mathbb{E}_X[\mathbb{P}(Y=1 \mid f(X)) \mid A] \ge \tilde a$.
        We next consider the case for the uncalibrated baseline model $f$.
        Since $f$ is uniformly underconfident on $A$, its corresponding expectation term can be written as
        \begin{align}
            \mathbb{E}_X[\lvert f(X) - \mathbb{P}(Y=1 \mid f(X))\rvert \mid A] &= \mathbb{E}_X[\mathbb{P}(Y=1 \mid f(X)) - f(X) \mid A].
        \end{align}
        Since $f(\bm{x}) \le \tilde{a} \to \tilde{a}-f(\bm{x}) \ge 0$ for all $\bm{x}\in A$, we have
        \begin{align}
            \mathbb{E}_X[\mathbb{P}(Y=1 \mid f(X)) - f(X) \mid A] &= \mathbb{E}_X[\mathbb{P}(Y=1 \mid f(X)) -\tilde{a} + \tilde{a} - f(X) \mid A] \\
            &\ge \mathbb{E}_X[\mathbb{P}(Y=1 \mid f(X)) -\tilde{a} + 0 \mid A] \label{eq:conditional_tce_on_A_inequality_tmp}\\
            &=\mathbb{E}_{X}[\mathbb{P}(Y=1 \mid f(X)) \mid A] - \tilde{a} \tag{\ref{eq:conditional_tce_on_A_pb}}.
        \end{align}
        Therefore, we have
        \begin{align}
            P(A)\mathbb{E}_X[\lvert f_{\mathrm{PB}}(X) -\mathbb{P}(Y=1 \mid f_{\mathrm{PB}}(X))\rvert \mid A] \le P(A)\mathbb{E}_X[\lvert f(X) -\mathbb{P}(Y=1 \mid f(X))\rvert \mid A].
            \label{eq:conditional_tce_on_A_inequality}
        \end{align}
        Similarly, we can obtain
        \begin{align}
            P(B)\mathbb{E}_X[\lvert f_{\mathrm{PB}}(X) -\mathbb{P}(Y=1 \mid f_{\mathrm{PB}}(X))\rvert \mid B] \le P(B)\mathbb{E}_X[\lvert f(X) -\mathbb{P}(Y=1 \mid f(X))\rvert \mid B].
            \label{eq:conditional_tce_on_B_inequality}
        \end{align}
        Since $f_{\mathrm{PB}}(\bm{x}) =f(\bm{x})$ on $M$, we have
        \begin{align}
             P(M)\mathbb{E}_X[\lvert f_{\mathrm{PB}}(X) -\mathbb{P}(Y=1 \mid f_{\mathrm{PB}}(X))\rvert \mid M] = P(M)\mathbb{E}_X[\lvert f(X) -\mathbb{P}(Y=1 \mid f(X))\rvert \mid M].
        \end{align}
        \Cref{eq:pb_tce_is_not_worse_than_baseline} follows immediately from these inequalities.
        If the lower bound is non-vacuous, i.e., $\mathbb{P}(f(X) < \tilde{a}) > 0$,~\cref{eq:conditional_tce_on_A_inequality_tmp} is strict and consequently~\cref{eq:conditional_tce_on_A_inequality} is also strict.
        Similarly, if the upper bound is non-vacuous,~\cref{eq:conditional_tce_on_B_inequality} is strict.
        This implies the overall inequality~\eqref{eq:pb_tce_is_not_worse_than_baseline} is strict if at least one of the bounds is non-vacuous.

        \paragraph{Proof of the risk inequality.}
        Assume that loss $\ell: \{0,1\}\times[0,1]\to\mathbb{R} \cup \{+\infty\}$ is a strictly proper convex loss~\citep{buja2005loss}: for every $q\in[0,1]$,
        \begin{align}
            L_{\ell}(q,p)\coloneqq q\,\ell(1,p)+(1-q)\,\ell(0,p)
        \end{align}
        is strictly convex and minimized at $p=q$.
        We prove that
        \begin{align}
                R_{\ell,\mathcal{D}}\bigl(f_{\mathrm{PB}}(\cdot; a,b)\bigr) \le  R_{\ell,\mathcal{D}}(f) \quad \forall a\le a^*, b\ge b^*,
            \tag{\ref{eq:pb_risk_is_not_worse_than_baseline}}
        \end{align}
        where $R_{\ell,\mathcal{D}}(h)\coloneqq \mathbb{E}_{(X,Y)\sim\mathcal{D}}[\ell(Y,h(X))]$ is the expected $\ell$-risk.
        As with~\cref{eq:tce_decomposition}, we can decompose $R_{\ell, \mathcal{D}}$ as
        \begin{align}
            R_{\ell, \mathcal{D}}(f_{\mathrm{PB}}) &= \mathbb{E}_{X,Y}[\ell(Y, f_{\mathrm{PB}}(X))] \\
            &= \mathbb{E}_{X}[L_\ell(\mathbb{P}(Y=1\mid f(X)), f_{\mathrm{PB}}(X))] \\
            &= P(A)\mathbb{E}_{X}[L_\ell(\mathbb{P}(Y=1\mid f(X)), f_{\mathrm{PB}}(X)) \mid A] \\
            &\quad+ P(M)\mathbb{E}_{X}[L_\ell(\mathbb{P}(Y=1\mid f(X)), f_{\mathrm{PB}}(X)) \mid M] \\
            &\quad + P(B)\mathbb{E}_{X}[L_\ell(\mathbb{P}(Y=1\mid f(X)), f_{\mathrm{PB}}(X)) \mid B],
        \end{align}
        where we used the law of total expectation for the second equality.
        Therefore, to prove~\cref{eq:pb_risk_is_not_worse_than_baseline}, it suffices to prove the following pointwise inequalities:
        \begin{align}
            L_\ell(q, f_{\mathrm{PB}}(\bm{x})) &\le  L_\ell(q, f(\bm{x}))\quad \forall \bm{x} \in A, q \ge f(\bm{x}) + \varepsilon_{\mathrm{UC}}, \label{eq:conditional_risk_inequality_cond_A_pointwise}\\
            L_\ell(q, f_{\mathrm{PB}}(\bm{x})) &=  L_\ell(q, f(\bm{x}))\quad \forall \bm{x} \in M, q \in [0,1], \label{eq:conditional_risk_inequality_cond_M_pointwise} \\
            L_\ell(q, f_{\mathrm{PB}}(\bm{x})) &\le  L_\ell(q, f(\bm{x}))\quad \forall \bm{x} \in B, q \le f(\bm{x}) - \varepsilon_{\mathrm{OC}}. \label{eq:conditional_risk_inequality_cond_B_pointwise}
        \end{align}
        We first derive~\cref{eq:conditional_risk_inequality_cond_A_pointwise}.
        Since $f(\bm{x}) \le f_{\mathrm{PB}}(\bm{x}) = \tilde{a}$, for all $q \ge f(\bm{x}) + \varepsilon_{\mathrm{UC}} \ge \tilde{a} \ge f(\bm{x})$, we can write $\tilde{a} = (1-\lambda)q + \lambda f(\bm{x})$ with $\lambda = (q-\tilde{a})/(q-f(\bm{x})) \in [0,1]$.
        Then, by strict convexity of $L_{\ell}(q, \cdot)$, we have
        \begin{align}
            L_\ell(q, f_{\mathrm{PB}}(\bm{x})) &= L_\ell(q, \tilde{a}) = L_{\ell}(q, (1-\lambda) q + \lambda f(\bm{x})) \\
            & \le (1-\lambda) L_\ell(q,q) + \lambda  L_\ell(q, f(\bm{x})) \\
            & \le L_\ell(q, f(\bm{x})),
            \tag{\ref{eq:conditional_risk_inequality_cond_A_pointwise}}
        \end{align}
        which proves~\cref{eq:conditional_risk_inequality_cond_A_pointwise}.
        The proof of~\cref{eq:conditional_risk_inequality_cond_B_pointwise} is analogous.
        Since $f(\bm{x}) = f_{\mathrm{PB}}(\bm{x})$ on $M$,~\cref{eq:conditional_risk_inequality_cond_M_pointwise} holds.
        These imply~\cref{eq:pb_risk_is_not_worse_than_baseline}.
        Assume that the lower bound is non-vacuous.
        Then, by strict convexity of $L_\ell$,~\cref{eq:conditional_risk_inequality_cond_A_pointwise} with $q=\mathbb{P}(Y=1 \mid f(X)) > f(X)$ is strict on some (measurable) subset $A^\prime$ of $A$ such that $P(A^\prime) > 0$.
        This means that~\cref{eq:pb_risk_is_not_worse_than_baseline} is strict.
        The same argument applies when the upper bound is non-vacuous.
        This concludes the proof of~\cref{eq:pb_risk_is_not_worse_than_baseline}.
    \end{proof}
    
\subsection{Proof of \texorpdfstring{\cref{thm:smooth_ce_bound_pb}}{Theorem \ref{thm:smooth_ce_bound_pb}}}
\subsubsection{Preliminary}
    As in the previous section, we focus on the binary case: $K=2$ and $\mathcal{Y}=\{0,1\}$.

    By~\cref{coro:bcsoftmax_is_clipped_sigmoid},~\cref{thm:smooth_ce_bound_pb} can be rewritten as follows.
    \begin{theoremmd}
        Given $g:\mathcal{X} \to \real$ and $D_{\mathrm{val}}$, let 
        \begin{align}
            f_{\mathrm{PB}}(\bm{x}; a, b) = \clip(\sigma(g(\bm{x})), \tilde{a}(a,b), \tilde{b}(a,b)) = \sigma(\clip(g(x),\tilde{c}(a,b), \tilde{C}(a,b))),
        \end{align}
        and $(\hat{a}, \hat{b})$ be the solution to~\cref{eq:pb_optimization} with the xent loss: $\ell: y \in \{0,1\} \times f \in [0,1] \mapsto-y \log f - (1-y) \log (1-f)$.
        For any $\delta > 0$, with probability at least $1-\delta$, 
        \begin{align}
            \begin{split}
                 \smCE_{\mathcal{D}}(\hat{f}_{\mathrm{PB}})^2 \le & \frac{1}{2}\left(\hat{R}_{\ell, D_{\mathrm{val}}}(\hat{f}_{\mathrm{PB}})-\inf_{\kappa \in \mathcal{K}}\hat{R}_{\ell,D_{\mathrm{val}}}(\hat{f}_{\mathrm{PB},\kappa})\right) \\
                &+\frac{1}{\sqrt{N_{\mathrm{val}}}}\left(42B_{k(\hat{a},\hat{b})} + 144 + \left(B_{k(\hat{a},\hat{b})}+2\right) \sqrt{\frac{\log (2^{k(\hat{a},\hat{b})+2}/\delta)}{2}} \right),
                \label{eq:smooth_ce_bound_pb_sigmoid}
            \end{split}
        \end{align}
        where $\hat{f}_{\mathrm{PB}} \coloneqq\hat{f}_{\mathrm{PB}}(\cdot; \hat{a},\hat{b})$, $\hat{f}_{\mathrm{PB},\kappa}\coloneqq \sigma\circ\kappa \circ \sigma^{-1} \circ \hat{f}_{\mathrm{PB}} = \sigma \circ \kappa \circ \clip(\cdot, \tilde{c}(\hat{a}, \hat{b}), \tilde{C}(\hat{a},\hat{b}))\circ g$,
        \begin{align}
            B_k &\coloneqq \sigma^{-1}(1-2^{-k}) = \log \left(\frac{1-2^{-k}}{2^{-k}}\right), \quad k \in \mathbb{N}_{>0}, \label{eq:B_k}\\
            k(a,b) &\coloneqq \min \{ k \in \{1, 2, \ldots\}\mid \tilde{a}(a,b) \ge 2^{-k} \iff \tilde{b}(a,b)\le 1-2^{-k}\}.
            \label{eq:k_of_a_b}
        \end{align}
        \label{thm:smooth_ce_bound_pb_sigmoid}
    \end{theoremmd}
    
    To prove~\cref{thm:smooth_ce_bound_pb_sigmoid}, we first present some important definitions and results from~\citep{blasiok2023does}.
    \begin{definitionmd}[Dual loss, Definition 4.3 in~\citep{blasiok2023does}]
        For a function $\psi: \real\to\real$, we define a dual loss function $\ell^{(\psi)}:\{0,1\}\times\real \to\real$ such that
        \begin{align}
            \ell^{(\psi)}(y, t) = \psi(t)-yt \quad \text{for every} \ y\in\{0,1\} \ \text{and} \ t \in \real.
        \end{align}
        Consequently, if a loss function $\ell:\{0,1\}\times V\to\real$ satisfies
        \begin{align}
            \ell(y, v) = \psi(\dual(v)) - y\dual(v) \ \text{for every} \ y \in \{0,1\} \ \text{and} \ v \in V
        \end{align}
        for some $V \subseteq [0,1]$, where $\dual(v) \coloneqq \ell(0,v)-\ell(1,v)$, then
        \begin{align}
            \ell(y, v) = \ell^{(\psi)}(y, \dual(v))
        \end{align}
        and we say $\ell^{(\psi)}$ is the dual loss of $\ell$.
    \end{definitionmd}
    \paragraph{Dual loss of $\ell_{\mathrm{bxent}}$.} Note that for the binary cross entropy loss $\ell_{\mathrm{bxent}}$, its dual prediction $t=\dual(v)$ is the logit of $v$:
    \begin{align}
        t=\dual(v)=\ell(0, v)-\ell(1,v)=-\log (1-v) + \log v = \log \left(\frac{v}{1-v}\right)  = \sigma^{-1}(v).
    \end{align}
    Since $\dual^{-1} = \sigma$, its dual loss is the logistic loss:
    \begin{align}
        \ell(y, \sigma^{-1}(t)) = \ell^{(\psi)}_{\mathrm{bxent}}(y,t) = \log(1+e^t)-yt \eqqcolon \ell_{\mathrm{logistic}}(y, t).
        \label{eq:bxent_on_p_is_logistic_on_logit}
    \end{align}
    
    We next define an important notion, dual post-processing gap.
    \begin{definitionmd}[Dual post-processing gap for binary cross entropy loss, Definition 2.5 and 4.4 in~\citet{blasiok2023does}]
        Let $\mathcal{K}$ denote the family of all post-processing functions $\kappa:\real\to\real$ such that the update function $\eta(v)=\kappa(v)-v$ is 1-Lipschitz and bounded $\lvert\eta(v)\rvert\le 4$ for all $v$.
        For a logit function $g: \mathcal{X}\to\real$ and a distribution $\mathcal{D}$ over $\mathcal{X}\times\mathcal{Y}$, we define the dual post-processing gap of $g$ w.r.t $\mathcal{D}$ as
        \begin{align}
            \dpGap_{\mathcal{D}}(g) \coloneqq R_{\ell_{\mathrm{logistic}}, \mathcal{D}}(g) - \inf_{\kappa \in \mathcal{K}} R_{\ell_{\mathrm{logistic}}, \mathcal{D}}(\kappa \circ g).
            \label{eq:dpgap}
        \end{align}
        \label{def:dpgap}
    \end{definitionmd}

    \citet{blasiok2023does} showed that the dual post-processing gap and the smooth calibration error are quadratically related.
    \begin{theoremmd}[Corollary 2.7 in~\citet{blasiok2023does}]
        Let $\dpGap$ and $\mathrm{smCE}$ be defined as in~\cref{eq:dpgap} and~\cref{eq:smCE}, respectively.
        For any logit function $g:\mathcal{X}\to\real$ and any distribution $\mathcal{D}$ over $\mathcal{X}\times\mathcal{Y}$,
        \begin{align}
            \smCE_{\mathcal{D}}(f)^2 \le \frac{1}{2}\dpGap_{\mathcal{D}}(g),
        \end{align}
        where $f\coloneqq \sigma \circ g$.
        \label{thm:dpgap_implies_small_smce}
    \end{theoremmd}

    By~\cref{coro:bcsoftmax_is_clipped_sigmoid} and~\cref{thm:dpgap_implies_small_smce}, we immediately obtain the following bound on $\mathrm{smCE}(f_{\mathrm{PB}})$.
    \begin{corollarymd}
        For any logit function $g:\mathcal{X}\to\real$, lower bound scalar $a \in [0, 1/K]$, upper bound scalar $b \in [1/K, 1]$, and distribution $\mathcal{D}$ over $\mathcal{X}\times\mathcal{Y}$,
        \begin{align}
            \smCE_{\mathcal{D}}(f_{\mathrm{PB}}(\cdot, a, b))^2 \le \frac{1}{2}\dpGap_{\mathcal{D}}(g_{\mathrm{clip}}(\cdot, a, b)),
        \end{align}
        where
        \begin{align}
            g_{\mathrm{clip}}(\bm{x}, a, b) = \clip(g(\bm{x}), \tilde{c}(a,b), \tilde{C}(a,b)).
        \end{align}
        \label{coro:smce_is_bounded_by_dpgap_pb}
    \end{corollarymd}

\subsubsection{Proof of \texorpdfstring{\cref{thm:smooth_ce_bound_pb_sigmoid}}{Theorem \ref{thm:smooth_ce_bound_pb_sigmoid}}}
    Hence, we prove~\cref{thm:smooth_ce_bound_pb_sigmoid}, which is equivalent to~\cref{thm:smooth_ce_bound_pb}, by upper bounding~$\dpGap_{\mathcal{D}}(g_{\mathrm{clip}}(\cdot, \hat{a}, \hat{b}))$.
    To simplify notation, we denote $\ell_{\mathrm{logistic}}$ as $\ell$, $R_{\ell, \mathcal{D}}$, the expected $\ell$-risk on $\mathcal{D}$, as $R$,  $\hat{R}_{\ell, D_{\mathrm{val}}}$, the empirical $\ell$-risk on $D_{\mathrm{val}}$, as $\hat{R}$, and $\dpGap_{\mathcal{D}}$ as $\dpGap$ in the rest of this section.
    
    The first term in~\cref{eq:dpgap} is the expected $\ell$-risk of $g_{\mathrm{clip}}(\cdot, \hat{a}, \hat{b})$.
    The lower and upper bounds $\hat{a}$ and $\hat{b}$ are obtained by minimizing the empirical $\ell$-risk on $D_{\mathrm{val}}$:
    \begin{align}
        \hat{a},\hat{b} &= \argmin_{a\in[0,1/2], b\in[1/2,1]} \hat{R}(g_{\mathrm{clip}}(\cdot, a, b)).
    \end{align}
    Therefore, we derive an upper bound on $\dpGap(g_{\mathrm{clip}}(\cdot, \hat{a}, \hat{b}))$ via generalization error analysis techniques~\citep{mohri2018foundations,shalev2014understanding}.

    We first decompose and bound $\dpGap(\hat{g}_{\mathrm{clip}})$ as (to simplify notation, we set $\hat{g}_{\mathrm{clip}}(\bm{x})=g_{\mathrm{clip}}(\bm{x}, \hat{a}, \hat{b})$)
    \begin{align}
        \dpGap(\hat{g}_{\mathrm{clip}}) &= R(\hat{g}_{\mathrm{clip}}) - \inf_{\kappa \in \mathcal{K}} R(\kappa \circ \hat{g}_{\mathrm{clip}})\\
        &= R(\hat{g}_{\mathrm{clip}}) - \hat{R}(\hat{g}_{\mathrm{clip}}) + \hat{R}(\hat{g}_{\mathrm{clip}}) - \inf_{\kappa \in \mathcal{K}} R(\kappa \circ \hat{g}_{\mathrm{clip}})\\
        &= R(\hat{g}_{\mathrm{clip}}) - \hat{R}(\hat{g}_{\mathrm{clip}}) + \hat{R}(\hat{g}_{\mathrm{clip}}) - \inf_{\kappa \in \mathcal{K}} \hat{R}(\kappa \circ \hat{g}_{\mathrm{clip}}) + \inf_{\kappa \in \mathcal{K}} \hat{R}(\kappa \circ \hat{g}_{\mathrm{clip}}) - \inf_{\kappa \in \mathcal{K}} R(\kappa \circ \hat{g}_{\mathrm{clip}}) \\
        &\le \phi(\hat{g}_{\mathrm{clip}}) + \widehat{\dpGap}(\hat{g}_{\mathrm{clip}}) + \sup_{\kappa \in \mathcal{K}} \phi(\kappa\circ\hat{g}_{\mathrm{clip}}),
        \label{eq:dpgap_decomposed}
    \end{align}
    where $\phi(g)$ is the generalization gap of $g$ on $D_{\mathrm{val}}$ and $\mathcal{G}_{\mathrm{clip}} \coloneqq \{g_\mathrm{clip}(\cdot, a, b): a\in[0,1/2], b\in[1/2,1]\}$ and $\widehat{\dpGap}$ is the empirical dual post-processing gap:
    \begin{align}
        \phi(g) & \coloneqq \lvert R(g) - \hat{R}(g)\rvert, \\
        \widehat{\dpGap}(\hat{g}_{\mathrm{clip}}) &\coloneqq \hat{R}(\hat{g}_{\mathrm{clip}}) - \inf_{\kappa \in \mathcal{K}} \hat{R}(\kappa \circ \hat{g}_{\mathrm{clip}}).
    \end{align}
    We prove~\cref{thm:smooth_ce_bound_pb_sigmoid} by bounding the first term and third term in~\cref{eq:dpgap_decomposed}.
    
    For the first term, we can derive the following bound.
    The proof will be shown later.
    \begin{lemmamd}
        For any $\delta \in (0,1)$, with probability at least $1-\delta$,
        \begin{align}
            \phi(\hat{g}_{\mathrm{clip}}) \le \frac{12B_{k(\hat{a},\hat{b})}}{\sqrt{N_{\mathrm{val}}}} + B_{k(\hat{a},\hat{b})}\sqrt{\frac{\log (2^{k(\hat{a},\hat{b})+1}/\delta)}{2N_{\mathrm{val}}}},
            \label{eq:non_uniform_bound_of_generalization_gap_g_clip}
        \end{align}
        where $B_k$ is defined as~\cref{eq:B_k} and $k(a,b)$ is defined in ~\cref{eq:k_of_a_b}.
        \label{lemma:non_uniform_bound_of_generalization_gap_g_clip}
    \end{lemmamd}
    
    For the third term, we have the following result, which will also be proved later.
    \begin{lemmamd}
        For any $\delta \in (0,1)$, with probability at least $1-\delta$,
        \begin{align}
            \sup_{\kappa \in \mathcal{K}} \phi(\kappa \circ \hat{g}_{\mathrm{clip}}) \le \frac{72\left(B_{k(\hat{a},\hat{b})}+4\right)}{\sqrt{N_{\mathrm{val}}}} + \left(B_{k(\hat{a},\hat{b})}+4\right)\sqrt{\frac{\log (2^{k(\hat{a},\hat{b})+1}/\delta)}{2N_{\mathrm{val}}}},
            \label{eq:non_uniform_bound_of_generalization_gap_kappa_g_clip}
        \end{align}
        \label{lemma:non_uniform_bound_of_generalization_gap_kappa_g_clip}
    \end{lemmamd}
    
    \begin{proof}[Proof of~\cref{thm:smooth_ce_bound_pb_sigmoid}]
        We can immediately obtain~\cref{eq:smooth_ce_bound_pb_sigmoid} by substituting~\cref{eq:non_uniform_bound_of_generalization_gap_g_clip,eq:non_uniform_bound_of_generalization_gap_kappa_g_clip} into~\cref{eq:dpgap_decomposed} with $\delta$ replaced by $\delta/2$ and combining it with~\cref{coro:smce_is_bounded_by_dpgap_pb}.
    \end{proof}

\subsubsection{Proof of \texorpdfstring{\cref{lemma:non_uniform_bound_of_generalization_gap_g_clip}}{Lemma \ref{lemma:non_uniform_bound_of_generalization_gap_g_clip}}}
    We first derive a non-uniform bound on $\phi(\hat{g}_{\mathrm{clip}})$.
    To do so, we derive a uniform bound on a small subset of $\mathcal{G}_{\mathrm{clip}}$, $\mathcal{G}_{\mathrm{clip},k}$. 
    \begin{lemmamd}
        Given $g$ and $k\in\mathbb{N}_{>0}$, we define $\mathcal{G}_{\mathrm{clip},k} \subseteq \mathcal{G}_{\mathrm{clip}}$ as 
        \begin{align}
            \mathcal{G}_{\mathrm{clip},k} &\coloneqq \{g_\mathrm{clip}(\cdot, a, b) \mid \tilde{a}=\max(a, 1-b)\in [2^{-k}, 1/2] \iff \tilde{b}=\min(1-a, b) \in  [1/2, 1-2^{-k}]\} \\
            &= \left\{\clip(g(\cdot), -C, C) \mid 0\le C \le B_k\right\} \quad\text{where}\quad B_k\coloneqq \log \left(\frac{1-2^{-k}}{2^{-k}}\right).
        \end{align}
        Then, for any $\delta_k \in (0,1)$, with probability at least $1-\delta_k$,
        \begin{align}
            \sup_{g_{\mathrm{clip}} \in \mathcal{G}_{\mathrm{clip},k}} \phi(g_{\mathrm{clip}}) \le \frac{12B_k}{\sqrt{N_{\mathrm{val}}}} + B_k\sqrt{\frac{\log (2/\delta_k)}{2N_{\mathrm{val}}}}.
            \label{eq:bound_of_generalization_gap_G_clip_k}
        \end{align}
        \label{lemma:bound_of_generalization_gap_G_clip_k}
    \end{lemmamd}
    
    \begin{proof}
        By the basic Rademacher bound on the maximal generalization gap (Theorem 3.3 in~\cite{mohri2018foundations}) and the 1-Lipschitz continuity of $\ell$, with probability at least $1-\delta_k$, we have
        \begin{align}
            \sup_{g_{\mathrm{clip}}\in\mathcal{G}_\mathrm{clip},k} \phi(g_{\mathrm{clip}})\le 2\cdot \mathfrak{R}_{N_\mathrm{val}}(\mathcal{G}_{\mathrm{clip}, k}) + M_k \sqrt{\frac{\log (2/\delta_k)}{2N_{\mathrm{val}}}},
            \label{eq:basic_rademacher_bound_G_clip_k}
        \end{align}
        where $\mathfrak{R}_{N_\mathrm{val}}(\mathcal{G}_{\mathrm{clip}, k})$ is the Rademacher complexity of $\mathcal{G}_{\mathrm{clip},k}$ on $\mathcal{D}$:
        \begin{align}
            \mathfrak{R}_{N}(\mathcal{G}_{\mathrm{clip},k}) &\coloneqq \mathbb{E}_{D \sim \mathcal{D}^{N}}[\hat{\mathfrak{R}}(\mathcal{G}_{\mathrm{clip}, k}(D))] \quad \text{for} \  N > 0, \label{eq:rademacher_complexity}\\
            \hat{\mathfrak{R}}(\mathcal{G}_{\mathrm{clip}, k}(D)) &\coloneqq \mathbb{E}_{\bm{\sigma}\in\{-1,1\}^{N}}\left[\sup_{\bm{g}\in\mathcal{G}_{\mathrm{clip}, k}(D)}\frac{1}{N}\sum_{n=1}^{N}\sigma_n\cdot g_n\right] \quad \text{for}\  D = \{(\bm{x}_n, y_n)\}_{n=1}^N, \label{eq:empirical_rademacher_complexity} \\
            \mathcal{G}_{\mathrm{clip}, k}(D) &\coloneqq \{g_{\mathrm{clip}}(D) \in [-B_k, B_k]^N \mid g_{\mathrm{clip}} \in \mathcal{G}_{\mathrm{clip},k}\}, \\
            g_{\mathrm{clip}}(D) &\coloneqq \left(g_{\mathrm{clip}}(\bm{x}_1), \ldots, g_{\mathrm{clip}}(\bm{x}_N)\right)^\top\in[-B_k, B_k]^{N}.
        \end{align}
        and $M_k$ is the maximum deviation of the loss function on $\mathcal{G}_{\mathrm{clip},k}$:
        \begin{align}
            M_k \coloneqq \sup_{g_{\mathrm{clip}}, (\bm{x},y), (\bm{x}^{\prime},y^{\prime})} \ell(y, g_{\mathrm{clip}}(\bm{x}))-\ell(y^{\prime}, g_{\mathrm{clip}}(\bm{x}^{\prime})).
        \end{align}
        The Rademacher variables $\sigma_{n}$ in~\cref{eq:empirical_rademacher_complexity} are sampled independently from the uniform Rademacher distribution.
        Since the output of $g_{\mathrm{clip}} \in \mathcal{G}_{\mathrm{clip},k}$ is bounded from $-B_k$ to $B_k$, we get
        \begin{align}
            M_k = \ell(0, B_k) - \ell(0, -B_k) = \log(1+\exp (B_k)) - \log (1+\exp(-B_k)) = B_k.
            \label{eq:maximum_deviation_of_loss}
        \end{align}
    
        We next derive an upper bound on the empirical Rademacher complexity in~\cref{eq:empirical_rademacher_complexity} by the chaining technique~\citep{shalev2014understanding}.
        Given $D\sim\mathcal{D}^{N}$, for any two functions $g_{\mathrm{clip},1}=\clip(g(\cdot), -C_1, C_1), g_{\mathrm{clip},2}=\clip(g(\cdot), -C_2, C_2) \in \mathcal{G}_{\mathrm{clip}, k}$, we have
        \begin{align}
            \lVert g_{\mathrm{clip},1}(D) - g_{\mathrm{clip},2}(D) \rVert_2 &= \left(\sum_{n=1}^N (g_{\mathrm{clip},1}(\bm{x}_n) - g_{\mathrm{clip},2}(\bm{x}_n))^2\right)^{1/2} \\
            &\le \sqrt{N}\max (|-C_1-(-C_2)|, |C_1-C_2|) \label{eq:bound_of_distance_of_G_clip_k_of_D} \\
            &\le \sqrt{N}B_k.
            \label{eq:diam_of_G_clip_k_of_D}
        \end{align}
        For given $r > 0$, let $I_{r}$ be $(r/\sqrt{N})$-coverings of $[0, B_k]$:
        \begin{align}
            \min_{C^\prime \in I_{r}} |C^\prime - C| \le \frac{r}{\sqrt{N}} \quad \forall C\in[0,B_k].
        \end{align}
        We define $\mathcal{G}_{\mathrm{clip}, k}(D, I_{r}) \subset [-B_k, B_k]^N$ as
        \begin{align}
            \mathcal{G}_{\mathrm{clip}, k}(D, I_{r}) \coloneqq \{g_{\mathrm{clip}}(D)\mid g_{\mathrm{clip}}=\clip(g(\cdot), -C, C), C \in I_{r}\}.
        \end{align}
        Then, by~\cref{eq:bound_of_distance_of_G_clip_k_of_D} and the definition of $I_{r}$, $\mathcal{G}_{\mathrm{clip}, k}(D, I_{r})$ is an $r$-covering of $\mathcal{G}_{\mathrm{clip}, k}(D)$:
        \begin{align}
            \min_{\bm{g}_{\mathrm{clip}}^{\prime} \in \mathcal{G}_{\mathrm{clip}, k}(D, I_{r})}  \lVert \bm{g}_{\mathrm{clip}}^{\prime} -\bm{g}_{\mathrm{clip}} \rVert_2 \le \sqrt{N}\max (r/\sqrt{N}, r/\sqrt{N}) = r \quad \forall \bm{g}_{\mathrm{clip}} \in \mathcal{G}_{\mathrm{clip}, k}(D).
            \label{eq:r_covering_of_G_clip_k_of_D}
        \end{align}
        Moreover, the covering number of $[0, B_k]$ with radius $r/\sqrt{N}$ is less than or equal to $\lceil B_k\sqrt{N}/(2r)\rceil$.
        Therefore, we have
        \begin{align}
            \inf_{I_{r}} \lvert \mathcal{G}_{\mathrm{clip}, k}(D, I_{r}) \rvert \le \left\lceil\frac{B_k\sqrt{N}}{2r}\right\rceil.
            \label{eq:covering_number_of_G_clip_k}
        \end{align}

        Then, we can derive an upper bound on~\cref{eq:empirical_rademacher_complexity} by the chaining technique (Lemma 27.4 in~\citet{shalev2014understanding}).
        For each $\bm{\sigma} \in \{-1,+1\}^N$, let $\bm{g}^{*}(\bm{\sigma}) = \argmax_{\bm{g} \in \mathcal{G}_{\mathrm{clip}, k}(D)} \bm{\sigma}^\top \bm{g}$ (if a maximizer does not exist, choose $\bm{g}^*(\bm{\sigma})$ such that $\bm{\sigma}^\top g^*(\bm{\sigma})$ is close enough to the supremum).
        We fix an integer $M>0$ and set $r_m \coloneqq \sqrt{N}B_k 2^{-m}$ for all $m \in [M]$.
        We pick a ($r_m/\sqrt{N}$)-covering of $[0, B_k]$, $I_{r_m}$, such that $|I_{r_m}|\le2^{m-1}$ for all $m \in [M]$.
        For $m=0$, we define $I_{r_0}\coloneqq\{0\}$, which is a $(r_0/\sqrt{N})$-covering of $[0, B_k]$.
        By the definition of $I_{r_m}$, $\mathcal{G}_{\mathrm{clip}, k, m}(D) \coloneqq \mathcal{G}_{\mathrm{clip}, k}(D, I_{r_m})$ is an $r_m$-covering of $\mathcal{G}_{\mathrm{clip},k}(D)$ and
        \begin{align}
            \lvert \mathcal{G}_{\mathrm{clip}, k, m}(D) \rvert \le \left\lceil\frac{\sqrt{N}B_k}{2r_m}\right\rceil = \left\lceil\frac{\sqrt{N}B_k}{2\sqrt{N}B_k2^{-m}}\right\rceil = 2^{m-1}.
            \label{eq:covering_number_of_G_clip_k_m_of_D}
        \end{align}
        We define $\bm{g}_m(\bm{\sigma})$ to be the nearest neighbor of $\bm{g}^*(\bm{\sigma})$ in:
        \begin{align}
            g_m(\bm{\sigma}) = \argmin_{\bm{g} \in \mathcal{G}_{\mathrm{clip}, k, m}(D)}  \lVert g^*(\bm{\sigma})-\bm{g})\rVert_2.
        \end{align}
        For all $m \in [M]$, we have
        \begin{align}
            \lVert g_m(\bm{\sigma}) - g_{m-1}(\bm{\sigma}) \lVert_2 &\le \lVert g_m(\bm{\sigma}) - g^*(\bm{\sigma}) \lVert_2 + \lVert g^*(\bm{\sigma}) - g_{m-1}(\bm{\sigma}) \lVert_2\\
            &\le \sqrt{N}B_k (2^{-m}+2^{-m+1}) \\
            &= 3\sqrt{N}B_k 2^{-m}.
        \end{align}
        For each $m$, we define the set
        \begin{align}
            \tilde{\mathcal{G}}_{m} \coloneqq \{\bm{g}-\bm{g}^{\prime}\mid \bm{g} \in \mathcal{G}_{\mathrm{clip},k,m}(D), \bm{g}^\prime \in \mathcal{G}_{\mathrm{clip},k,m-1}(D), \lVert \bm{g}-\bm{g}^\prime\rVert_2\le3\sqrt{N}B_k2^{-m}\}.
        \end{align}
        Clearly, $g_m(\bm{\sigma}) - g_{m-1}(\bm{\sigma}) \in \tilde{\mathcal{G}}_m$.
        Then, we have
        \begin{align}
            \hat{\mathfrak{R}}(\mathcal{G}_{\mathrm{clip}, k}(D)) &= \mathbb{E}_{\bm{\sigma}}\left[\frac{1}{N}\bm{\sigma}^\top\cdot g^{*}(\bm{\sigma})\right]\\
            &= \frac{1}{N}\mathbb{E}_{\bm{\sigma}}\left[\bm{\sigma}^\top\{g^{*}(\bm{\sigma})-g_{M}(\bm{\sigma})\} + \sum_{m=1}^{M}\bm{\sigma}^\top\{g_{m}(\bm{\sigma})-g_{m-1}(\bm{\sigma})\}\right]\\
            &\le \frac{1}{N}\mathbb{E}\left[\lVert\bm{\sigma}\rVert_2\cdot\lVert g^{*}(\bm{\sigma})-g_{M}(\bm{\sigma})\rVert_2\right] + \sum_{m=1}^M\frac{1}{N}\mathbb{E}\left[\sup_{\bm{g}\in\tilde{\mathcal{G}}_m} \bm{\sigma}^\top\bm{g}\right]\\
            &=\frac{1}{N}\mathbb{E}\left[\lVert\bm{\sigma}\rVert_2\cdot\lVert g^{*}(\bm{\sigma})-g_{M}(\bm{\sigma})\rVert_2\right] + \sum_{m=1}^M\hat{\mathfrak{R}}(\tilde{\mathcal{G}}_m).
            \label{eq:bound_of_empirical_rademacher_complexity_of_G_clip_k_D_1}
        \end{align}
        Since $\lVert \bm{\sigma}\rVert_2=\sqrt{N}$ and $\lVert g^{*}(\bm{\sigma})-g_{M}(\bm{\sigma})\rVert_2 \le \sqrt{N}B_k2^{-M}$, the first term in~\cref{eq:bound_of_empirical_rademacher_complexity_of_G_clip_k_D_1} is at most $B_k2^{-M}$.
        By Massart's lemma and~\cref{eq:covering_number_of_G_clip_k_m_of_D}, we have
        \begin{align}
            \hat{\mathfrak{R}}(\tilde{\mathcal{G}}_m) &\le \max_{\bm{g} \in \tilde{\mathcal{G}}_m} \lVert \bm{g} - \bar{\bm{g}}\rVert_2\frac{\sqrt{2\log |\tilde{\mathcal{G}}_m|}}{N}\\
            &\le 3\sqrt{N}B_k2^{-m}\frac{\sqrt{2\log |\mathcal{G}_{\mathrm{clip},k,m}(D)|^2}}{N} \\
            &= 6\sqrt{N}B_k2^{-m}\frac{\sqrt{\log |\mathcal{G}_{\mathrm{clip},k,m}(D)|}}{N}\\
            &\le \frac{6B_k2^{-m}}{\sqrt{N}}\sqrt{\log 2^{(m-1)}},
        \end{align}
        where $\bar{\bm{g}}\coloneqq \lvert\tilde{\mathcal{G}}_m\rvert^{-1}\sum_{\bm{g}\in\tilde{\mathcal{G}}_m}\bm{g}$.
        Therefore, we obtain
        \begin{align}
            \hat{\mathfrak{R}}(\mathcal{G}_{\mathrm{clip}, k}(D)) &\le\frac{1}{N}\mathbb{E}\left[\lVert\bm{\sigma}\rVert_2\cdot\lVert g^{*}(\bm{\sigma})-g_{M}(\bm{\sigma})\rVert_2\right] + \sum_{m=1}^M\hat{\mathfrak{R}}(\tilde{\mathcal{G}}_m)
            \tag{\ref{eq:bound_of_empirical_rademacher_complexity_of_G_clip_k_D_1}}\\
            &\le B_k2^{-M}+\frac{6B_k}{\sqrt{N}}\sum_{m=1}^M2^{-m}\sqrt{\log \lvert \mathcal{G}_{\mathrm{clip},k,m}(D)\rvert} \\
            &\le B_k2^{-M} + \frac{6B_k}{\sqrt{N}}\sum_{m=1}^M 2^{-m}\sqrt{\log 2^{m-1}} \\
            &= B_k2^{-M} + \frac{6B_k}{\sqrt{N}}\sum_{m=1}^M 2^{-m}\sqrt{(m-1)\log 2}\\
            & =B_k2^{-M} + \frac{6B_k}{\sqrt{N}}\left(2^{-1}\sqrt{\log 2}\right)\sum_{m=0}^{M-1}\sqrt{m}\cdot2^{-m} \\
            & =B_k2^{-M} + \frac{6B_k}{\sqrt{N}}\left(2^{-1}\sqrt{\log 2}\right)\sum_{m=1}^{M-1} \sqrt{m}\cdot 2^{-m/2}\cdot2^{-m/2} \\
            &\le B_k2^{-M} + \frac{6B_k}{\sqrt{N}}\left(2^{-1}\sqrt{\log 2}\right)\sqrt{\left(\sum_{m=1}^{M-1}2^{-m}\right)\left(\sum_{m=1}^{M-1}m2^{-m}\right)},
        \end{align}
        where the last inequality follows from the Cauchy–Schwarz inequality.
        By taking $M\to\infty$ and the following equations
        \begin{align}
            \sum_{m=1}^{\infty}2^{-m} = 1,\quad           \sum_{m=1}^{\infty}m2^{-m} = 2,
        \end{align}
        we obtain the following upper bounds of the empirical Rademacher complexity and Rademacher complexity:
        \begin{align}
            \hat{\mathfrak{R}}(\mathcal{G}_{\mathrm{clip},k}(D)) &\le 6B_k \sqrt{\frac{2^{-1}\log 2}{N}}\le 6B_k \sqrt{\frac{1}{N}}, \\
            \mathfrak{R}_{N}(\mathcal{G}_{\mathrm{clip},k}) &= \mathbb{E}_{D \sim \mathcal{D}^{N}}[\hat{\mathfrak{R}}(\mathcal{G}_{\mathrm{clip},k}(D))] \le 6B_k \sqrt{\frac{1}{N}}.
        \end{align}
        By applying the above inequality and~\cref{eq:maximum_deviation_of_loss} to~\cref{eq:basic_rademacher_bound_G_clip_k}, we immediately obtain~\cref{eq:bound_of_generalization_gap_G_clip_k}.
    \end{proof}

    By the above lemma and Theorem 7.4 in~\cite{shalev2014understanding}, we can obtain~\cref{eq:non_uniform_bound_of_generalization_gap_g_clip}.
    \begin{proof}[Proof of~\cref{lemma:non_uniform_bound_of_generalization_gap_g_clip}]
        By the definition of $\mathcal{G}_{\mathrm{clip},k}$, we have $\mathcal{G}_{\mathrm{clip},1}\subset \mathcal{G}_{\mathrm{clip},2} \subset \cdots$ and  $\mathcal{G}_{\mathrm{clip}}=\bigcup_{k=1}^{\infty}\mathcal{G}_{\mathrm{clip},k}$.
        Then, for all $k \ge 1$ and $g_\mathrm{clip} \in \mathcal{G}_{\mathrm{clip},k}$, the following bounds simultaneously hold:
        \begin{align}
            \phi(g_{\mathrm{clip}}) \le \frac{12B_{k}}{\sqrt{N_{\mathrm{val}}}} + B_k\sqrt{\frac{\log (2/(2^{-k}\delta_k))}{2N_{\mathrm{val}}}},
        \end{align}
        and this implies~\cref{eq:non_uniform_bound_of_generalization_gap_g_clip} (Theorem 7.4 in~\citet{shalev2014understanding}).
    \end{proof}

\subsubsection{Proof of \texorpdfstring{\cref{lemma:non_uniform_bound_of_generalization_gap_kappa_g_clip}}{Lemma \ref{lemma:non_uniform_bound_of_generalization_gap_kappa_g_clip}}}
    We next consider an upper bound on the third term in~\cref{eq:dpgap_decomposed}, $\sup_{\kappa \in \mathcal{K}} \phi(\kappa \circ\hat{g}_{\mathrm{clip}})$.
    Fortunately, we can derive an upper bound on this term similarly to~\cref{lemma:non_uniform_bound_of_generalization_gap_g_clip}.
    
    We first present an upper bound on the covering number of the set of all $1$-Lipschitz functions from $[-B, B]$ to $[-4,4]$.
    \begin{lemmamd}[\citet{shiryaev1992selected}]
        Let $B>0$ and $\mathcal{E}_{B}\coloneqq\{\eta:[-B,B]\to[-4,4] \mid \eta \text{\ is 1-Lipschitz}\}$ denote the family of all $1$-Lipschitz functions from $[-B, B]$ to $[-4,4]$.
        For any $r>0$, let $\mathcal{N}(r, \mathcal{E}_B, \lVert \cdot \rVert_{\infty})$ denote the $r$-covering number with respect to the sup norm.
        Then, for all $r > 0$,
        \begin{align}
            \mathcal{N}(r, \mathcal{E}_B, \lVert \cdot \rVert_{\infty}) \le \left(\left\lceil\frac{4}{r}\right\rceil\right) \cdot 2^{\lceil 2B/r\rceil}.
        \end{align}
        \label{lemma:covering_number_of_bounded_lipschitz_functions}
    \end{lemmamd}
    Then, we obtain the following result.
    \begin{lemmamd}
        Given $g$, we define $\mathcal{G}_{\mathrm{clip}, k}$ as in~\cref{lemma:bound_of_generalization_gap_G_clip_k}.
        Then, for any $\delta_k \in (0,1)$, with probability at least $1-\delta_k$,
        \begin{align}
            \sup_{g_{\mathrm{clip}} \in \mathcal{G}_{\mathrm{clip},k},\kappa \in \mathcal{K}} \phi(\kappa\circ g_{\mathrm{clip}}) \le \frac{72(B_k+4)}{\sqrt{N_{\mathrm{val}}}} + (B_k+4)\sqrt{\frac{\log (2/\delta_k)}{2N_{\mathrm{val}}}}.
            \label{eq:bound_of_generalization_gap_K_G_clip_k}
        \end{align}
        \label{lemma:bound_of_generalization_gap_K_G_clip_k}
    \end{lemmamd}

    \begin{proof}
        We can prove this lemma in the same way as~\cref{lemma:bound_of_generalization_gap_G_clip_k}.
        By the basic Rademacher bound on the maximal generalization gap, with probability at least $1-\delta_k$, we have
        \begin{align}
            \sup_{\kappa \in \mathcal{K}} \phi(\kappa \circ\hat{g}_{\mathrm{clip}}) \le 2\cdot \mathfrak{R}_{N_\mathrm{val}}(\mathcal{K}\circ\mathcal{G}_{\mathrm{clip}, k}) + M_{\mathcal{K},k} \sqrt{\frac{\log (2/\delta_k)}{2N_{\mathrm{val}}}},
            \label{eq:basic_rademacher_bound_K_G_clip_k}
        \end{align}
        where
        \begin{align}
            \mathcal{K}\circ\mathcal{G}_{\mathrm{clip}, k} &\coloneqq \{\kappa \circ g_{\mathrm{clip}}\mid \kappa \in \mathcal{K}, g_{\mathrm{clip}} \in \mathcal{G}_{\mathrm{clip},k}\}, \\
            M_{\mathcal{K},k} &\coloneqq \sup_{h_{\mathrm{clip}} \in \mathcal{K}\circ\mathcal{G}_{\mathrm{clip}, k}, (\bm{x},y), (\bm{x}^{\prime},y^{\prime})} \ell(y, h_{\mathrm{clip}}(\bm{x}))-\ell(y^{\prime}, h_{\mathrm{clip}}(\bm{x}^{\prime})).
        \end{align}
        By the definition of $\mathcal{{K}}$ in~\cref{def:dpgap}, $\kappa(v) = v + \eta(v)$ and $|\eta(v)| \le 4$, we have
        \begin{align}
            M_{\mathcal{K}, k} = \ell(0, B_k+4) - \ell(0, -B_k-4) = B_k+4.
            \label{eq:maximum_deviation_of_loss_kappa}
        \end{align}
        
        For any two functions $h_{\mathrm{clip},1} , h_{\mathrm{clip},2}\in \mathcal{K}\circ\mathcal{G}_{\mathrm{clip}, k}$, we have
        \begin{align}
            \lVert h_{\mathrm{clip},1}(D) - h_{\mathrm{clip},2}(D) \rVert_2 &= \lVert (\kappa_1 \circ g_{\mathrm{clip},1})(D) - (\kappa_2 \circ g_{\mathrm{clip},2})(D) \rVert_2 \\
            &\le \lVert (\kappa_1 \circ g_{\mathrm{clip},1})(D) - (\kappa_1 \circ g_{\mathrm{clip},2})(D) \rVert_2 + \lVert (\kappa_1 \circ g_{\mathrm{clip},2})(D) - (\kappa_2 \circ g_{\mathrm{clip},2})(D) \rVert_2\\
            &\le 2 \lVert g_{\mathrm{clip},1}(D) - g_{\mathrm{clip},2}(D) \rVert_2 + \sqrt{N}\max_{v \in [-B_k, B_k]} |\kappa_1(v) - \kappa_2(v)|.
            \label{eq:bound_of_distance_of_K_G_clip_k_of_D}
        \end{align}
        For given $r^\prime_1> 0$, let $I_{r^\prime_1}$ be a $(r^\prime_1/\sqrt{N})$-covering of $[0, B_k]$.
        Moreover, for given $r^\prime_2$, let $\mathcal{K}_{k,r^\prime_2}$ be a $(r^\prime_2/\sqrt{N})$-covering of $\mathcal{K}$ on the input domain $[-B_k, B_k]$:
        \begin{align}
            \min_{\kappa^\prime\in\mathcal{K}_{k,r^\prime_2}}\max_{v\in[-B_k, B_k]} |\kappa(v)-\kappa^\prime(v)| \le \frac{r^\prime_2}{\sqrt{N}} \quad \forall \kappa \in \mathcal{K}.
        \end{align}
        We define $\mathcal{K}_{k,r^\prime_2}\circ\mathcal{G}_{\mathrm{clip}, k}(D, I_{r^\prime_1})$ as
        \begin{align}
            \mathcal{K}_{k,r^\prime_2}\circ\mathcal{G}_{\mathrm{clip}, k}(D, I_{r^\prime_1}) \coloneqq \{\kappa \circ g_{\mathrm{clip}}(D)\mid \kappa \in \mathcal{K}_{k,r^\prime_2}, g_{\mathrm{clip}}=\clip(g(\cdot), -C, C), C \in I_{r^\prime_1}\}.
        \end{align}
        Then, by~\cref{eq:bound_of_distance_of_K_G_clip_k_of_D,eq:r_covering_of_G_clip_k_of_D}, $\mathcal{K}_{k,r^\prime_2}\circ\mathcal{G}_{\mathrm{clip}, k}(D, I_{r^\prime_1})$ is a $(2r^\prime_1+r^\prime_2)$-covering of $\mathcal{K}\circ\mathcal{G}_{\mathrm{clip}, k}(D)$:
        \begin{align}
            &\min_{\bm{h}_{\mathrm{clip}}^{\prime} \in \mathcal{K}_{k,r^\prime_2}\circ\mathcal{G}_{\mathrm{clip}, k}(D, I_{r^\prime_1})}  \lVert \bm{h}_{\mathrm{clip}}^{\prime} -\bm{h}_{\mathrm{clip}} \rVert_2 \\
            &\le 2 \min_{\bm{g}^\prime \in \mathcal{G}_{\mathrm{clip}, k}(D, I_{r^\prime_1})} \lVert \bm{g}_{\mathrm{clip}}^{\prime} -\bm{g}_{\mathrm{clip}} \rVert_2 + \sqrt{N}\min_{\kappa^\prime \in \mathcal{K}_{k,r^\prime_2}}\max_{v \in [-B_k,B_k]} |\kappa^\prime(v)-\kappa(v)|\\
            &\le 2r^\prime_1+r^\prime_2,
        \end{align}
        for all $\bm{h}_{\mathrm{clip}} = \kappa \circ g_{\mathrm{clip}}(D)\in \mathcal{K} \circ (\mathcal{G}_{\mathrm{clip}, k}(D))$.
        The covering number $[0, B_k]$ with radius $r^\prime_1/\sqrt{N}$ is less than or equal to $\left\lceil B_k\sqrt{N}/(2r^\prime_1)\right\rceil$.
        By~\cref{lemma:covering_number_of_bounded_lipschitz_functions}, the covering number of $\mathcal{K}$ with radius $r^\prime_2/\sqrt{N}$ is less than or equal to $\left(\left\lceil4\sqrt{N}/r^\prime_2\right\rceil\right) \cdot 2^{\left\lceil 2B_k\sqrt{N}/r^\prime_2\right\rceil}$.
        Therefore, we have
        \begin{align}
            \inf_{I_{r^\prime_1}, \mathcal{K}_{k,r^\prime_2}} |\mathcal{K}_{k,r^\prime_2}\circ\mathcal{G}_{\mathrm{clip}, k}(D, I_{r^\prime_1})| \le \left\lceil\frac{B_k\sqrt{N}}{2r^\prime_1}\right\rceil\left(\left\lceil\frac{4\sqrt{N}}{r^\prime_2}\right\rceil\right) \cdot 2^{\lceil 2B_k\sqrt{N}/r^\prime_2\rceil}.
            \label{eq:covering_number_of_K_G_clip_k_of_D}
        \end{align}
        Thus, we can construct an $r$-covering of $\mathcal{K}\circ\mathcal{G}_{\mathrm{clip},k}(D)$ with the cardinality being less than or equal to $\lceil 3B_k\sqrt{N}/r\rceil\cdot\lceil6\sqrt{N}/r\rceil 2^{\lceil 3 B_k\sqrt{N}/r\rceil}$ by setting $r^\prime_1=r/6$ and $r^\prime_2=2r/3$ in~\cref{eq:covering_number_of_K_G_clip_k_of_D}.

        Then, by the same way as~\cref{lemma:bound_of_generalization_gap_G_clip_k}, for any $M>0$ and $r_m\coloneqq \sqrt{N}(B_k+4)/2^{m}$, we have
        \begin{align}
            \hat{\mathfrak{R}}(\mathcal{K}\circ\mathcal{G}_{\mathrm{clip}, k}(D)) \le &\frac{\sqrt{N}(B_k+4)2^{-M}}{\sqrt{N}} + \frac{6\sqrt{N}(B_k+4)}{N}\sum_{m=1}^M 2^{-m} \cdot \sqrt{\log \left(\left\lceil\frac{ 3B_k\sqrt{N}}{r_m}\right\rceil \cdot \left\lceil\frac{6\sqrt{N}}{r_m}\right\rceil \cdot 2^{\lceil 3 B_k\sqrt{N}/r_m\rceil}\right)}\\
            =&(B_k+4)2^{-M} + \frac{6(B_k+4)}{\sqrt{N}}\sum_{m=1}^M 2^{-m} \cdot \sqrt{\log \left(\left\lceil\frac{3B_k\cdot2^{m}}{B_k+4}\right\rceil\cdot\left\lceil\frac{6\cdot2^{m}}{B_k+4}\right\rceil \cdot 2^{\lceil 3 B_k\cdot 2^{m}/(B_k+4)\rceil}\right)} \\
            \le&(B_k+4)2^{-M} + \frac{6(B_k+4)}{\sqrt{N}}\sum_{m=1}^M 2^{-m} \cdot \sqrt{\log \left((3\cdot2^{m})\cdot (2\cdot2^{m})\cdot 2^{(3 \cdot 2^{m})}
            )\right)}\\
            =&(B_k+4)2^{-M} + \frac{6(B_k+4)}{\sqrt{N}}\sum_{m=1}^M 2^{-m} \cdot \sqrt{\log 3+m\log 2 + (m+1)\log 2 + (3\cdot 2^{m}) \log 2}\\
            \le& (B_k+4)2^{-M} + \frac{6(B_k+4)}{\sqrt{N}}\sum_{m=1}^M 2^{-m} \cdot \sqrt{3+3m + 3\cdot 2^{m}} \\
            =& (B_k+4)2^{-M} + \frac{6(B_k+4)}{\sqrt{N}}\sum_{m=1}^M \sqrt{3}\cdot2^{-m/2} \cdot \sqrt{2^{-m}(1+m)+1} \\
            \le& (B_k+4)2^{-M} + \frac{6(B_k+4)}{\sqrt{N}}\sum_{m=1}^M \sqrt{3}\cdot2^{-m/2} \cdot \sqrt{2}\\
            =& (B_k+4)2^{-M} + \frac{6(B_k+4)}{\sqrt{N}}\sqrt{6}(\sqrt{2}+1)\left(1-2^{-M/2}\right) \\
            \le & (B_k+4)2^{-M} + \frac{36(B_k+4)}{\sqrt{N}}\left(1-2^{-M/2}\right).
        \end{align}
        By taking $M\to\infty$, we have
        \begin{align}
            \hat{\mathfrak{R}}(\mathcal{K}\circ\mathcal{G}_{\mathrm{clip}, k}(D)) &\le \frac{36(B_k+4)}{\sqrt{N}}, \\
            \mathfrak{R}(\mathcal{K}\circ\mathcal{G}_{\mathrm{clip}, k}) &\le \frac{36(B_k+4)}{\sqrt{N}}.
        \end{align}
    By applying the above inequality and~\cref{eq:maximum_deviation_of_loss_kappa} to~\cref{eq:basic_rademacher_bound_K_G_clip_k}, we immediately obtain~\cref{eq:bound_of_generalization_gap_K_G_clip_k}.
    \end{proof}
        
    Similar to~\cref{lemma:non_uniform_bound_of_generalization_gap_g_clip}, by the above lemma and Theorem 7.4 in~\cite{shalev2014understanding}, we can obtain~\cref{eq:non_uniform_bound_of_generalization_gap_kappa_g_clip}.
    \begin{proof}[Proof of~\cref{lemma:non_uniform_bound_of_generalization_gap_kappa_g_clip}]
        By the definition of $\mathcal{K}\circ\mathcal{G}_{\mathrm{clip},k}$, we have $\mathcal{K}\circ\mathcal{G}_{\mathrm{clip},1}\subset \mathcal{K}\circ\mathcal{G}_{\mathrm{clip},2} \subset \cdots$ and  $\mathcal{G}_{\mathrm{clip}}=\bigcup_{k=1}^{\infty}\mathcal{K}\circ\mathcal{G}_{\mathrm{clip},k}$.
        Then, for all $k \ge 1$ and $h_\mathrm{clip} \in \mathcal{K}\circ\mathcal{G}_{\mathrm{clip},k}$, the following bounds hold simultaneously:
        \begin{align}
            \phi(h_{\mathrm{clip}}) \le \frac{72(B_{k}+4)}{\sqrt{N_{\mathrm{val}}}} + (B_k+4)\sqrt{\frac{\log (2/(2^{-k}\delta_k))}{2N_{\mathrm{val}}}}.
        \end{align}
        This implies~\cref{eq:non_uniform_bound_of_generalization_gap_kappa_g_clip} (Theorem 7.4 in~\citet{shalev2014understanding}).
    \end{proof}

\subsection{Proof of \texorpdfstring{\cref{thm:relationship_bc_soft_scalar}}{Theorem \ref{thm:relationship_bc_soft_scalar}}}
    \begin{proof}
        By~\cref{thm:relationship_bc_soft}, there exists $\gamma \in \real^K$ such that \begin{align}
        \BCSoftmax_{\tau}(\bm{g}; (a\cdot \bm{1}_K, b\cdot \bm{1}_K))[i] = \Softmax_{\tau}(\bm{g}-\bm{\gamma})[i]\\
        =\begin{cases}
            a & \gamma_i < 0 \\
            b & \gamma_i > 0 \\
            \frac{\exp(g_i/\tau)}{Z} & \gamma_i = 0
        \end{cases},
    \end{align}
    where $Z>0$ is the normalization term.
    Since the lower and upper bounds are the same in all classes, there exist two scalars $c$ and $C$ such that $c \le C$ and
    \begin{align}
        g_i - \gamma_i = c \quad &\forall i \in \{i: \gamma_i < 0\}, \\
        g_i - \gamma_i = C\quad &\forall i \in \{i: \gamma_i > 0\}, \\
        c < g_i < C \quad &\forall i \in \{i: \gamma_i = 0\}.
    \end{align}
    Therefore, we have
    \begin{align}
        \BCSoftmax_\tau(\bm{g}; (a\cdot\bm{1}_K, b\cdot\bm{1}_K)) &= \Softmax_\tau(\bm{g}-\bm{\gamma}) \\
        &=\Softmax_\tau(\clip(\bm{g}, c, C)).
    \end{align}
    \end{proof}

\subsection{Proof of \texorpdfstring{\cref{prop:lb_and_fc}}{Proposition \ref{prop:lb_and_fc}}}
    \begin{proof}
        For any $\bm{u} \in \real^K$, $i \in [K]$, and $c\ge 0$, the clip function and soft-thresholding operator can be written as
        \begin{align}
            \clip(\bm{u}, -c, c)[i] = \begin{cases}
                c &\ \ \ c < u_i \\
                u_i & -c < u_i < c \\
                -c &\ \ \ \ \ \ \ \ \ \ u_i < -c
            \end{cases}, \quad 
            S_c(\bm{u})[i] = \begin{cases}
                u_i - c &\ \ \ c < u_i \\
                0 & -c < u_i < c \\
                u_i + c &\ \ \ \ \ \ \ \ \ \ u_i < -c
            \end{cases}.
        \end{align}
        Therefore, for the clipping function $\clip(\cdot, -c, c)$ and the soft-thresholding operator $S_c$, we have
        \begin{align}
            \clip(\bm{u}, -c, c) = \bm{u} - S_c(\bm{u}).
        \end{align}
        Then, we immediately obtain
        \begin{align}
            \bm{g}_{\mathrm{FC}} &= \bm{W}\clip(\bm{z}, -c, c) + \bm{w}= \bm{W}(\bm{z} - S_c(\bm{z})) + \bm{w} = \left(\bm{W}\bm{z}+\bm{w}\right) -\bm{W}S_c(\bm{z}) = \bm{g}-\bm{\gamma}_{\mathrm{FC}}, \\
            \bm{g}_{\mathrm{LB}} &= \clip(\bm{g}, -c, c) = (\bm{g} - S_c(\bm{g})) = \bm{g}-\bm{\gamma}_{\mathrm{LB}}.
        \end{align}
    \end{proof}

\subsection{Counterexample for the Algorithm Proposed by \texorpdfstring{\citet{parra2025deep}}{Parra-Diaz and Castro-Iragorri}}
    \label{sec:counterexample_for_lb}
    \citet{parra2025deep} proposed the lower-bounded softmax ($\LBSoftmax$), which is the $\BCSoftmax$ with $\bm{b}=\bm{1}_K$ and a uniform lower bound:
    \begin{align}
        \LBSoftmax_{\tau}(\bm{g}, a) \coloneqq \BCSoftmax_{\tau}(\bm{g}; (a\cdot\bm{1}_K, \bm{1}_K)).
        \label{eq:lbsoftmax_parra}
    \end{align}

    \begin{algorithm}
        \caption{$O(K)$ algorithm for computing $\LBSoftmax_\tau$~\citep{parra2025deep}}
        \label{alg:lbsoftmax_parra}
        \begin{algorithmic}[1]
            \Input $\bm{g} \in \real^K, a \in [0, 1/K]$
            \State $\bm{p}^\prime \gets \Softmax_{\tau}(\bm{g})$
            \State $A \gets \{i \in [K]: p^\prime_i \ge a\}$
            \State $p_i \gets a$ for all $i \notin A$
            \State $Z \gets (1- (K-|A|)a)^{-1}\cdot \sum_{i \in A} \exp(g_i/\tau)$
            \State $p_i \gets \exp(g_i/\tau)/Z$ for all $i \in A$
            \Output $\bm{p}$
        \end{algorithmic}
    \end{algorithm}
    \Cref{alg:lbsoftmax_parra} shows the algorithm for~\cref{eq:lbsoftmax_parra} proposed by~\citet{parra2025deep}.
    This algorithm first computes the canonical softmax probabilities $\bm{p}^\prime = \Softmax_{\tau}(\bm{g})$ (in line 1), determines the indices that are not lower-bounded, $A$, by comparing each element in $\bm{p}^\prime$ with the scalar $a$ (in line 2), sets the probabilities to $a$ for all $i \notin A$ (in line 3), and finally re-normalizes the remaining probabilities (in lines 4 and 5).
    
    The core issue with~\cref{alg:lbsoftmax_parra} is its one-step greedy procedure.
    By setting the violating elements to the lower bound $a$, the probability mass available for the remaining elements is reduced.
    Consequently, the renormalization step can push some previously feasible elements (those with $p^\prime_i \ge a$) into the infeasible region ($p_i < a$).
    Since the algorithm terminates without re-checking these violations, it returns an infeasible solution that violates the lower bound constraint.
    Our proposed algorithm,~\cref{alg:bcsoftmax}, avoids this by iteratively identifying the bounded indices.
    
    \paragraph{Counterexample.} Let $K=6$, $\tau=1$, $\bm{g}=\log ((0.01, 0.01, 0.01, 0.01, 0.06, 0.90))^\top$, and $a=0.055$.
    Then, $\bm{p}^\prime = (0.01, 0.01, 0.01, 0.01, 0.06, 0.90)^\top$, $A = \{5,6\}$, and $Z=(1-4\cdot0.055)^{-1}(0.06+0.90)=96/78=16/13$.
    We thus have
    \begin{align}
        \bm{p} &=\left(0.055, 0.055, 0.055, 0.055, \frac{0.06}{Z}, \frac{0.90}{Z}\right)^\top \\
        &=\left(0.055, 0.055, 0.055, 0.055, \frac{78}{1600}, \frac{1170}{1600}\right)^\top,
    \end{align}
    and $p_5 = 78/1600 = 0.04875 < 0.055=a$, i.e., $a\cdot\bm{1}_K \not \preceq \bm{p}$.
    Therefore,~\cref{alg:lbsoftmax_parra} can yield incorrect solutions.

%% file: appendix_algorithms.tex
\section{Algorithms for \texorpdfstring{$\BCSoftmax$}{BCSoftmax}}
\subsection{\texorpdfstring{$O(K)$}{O(K)} Algorithm for Computing \texorpdfstring{$\UBSoftmax$}{UBSoftmax}}
    \label{sec:ubsoftmax_linear}
     Since $\mathsf{Cand}(\bm{g}, \bm{b})$ is monotone by \cref{lemma:feasibility_of_p_of_k}, we can find $\rho$ by a quickselect-like procedure, as shown in \cref{alg:ubsoftmax_linear}.
    In the while loop (lines 6-24), this algorithm (1) picks a pivot $i$ (line 7), (2) divides the candidate set $C$ of $\rho$ into $L$ and $R$ based on $t_k=b_k/\exp(g_k)$ (line 8), (3) determines whether $\rho$ is in $L$ or $R$ (lines 11, 15, and 17), (4) updates the candidate set $C$ and two caches $s$ and $r$ (lines 12, 16, and 18).
    The time complexity of one loop is $O(|C|)$.
    Therefore, this algorithm runs in $O(K)$ time in expectation.
    \begin{algorithm}[t]
    \caption{$O(K)$ algorithm for computing $\UBSoftmax_\tau$}
        \label{alg:ubsoftmax_linear}
        \begin{algorithmic}[1]
            \Input $\bm{g} \in \real^K, \bm{b} \in U^K$
            \State $\bm{g} \gets \bm{g}/\tau$
            \State $t_k \gets b_k / \exp(g_k) \ \forall k \in [K]$, $t_0 \gets -\infty$, $x_0 \gets -\infty$, $b_0 \gets 0$
            \State $s \gets 1$, $r \gets \sum_{k=1}^{K} \exp(g_k)$
            \State $\rho \gets 0, i' \gets \argmax_{k \in [K]} t_k$
            \State $C \gets [0, K]/ \{i'\}$
            \While{$C \neq \emptyset$}
                \State Pick $i$ from $C$
                \State $L \gets \{k\in C: t_k \le t_i\}$, $R \gets C\setminus L = \{k \in C: t_k > t_i\}$
                \State $s' \gets s - \sum_{k \in L} b_k$, $r' \gets  r - \sum_{k \in L} \exp(g_k)$
                \State $j \gets \argmin_{k \in R \cup \{i'\}} t_k$
                \If{$s' \le 0$}
                    \State $C \gets \{k \in L: t_k < t_i\}$, $i'\gets i$
                \Else
                    \State $Z \gets r'/s'$
                    \If{$1/Z \le t_j$}
                        \State $C \gets \{k \in L: t_k < t_i\}$, $i'\gets i$,  $\rho \gets i$                
                    \Else
                        \State $C \gets R$, $s \gets s'$, $r \gets r'$
                    \EndIf
                \EndIf
            \EndWhile
            \State $Z \gets r / s$
            \State $p_k = b_k$ for all $k \in  \{i \in [K]: t_i \le t_\rho\}$
            \State $p_k = \exp(g_k)/Z$ for all $k \in  \{i \in [K]: t_i > t_\rho\}$
            \Output $\bm{p}$
        \end{algorithmic}
    \end{algorithm}
\newpage
\subsection{\texorpdfstring{$O(K^2 \log K)$}{O(K \^ 2 log K)} Algorithm for Computing \texorpdfstring{$\BCSoftmax$}{BCSoftmax}}
    \label{sec:bcsoftmax_gpu_friendly}
    Although~\cref{alg:bcsoftmax} runs in $O(K \log K)$ time, it is not GPU-friendly.
    Thus, in our experiments, we alternatively used a more GPU-friendly but $O(K^2 \log K)$ time algorithm, shown in~\cref{alg:bcsoftmax_gpu_friendly}.
    For the computation of $\UBSoftmax$ (line 4), we used~\cref{alg:ubsoftmax}.
    \begin{algorithm}[t]
        \caption{$O(K^2 \log K)$ algorithm for computing $\BCSoftmax_\tau$}
        \label{alg:bcsoftmax_gpu_friendly}
        \begin{algorithmic}[1]
            \Input $\bm{g} \in \real^K, (\bm{a}, \bm{b}) \in B^K$
            \State $\bm{g} \gets \bm{g}/\tau$
            \State Sort $\bm{g}$, $\bm{a}$, and $\bm{b}$ as $a_{1}/\exp (g_{1}) \ge \cdots \ge a_{K}/\exp (g_{K})$;
            \State $s_k \gets 1 - \sum_{i=1}^k a_i$ for all $k \in [0, K-1]$
            \State Compute $p(k)$ for all $k \in [0, K-1]$
            \State $\rho \gets \min \{k \in [0, K-1]: \bm{a} \preceq p(k) \preceq \bm{b}, p(k) \in \Delta^K\}$
            \State $\bm{p} \gets p(\rho)$
            \State Undo sorting $\bm{p}$
            \Output $\bm{p}$
        \end{algorithmic}
    \end{algorithm}